\documentclass[10pt,twocolumn]{article}

\usepackage[utf8]{inputenc}
\usepackage[T1]{fontenc}
\usepackage[a4paper,margin=0.6in,columnsep=0.3in]{geometry}
\usepackage{graphicx}
\usepackage{amsmath,amssymb}
\usepackage{booktabs}
\usepackage{tabularx,array}
\usepackage{graphicx}
\usepackage{subcaption}
\usepackage{amsmath}
\usepackage{amssymb}

\usepackage[numbers,sort&compress]{natbib}
\usepackage[colorlinks=true,linkcolor=blue,citecolor=blue,urlcolor=blue]{hyperref}

\title{Performance at What Cost? A Sustainability-Aware Performance Index for Cell and Nucleus Instance Segmentation}
\author{%
  Eiram Mahera Sheikh, Alaa Tharwat, Wolfram Schenck\\[2pt]
  {\small Department of Engineering and Mathematics, Bielefeld University of Applied Sciences and Arts,} \\
  	{\small Interaktion 1, 33619 Bielefeld}\\
  {\small \texttt{\{eiram\_mahera.sheikh, alaa.othman, wolfram.schenck\}@hsbi.de}}
}
\date{}

\begin{document}

\twocolumn[
  \maketitle
  \begin{center}
  \begin{minipage}{0.88\textwidth}
    \begin{abstract}
      Pretrained models for cell and nuclear instance segmentation differ substantially in architecture, pretraining data and objectives, parameter count, inference strategy, adaptation requirements, post-processing pipeline, and computational demand. Large pretrained and foundation models are increasingly adopted because of their strong zero-shot capabilities, but their use also imposes greater energy consumption, memory requirements, computational demands, adaptation costs, and operational carbon emissions. Whether these additional demands are justified by meaningful gains in segmentation performance remains unclear. We address this question by introducing the Sustainability-Aware Performance Index (SAPI), a configurable metric that combines segmentation performance, energy consumption, and model size. We benchmark 19 pretrained and foundation models across six CellBinDB datasets under zero-shot inference and evaluate 16 fine-tunable models using few-shot adaptation with both frozen-encoder and full-model fine-tuning. We estimate energy consumption for GPU, CPU, and RAM using software-based monitoring tools. Our results show that larger and more computationally demanding models do not consistently achieve proportionate improvements in segmentation quality. While few-shot adaptation benefits several models, the gains and resource costs vary considerably across architectures, datasets, and adaptation strategies, causing SAPI-based rankings to differ from rankings based on performance alone. This study provides a practical framework for comparing segmentation models more comprehensively and supports more computationally accessible and environmentally responsible model selection in biomedical image analysis.
    \end{abstract}
  \end{minipage}
  \end{center}
  \vspace{1em}
]

\section{Introduction}
Cell and nucleus instance segmentation is a foundational step in quantitative microscopy because many downstream analyses require individual objects to be identified and delineated. Reliable object instance boundaries enable cell counting, morphometric measurement, phenotypic profiling, spatial analysis of cell populations, and the assignment of molecular signals to individual cells. In digital pathology, nuclear segmentation supports the extraction of interpretable cellular features and the characterization of tissue architecture; in multiplex tissue imaging, whole-cell and nuclear masks convert spatially resolved marker measurements into cell-level data; and in experimental biology, segmentation underpins high-content assays in which large image collections must be transformed into reproducible quantitative readouts. These applications span basic cell biology, computational pathology, spatial omics, drug discovery, and biopharmaceutical development, but they share the same practical requirement: segmentation must remain reliable across variation in tissue, staining chemistry, imaging modality, cell morphology, object density, and image quality. Errors introduced at this stage propagate into subsequent measurements, making segmentation quality consequential for the validity of the biological or clinical analysis constructed from the resulting instances \citep{mitchel2026impact}.

Cell and nucleus segmentation was initially approached through classical image-processing pipelines that combined intensity thresholding, filtering, morphological operations, contour detection, distance transforms, and watershed-based separation. These methods were often computationally efficient and interpretable, but their performance depended strongly on manually selected parameters and assumptions about image contrast, object shape, and boundary visibility. Variations in staining, illumination, noise, cell morphology, and object density could therefore require extensive retuning and limit transfer across datasets. Deep convolutional neural networks changed this paradigm by learning hierarchical image representations directly from annotated data and combining contextual encoding with precise spatial localization, as demonstrated by the encoder–-decoder architecture and skip connections introduced in U-Net \citep{ronneberger2015u}. Once reliable pixel-level prediction became possible, subsequent work increasingly addressed the more difficult problem of converting foreground predictions into distinct cellular instances. Convolutional architectures therefore incorporated object-centered geometric representations \citep{schmidt2018cell}, learned displacement fields \citep{stringer2021cellpose}, and dedicated output branches to distinguish adjoining or overlapping nuclei in crowded microscopy and histopathology images \citep{graham2019hover}. These developments moved instance segmentation beyond generic foreground classification by allowing object shape, relative pixel position, and instance boundaries to be learned as part of the prediction task. The next major architectural shift arose with vision transformers, which represented images as sequences of patches and used self-attention to learn interactions among image regions, particularly when pre-trained on large image collections \citep{dosovitskiy2020image}. Transformer encoders were then incorporated into biomedical image segmentation architectures by combining pre-trained encoders with task-specific decoders for cell and nucleus delineation \citep{horst2024cellvit}. Large-scale promptable segmentation introduced a foundation-model paradigm in which a high-capacity image encoder trained on an extensive collection of images and masks, could support mask generation across varied image distributions and segmentation tasks \citep{kirillov2023segment}. This paradigm was then extended to microscopy by adapting foundation--model representations and segmentation interfaces to biological images and microscopy-specific annotation workflows \citep{archit2025segment}. This progression has culminated in large vision transformer–-based foundation models pre-trained on extensive and diverse image collections.

The practical appeal of pre-trained and foundation models lies in their ability to segment images from a new dataset without additional target-specific training. In this zero-shot setting, a model is applied using knowledge acquired during pre-training, thereby avoiding the time and annotation effort required to construct a new model for each imaging study. However, recent comparative evaluations indicate that zero-shot performance remains strongly dependent on the characteristics of the target data \citep{shi2025cellbindb}. When zero-shot predictions do not provide sufficient accuracy, the pre-trained representation can instead serve as an initialization for adaptation to the target domain. Fine-tuning with a small support set of annotated target images allows the model to learn dataset-specific appearance and morphological characteristics without requiring the large annotation collections typically needed to train a segmentation network from scratch. Few-shot fine-tuning studies have shown that pre-trained models can be adapted using only a limited number of annotated samples from the target dataset, although the resulting performance depends on which images are selected for annotation and how well they represent the target distribution \citep{sheikh2026cgmd}.

The growing use of large pre-trained models has made the computational cost of artificial intelligence an increasingly important consideration alongside predictive performance. This concern motivated the emergence of Green AI, which argues that methodological progress should be evaluated not only by improvements in accuracy, but also by the computational resources required to develop and use a model \citep{schwartz2020green}. Early empirical evidence from natural language processing showed that training, hyperparameter optimization, and repeated experimentation with large neural networks could entail substantial energy consumption, financial cost, and associated carbon emissions \citep{strubell-etal-2019-energy}. Subsequent analyses of large-scale neural network training further demonstrated that environmental impact is shaped not only by the amount of computation performed, but also by model architecture, hardware efficiency, data-centre infrastructure, and the carbon intensity of the electricity supply \citep{patterson2021}. These findings led to calls for more systematic measurement and reporting of energy use and carbon emissions in machine-learning research, so that improvements in model quality can be interpreted in relation to the resources required to achieve them \citep{henderson2020towards}. Accordingly, improvements in predictive performance should be interpreted together with the computational resources required to obtain them. This requires explicit consideration of the costs associated with model development, target-domain adaptation, and repeated inference, particularly when large pretrained models are intended for broad or sustained use.

Despite substantial progress in cell and nucleus instance segmentation, model evaluation remains dominated by predictive performance. Existing benchmarks compare segmentation accuracy across imaging modalities, tissues, and model families, but they provide limited evidence about the computational resources required to obtain and sustain that performance. This omission is increasingly important because contemporary segmentation methods differ markedly in model scale, inference requirements, and the amount of target-domain adaptation they require. A model that achieves the highest segmentation score may also consume considerably more energy during inference, require costly fine-tuning, or depend on hardware that is unavailable in many research and deployment settings. Conversely, a computationally efficient model may provide comparable performance at substantially lower operational cost. However, these trade-offs cannot be determined from accuracy, parameter count, or runtime alone. To our knowledge, no controlled benchmark has jointly evaluated zero-shot performance, few-shot adaptation, inference and fine-tuning energy, and model complexity across a diverse collection of specialist, generalist, and foundation models for cell and nucleus instance segmentation. As a result, it remains unclear whether the performance gains offered by larger pretrained models are proportionate to the additional resources required to adapt and deploy them. Addressing this gap is necessary to support model selection that reflects not only segmentation quality, but also computational accessibility, operational efficiency, and the practical requirements of the intended application.

Motivated by this gap, we ask whether the segmentation performance gains offered by larger pretrained and foundation models justify their additional inference energy, adaptation energy, and model complexity. To address this question, our study makes three main contributions. First, we conduct a controlled benchmark of 19 cell and nucleus instance-segmentation models across six CellBinDB datasets \citep{shi2025cellbindb} covering DAPI, single-stranded DNA, multiplex immunofluorescence, and H\&E imaging modalities, with 1,044 images distributed across 30 tissue types and containing over 100,000 annotated cell and nuclear instances. Specifically, we evaluate the models under zero-shot condition to determine the performance provided by their existing pretrained representations without target-domain optimization. For the 16 models with fine-tuning API available to us, we compare frozen-encoder and full-model fine-tuning using tissue-stratified few-shot support sets, thereby examining both the benefits and the computational costs of adapting pretrained models to unknown datasets. Second, we measure energy consumption (GPU, CPU, and RAM) during inference and fine-tuning on a fixed hardware using software based estimates, and quantify model complexity via total parameter count. Third, we introduce the Sustainability-Aware Performance Index (SAPI), a configurable metric that associates segmentation performance with energy demand and model scale. Together, these contributions provide a unified framework for determining when additional model capacity and computational expenditure produce a proportionate improvement in segmentation performance.

The broader objective is to make sustainability and computational accessibility integral components of methodological quality in cell and nuclear segmentation, rather than considerations introduced only after a model has been selected. This perspective does not imply that the smallest model is inherently preferable or that energy efficiency should take precedence over segmentation quality. Instead, additional computational demand should be justified by improvements that are meaningful for the intended application. This principle is relevant to high-throughput pathology, spatial molecular profiling, multiplex imaging, pharmaceutical screening, and academic research, where segmentation models may be applied repeatedly to large image collections under substantially different hardware, memory, and energy constraints. Resource-aware evidence can also strengthen reproducibility by revealing whether reported performance gains depend on computationally intensive adaptation or deployment conditions that may not be accessible across institutions. By jointly evaluating segmentation performance, adaptation requirements, operational energy consumption, and model scale, this study provides a basis for determining whether improvements in accuracy are proportionate to the resources required to achieve and sustain them. More broadly, it frames model selection as a balance among technical performance, generalizability, computational accessibility, and environmental responsibility. Such an approach can support the development and adoption of segmentation methods that are not only accurate, but also reproducible, practically feasible, and appropriate for the biological, clinical, or industrial setting in which they will be used. 

Finally, the remainder of this paper is organized as follows. Section~\ref{sec:Methods} describes the CellBinDB datasets and the model zoo, together with the segmentation performance metrics, energy measurement protocols, the Sustainability-Aware Performance Index, and the experimental study design. Section~\ref{sec:results} presents the results of zero-shot and few-shot evaluations, including segmentation performance and energy consumption. Section~\ref{sec:discussion} discusses the implications of our findings for sustainable model selection, and Section~\ref{sec:conclusion} concludes with a summary of the key findings.

	\section{Methods}
\label{sec:Methods}

\subsection{Datasets}
\label{sec:datasets}

\begin{table}[t]
	\centering
	\caption{Details of the six CellBinDB datasets.}
	\label{tbl:datasets}
	
	\small
	\setlength{\tabcolsep}{2.5pt}
	\renewcommand{\arraystretch}{1.0}
	
	\begin{tabular}{@{}lrrrc@{}}
		\toprule
		Dataset &
		Images &
		\shortstack{Cell\\instances} &
		Tissues &
		\shortstack{Image\\size} \\
		\midrule
		mIF               &  60 &  6{,}013 &  3 & $256 \times 256$ \\
		10xGenomics\_HE  & 100 &  7{,}087 &  3 & $256 \times 256$ \\
		10xGenomics\_DAPI  & 100 &  7{,}745 &  2 & $256 \times 256$ \\
		DAPI              & 203 & 16{,}657 &  6 & $256 \times 256$ \\
		ssDNA             & 276 & 23{,}867 & 20 & $256 \times 256$ \\
		HE              & 305 & 41{,}111 &  6 & $512 \times 512$ \\
		\bottomrule
	\end{tabular}
\end{table}

\begin{figure*}[tp]
	\centering
	\includegraphics[width=1.0\textwidth]{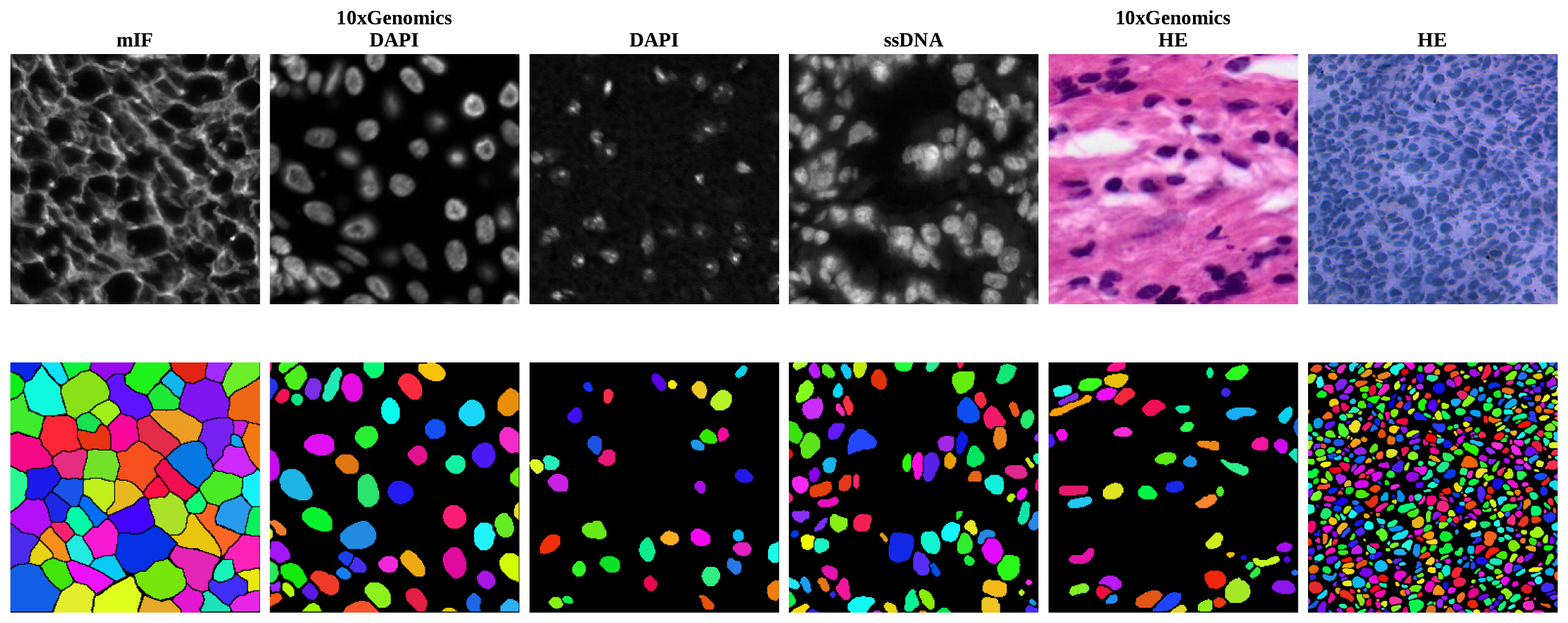}
	\caption{Sample images and their corresponding instance segmentation masks from the six CellBinDB datasets. Each column shows one dataset. The top row shows the microscopy image and the bottom row shows the reference instance masks.}
	\label{fig:dataset_examples}
\end{figure*}

We evaluate all models on the CellBinDB datasets \citep{shi2025cellbindb}, which is a multimodal benchmark for cell instance segmentation that contains expert instance annotations across six imaging modalities. The collection combines in-house Stereo-seq acquisitions with publicly released 10xGenomics FFPE data, spanning four fluorescence modalities (multiplex immunofluorescence (mIF), single-stranded DNA (ssDNA), DAPI, and 10xGenomics\_DAPI) and two haematoxylin--eosin modalities (HE and 10xGenomics\_HE). In total, the datasets have 1{,}044 images containing 102{,}480 annotated cell instances, distributed across 30 distinct tissue types over the six modalities. Table~\ref{tbl:datasets} summarizes the composition of each dataset and Figure~\ref{fig:dataset_examples} shows one sample image with its reference mask from each dataset.

We chose CellBinDB for three reasons. First, it spans several microscopy domains within a single resource that was consistently annotated, so the same evaluation protocol applies unchanged to every model and every domain. Second, none of the evaluated checkpoints was trained on CellBinDB, which makes it an independent test set for both zero-shot evaluation and target-specific adaptation. Third, it varies along the axes that drive the difficulty of the task: staining, tissue type, image source, and image size. The fluorescence and H\&E images differ in their intensity distributions and visual structure, and cell appearance changes further with the staining method and the tissue type. Many images contain densely packed or touching cells, which are hard to separate into individual instances. Others show large variation in cell size, shape, and staining intensity. Together, these differences amount to real domain shifts between the six datasets and tests whether a model holds its performance across heterogeneous microscopy data.

We inverted the mIF images following the same protocol as in the original CellBinDB paper. For the H\&E images we used colour-deconvolution \citep{ruifrok2001quantification} to extract the hematoxylin channel. No other preprocessing was applied by us to any dataset.

\subsection{Model zoo}
\label{sec:model_zoo}

\begin{figure*}[tp]
	\centering
	\includegraphics[width=1.0\textwidth]{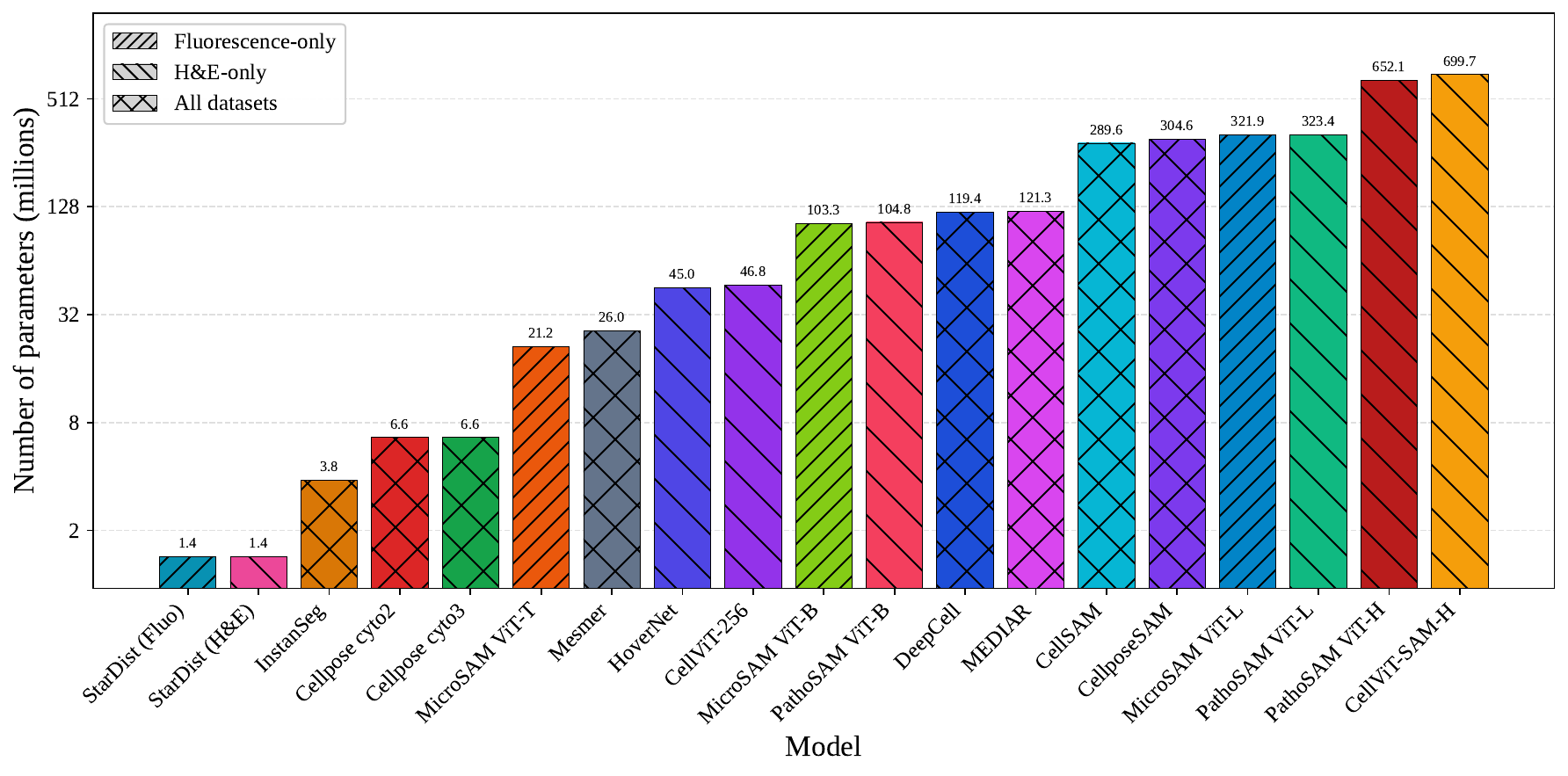}
	\caption{Model size and dataset compatibility of the 19 evaluated models. Bar height represents the total number of parameters ($N_{\theta}$) on a logarithmic scale, with exact counts reported in millions above each bar. Hatch patterns indicate whether a model was used with the fluorescence datasets, the H\&E datasets, or all six CellBinDB datasets.}
	\label{fig:model_zoo}
\end{figure*}

The model zoo was designed to represent the principal methodological choices currently encountered in cell and nuclear instance segmentation. The benchmark includes 19 pretrained models from 11 model families, spanning task-specific convolutional networks, transformer-based segmentation models, and adaptations of general-purpose vision foundation models. These models differ in their encoder architecture and scale, instance-reconstruction strategy, separation of touching objects, pretraining domain, and suitability for different microscopy modalities. Their evaluation across the six datasets was therefore determined by the modality and domain of the data on which the corresponding models were originally trained. Models developed specifically for fluorescence microscopy were evaluated on the fluorescence datasets, models developed for histopathology were evaluated on the H\&E datasets, and models with pretraining that supports both modalities were evaluated across all six CellBinDB datasets. This separation avoids evaluating models outside their intended input domain, where differences in performance could reflect unsupported modality shifts rather than the segmentation capability of the model itself. At the same time, it preserves the practical model-selection setting in which users choose models according to the imaging modality and training domain for which they were developed. As shown in Figure~\ref{fig:model_zoo}, their total parameter counts range from 1.4 million to approximately 700 million, covering more than two orders of magnitude in model size. The patterned bars additionally indicate whether each model was evaluated on the fluorescence datasets only, the H\&E datasets only, or all the six datasets. 

A central distinction among the evaluated models is how they convert pixel-level predictions into separate object instances. StarDist \citep{schmidt2018cell} imposes an explicit geometric prior by representing each object as a star-convex polygon described by radial distances from a candidate centre. Cellpose \citep{stringer2021cellpose} on the other hand predicts spatial flows that guide pixels towards their corresponding object centres. Cellpose-SAM \citep{pachitariu2025cellpose} follows this same decoding and post-processing pipeline; however, it uses a vision transformer encoder. MEDIAR \citep{pmlr-v212-lee23a} also uses a similar flow-based formulation, while combining it with a hierarchical transformer encoder and a multiscale decoder developed for heterogeneous microscopy images. HoVer-Net \citep{graham2019hover} predicts nuclear foreground together with horizontal and vertical displacement maps, which are used to separate touching nuclei in histopathology images. Mesmer \citep{greenwald2022whole} and DeepCell \citep{van2016deep} use multiscale convolutional segmentation networks and distance-based post-processing. InstanSeg \citep{goldsborough2024instanseg} instead learns object-centred pixel representations from which individual instances are recovered. CellSAM  \citep{marks2025cellsam} follows a different design: its CellFinder object detector first localizes cells and generates bounding-box prompts, after which the Segment Anything model (SAM) \citep{kirillov2023segment} produces the corresponding masks. These approaches address the same instance separation problem through different inductive biases. CellViT-256 and CellViT-SAM-H \citep{horst2024cellvit} share the same general histopathology segmentation framework but use markedly different encoders: CellViT-256 employs a comparatively compact DINO-pretrained vision transformer, whereas CellViT-SAM-H uses the substantially larger ViT-H encoder initialized from SAM. Micro-SAM \citep{archit2025segment} and PathoSAM \citep{griebel2025segment} adapt SAM directly to biomedical image domains through finetuning. The three Micro-SAM configurations use light-microscopy-adapted ViT-T, ViT-B, and ViT-L encoders with an automatic instance-segmentation decoder. The three PathoSAM configurations use ViT-B, ViT-L, and ViT-H encoders adapted specifically for histopathology. Within each family, varying the encoder while retaining a common application domain and segmentation formulation creates a controlled model-scale comparison. A comparison of all these models therefore helps to determine whether computationally more elaborate encoders provide greater practical value than carefully designed object representations and post-processing procedures. 

Comparing these models is important because a larger foundation-model encoder may increase representational capacity, but also increases parameter storage, inference energy, memory requirements, and fine-tuning cost. Overall, the model zoo provide structured variation along four dimensions: instance-reconstruction strategy, encoder architecture and scale, modality specialization, and adaptation regime. The inclusion of related variants within Cellpose, CellViT, Micro-SAM, and PathoSAM permits relatively controlled within-family comparisons, whereas the broader set supports comparisons across fundamentally different formulations.

\subsection{Segmentation performance evaluation}
\label{subsec:segmentation_evaluation}

We evaluate segmentation performance using the Aggregated Jaccard Index plus (AJI$^{+}$) \citep{graham2019hover}, Panoptic Quality (PQ) \citep{kirillov2019}, and Normalized Surface Dice (NSD) \citep{nikolov2021} metrics, which capture complementary aspects of segmentation quality. AJI$^{+}$ measures instance-level overlap while accounting for unmatched reference and predicted objects. PQ jointly evaluates instance recognition and the segmentation quality of correctly matched objects. NSD measures agreement between the predicted and reference boundaries within a specified spatial tolerance. All three metrics range from zero to one, with higher values indicating better segmentation quality.

Let \(\mathcal{G}=\{G_1,\ldots,G_K\}\) denote the set of reference instances in an image, and let \(\mathcal{R}=\{R_1,\ldots,R_L\}\) denote the corresponding set of predicted instances, where $K$ and $L$ are the total numbers of reference and predicted instances, respectively. AJI$^{+}$ uses a one-to-one assignment between reference and predicted instances. In our implementation, this assignment was obtained using the Hungarian algorithm without imposing an intersection-over-union threshold. Let \(\mathcal{M}\) denote the resulting set of matched pairs, and let \(\mathcal{U}_{G}\) and \(\mathcal{U}_{R}\) denote the sets of unmatched reference and predicted instances, respectively. AJI$^{+}$ is defined as:
\begin{equation}
	\mathrm{AJI}^{+}
	=
	\frac{
		\displaystyle
		\sum_{(k,l)\in\mathcal{M}}
		\left|G_k \cap R_l\right|
	}{
		\displaystyle
		\sum_{(k,l)\in\mathcal{M}}
		\left|G_k \cup R_l\right|
		+
		\sum_{k\in\mathcal{U}_{G}}\left|G_k\right|
		+
		\sum_{l\in\mathcal{U}_{R}}\left|R_l\right|
	}
	\label{eq:aji_plus}
\end{equation}
The one-to-one assignment ensures that each reference or predicted instance participates in at most one match. Unmatched instances contribute only to the denominator. Consequently, AJI$^{+}$ penalizes missed instances, false-positive detections, merges, and splits while measuring the overall agreement between the reference and predicted instance segmentations.

PQ is calculated using the standard instance-matching criterion. A reference instance and a predicted instance are considered a true-positive match when their intersection over union is strictly greater than \(0.5\). Let \(\mathrm{TP}\) denote the set of matched instance pairs, and let \(\mathrm{FP}\) and \(\mathrm{FN}\) denote the sets of unmatched predicted and reference instances, respectively. Let \(\mathrm{IoU}\) denote the intersection-over-union between the two masks. PQ is defined as:
\begin{equation}
	\mathrm{PQ}
	=
	\frac{
		\displaystyle
		\sum_{(G_k,R_l)\in\mathrm{TP}}
		\operatorname{IoU}(G_k,R_l)
	}{
		\left|\mathrm{TP}\right|
		+
		\frac{1}{2}\left|\mathrm{FP}\right|
		+
		\frac{1}{2}\left|\mathrm{FN}\right|
	}
	\label{eq:pq}
\end{equation}
The numerator measures the segmentation quality of the matched instances, whereas the denominator accounts for both matched instances and recognition errors arising from false-positive and false-negative detections. PQ therefore reflects both the accuracy with which individual objects are identified and the spatial quality of their predicted masks.

For the NSD, the reference and predicted instance label maps are converted into binary foreground masks. NSD therefore evaluates the boundary of the complete foreground region without considering the identities of individual instances. Let \(S_G\) and \(S_R\) denote the surfaces of the reference and predicted foreground masks, respectively. Let \(w(x)\) denote the length associated with surface element \(x\), and let \(d(x,S)\) denote the shortest distance from \(x\) to surface \(S\). For a spatial tolerance \(\tau\), NSD is defined as:
\begin{equation}
	\begin{array}{l}
		\mathrm{NSD}_{\tau}
		=
		\\[0.5ex]
		\displaystyle
		\frac{
			\sum_{x\in S_G}
			w(x)\,
			\mathbb{I}\!\left[d(x,S_R)\leq\tau\right]
			+
			\sum_{y\in S_R}
			w(y)\,
			\mathbb{I}\!\left[d(y,S_G)\leq\tau\right]
		}{
			\sum_{x\in S_G} w(x)
			+
			\sum_{y\in S_R} w(y)
		}
	\end{array}
	\label{eq:nsd}
\end{equation}
Here, \(\mathbb{I}[\cdot]\) denotes the indicator function. A surface element contributes to the numerator when its nearest point on the opposing surface lies within the specified tolerance. In our experiments, NSD is calculated using a tolerance of \(\tau=2\) pixels and a pixel spacing of \([1,1]\) for all the datasets.

AJI$^{+}$, PQ, and NSD are first calculated independently for each test image. Let \(A_i\), \(Q_i\), and \(D_i\) denote the corresponding AJI$^{+}$, PQ, and NSD values for image \(i\). When both the reference and predicted masks contained instances but no valid instance matches were identified, we set AJI$^{+}$ and PQ to zero, and calculate NSD from the corresponding binary foreground surfaces. For a test set containing \(N\) images, the dataset-level metrics are calculated as:
\begin{equation}
	\overline{A}
	=
	\frac{1}{N}\sum_{i=1}^{N} A_i,
	\qquad
	\overline{Q}
	=
	\frac{1}{N}\sum_{i=1}^{N} Q_i,
	\qquad
	\overline{D}
	=
	\frac{1}{N}\sum_{i=1}^{N} D_i
	\label{eq:metric_macro_averages}
\end{equation}
This aggregation assigns equal weight to every test image, preventing images with a larger number of cells or a greater foreground area from disproportionately influencing the dataset-level results.

Because no individual metric provides a complete characterization of segmentation quality \citep{reinke2024understanding, maier2024metrics}, we combined the three dataset-level metrics using their harmonic mean into a single segmentation performance score, denoted by \(P\). The harmonic mean is used because it requires balanced performance across instance overlap, instance recognition, and boundary agreement. A low value for any one component substantially reduces the combined score and cannot be fully compensated for by high values in the remaining components. The performance score is defined as:
\begin{equation}
	P
	=
	\begin{cases}
		0,
		&
		\min\left(\overline{A},\overline{Q},\overline{D}\right)=0,
		\\[1.2ex]
		\displaystyle
		\frac{3}{
			\overline{A}^{-1}
			+
			\overline{Q}^{-1}
			+
			\overline{D}^{-1}
		},
		&
		\text{otherwise}.
	\end{cases}
	\label{eq:performance_score}
\end{equation}

\subsection{Energy measurements}
\label{subsec:energy_measurements}

We measure energy consumption separately for each inference and fine-tuning phase using hardware-level energy counters exposed through the operating system and GPU management interface. Direct wall-socket measurement with an external power meter was not used because the experiments were conducted on a shared data-center compute cluster rather than on a dedicated workstation with an accessible, isolated power connection. In such an environment, an external meter installed at the rack, power-distribution unit, or server power input would measure the aggregate electrical consumption of the corresponding infrastructure and would not directly isolate the energy attributable to the selected GPU and processor domains. Obtaining component-level measurements with external instrumentation would additionally require physical access to the server power-delivery path and coordination with the data-center infrastructure, which was outside the scope of available experimental setup. This is why, we used software accessible hardware counters to provide a practical and reproducible means of obtaining component-level measurements without modifying the cluster infrastructure.

GPU energy measurement was obtained using the NVIDIA Management Library (NVML), accessed through the \texttt{nvidia-ml-py} Python bindings \citep{nvidia_ml_py,nvidia_nvml}. NVML is the NVIDIA-supported programmatic interface underlying \texttt{nvidia-smi} and provides access to GPU power measurements for supported data-center GPUs, including the Tesla V100 used in this study \citep{nvidia_nvml}. The use of NVML is also established in empirical studies of energy consumption on NVIDIA GPUs \citep{tang2019impact,fahad2019comparative}. Processor-package and DRAM energy were obtained from Intel Running Average Power Limit (RAPL) energy counters exposed through the Linux \texttt{powercap} interface \citep{hahnel2012measuring,khan2018rapl,arafa2020verified}. RAPL has been extensively evaluated against external power measurements and has been reported to exhibit strong correlation with plug-level power consumption while introducing negligible measurement overhead, making it suitable for energy monitoring on servers and computing clusters \citep{khan2018rapl}. These interfaces measure energy at the hardware-domain level and thus do not include power consumed by the power supply, cooling system, storage devices, networking equipment, or other infrastructure outside the monitored GPU, CPU-package, and DRAM domains.

The compute node was allocated exclusively to the experiments, and each experiment used one selected GPU without competing GPU processes. The complete energy increment recorded for the selected GPU, CPU packages, and DRAM domains during each measurement interval was attributed to the benchmark workload. This experimental isolation reduces interference from concurrent computational workloads and makes the component-level measurements comparable across models. The same hardware, measurement interfaces, measurement procedure, and measurement scope were used for all evaluated models, which is particularly important for the relative comparison of energy consumption.

The measurement interval encompassed the complete inference or fine-tuning phase executed by the experiment. For inference, this included network instantiation, checkpoint loading, transfer to the GPU, CUDA initialization, model-specific normalization, augmentations, channel adaptation, resizing, rescaling, tiling, forward propagation, transfer of outputs to the CPU, and model-specific instance reconstruction and post-processing. For fine-tuning, the measured interval additionally included loss calculation, backpropagation, parameter updates, and all training epochs. Thus, model-specific computational operations such as Cellpose flow integration, StarDist non-maximum suppression, and SAM-based mask generation were included in the measured energy. Dataset-level transformations performed before model invocation, logging, and metric calculations were excluded because they were not part of the model execution pipeline being evaluated. GPU operations were synchronized immediately before the initial and final counter readings to ensure that asynchronously queued GPU operations were completed before the corresponding energy values were recorded.

Because NVML and RAPL report hardware-domain energy rather than wall-socket electricity, their measurements are not assumed to be exact substitutes for external power-meter measurements. Previous validation studies have shown that the agreement between on-chip energy sensors and external meters depends on the hardware architecture and workload, particularly for GPU measurements \citep{fahad2019comparative}. The present study therefore emphasizes consistency of the measurement procedure and controlled comparison across models rather than interpreting the reported values as exact facility-level electricity consumption.

\subsubsection{GPU energy}

GPU energy was measured using the cumulative device-energy counter provided by NVML. After synchronizing the selected GPU, the counter was read at the start of each measured inference or fine-tuning phase. The GPU was synchronized again after the phase to ensure that all queued operations had completed before the counter was read a second time.

Let $G_{\mathrm{start}}$ and $G_{\mathrm{end}}$ denote the counter readings at the start and end boundaries, respectively. Both values represent the cumulative energy consumed by the GPU since the NVIDIA driver was loaded and are reported by NVML in millijoules (mJ). They do not represent instantaneous GPU power. The GPU energy consumed during the measured phase is calculated as:
\begin{equation}
	E_{\mathrm{GPU}}
	=
	G_{\mathrm{end}}-G_{\mathrm{start}}
	\label{eq:gpu_energy}
\end{equation}
where $E_{\mathrm{GPU}}$ is expressed in millijoules (mJ). Although the counter is cumulative, subtracting the start reading from the end reading isolates the energy consumed between the two measurement boundaries.

This device-level measurement represents the energy recorded for the complete selected GPU rather than for an individual process. Because the selected GPU was used exclusively by our experiment, the complete counter increment was assigned to the evaluated workload.

\subsubsection{CPU and memory energy}

CPU and host-memory energy were obtained using the cumulative RAPL energy counters exposed through the Linux \texttt{powercap} interface. Counter values were recorded at the beginning and end of each measured phase. Processor-package energy is calculated as:
\begin{equation}
	E_{\mathrm{CPU}}
	=
	\frac{
		\displaystyle\sum_{s\in\mathcal{S}}
		\Delta R_{\mathrm{pkg},s}
	}{
		10^{3}
	}
	\label{eq:cpu_energy}
\end{equation}
where $\mathcal{S}$ is the set of detected physical CPU packages and $\Delta R_{\mathrm{pkg},s}$ is the change in the package energy counter for the package $s$, expressed in microjoules. Dividing by $10^{3}$ converts
microjoules to millijoules. The package domain includes the processor cores and other components within the CPU package.

Host-memory energy was calculated separately by summing the increments of all detected DRAM energy counters:
\begin{equation}
	E_{\mathrm{RAM}}
	=
	\frac{
		\displaystyle\sum_{d\in\mathcal{D}}
		\Delta R_{\mathrm{DRAM},d}
	}{
		10^{3}
	}
	\label{eq:ram_energy}
\end{equation}
where $\mathcal{D}$ is the set of detected DRAM energy domains and $\Delta R_{\mathrm{DRAM},d}$ is the counter increment for DRAM domain $d$, expressed in microjoules. Thus, $E_{\mathrm{RAM}}$ is expressed in millijoules. Counter wraparound was handled using the maximum counter range reported by the \texttt{powercap} interface.

\subsubsection{Total IT energy}

The total energy consumed by either a training or an inference phase is calculated as:
\begin{equation}
	E_{\mathrm{IT}}
	=
	E_{\mathrm{GPU}}
	+
	E_{\mathrm{CPU}}
	+
	E_{\mathrm{RAM}}
	\label{eq:it_energy}
\end{equation}

It excludes datacenter infrastructure overhead and uninstrumented components such as storage, networking, motherboard circuitry, and power-supply conversion losses.

\subsubsection{Inference energy}
\label{subsec:inference_energy}

Inference energy is the total energy consumed by the inference phase amortized over the number of images in the test set:
\begin{equation}
	E_i
	=
	\frac{
		E_{\mathrm{IT}}
	}{
		N_{\mathrm{test}}
	}
	\label{eq:inference_energy_per_image}
\end{equation}
where $E_{\mathrm{IT}}$ is the total IT energy measured over the complete inference phase and $N_{\mathrm{test}}$ is the number of test images processed during that phase. Therefore, $E_i$ is expressed in millijoules (mJ) per test image.

\subsubsection{Fine-tuning energy}
\label{subsec:finetuning_energy}
Fine-tuning energy is the total energy consumed during the fine-tuning phase amortized over the number of images in the support set:
\begin{equation}
	E_{\mathrm{ft}}
	=
	\frac{
		E_{\mathrm{IT}}
	}{
		N_{\mathrm{support}}
	}
	\label{eq:training_energy_per_image}
\end{equation}
where $E_{\mathrm{IT}}$ is the total IT energy measured over the complete fine-tuning phase, including all training epochs, and $N_{\mathrm{support}}$ is the number of unique images in the support set. Therefore, $E_{\mathrm{ft}}$ is expressed in millijoules (mJ) per support image and represents the total energy of the complete fine-tuning procedure. 

\subsection{Sustainability-Aware Performance Index (SAPI)}
\label{sec:sapi}

Several composite metrics have been proposed to move model evaluation beyond predictive performance alone. NetScore \citep{wong2019netscore} was introduced to quantify the balance between accuracy, architectural complexity, and computational complexity. For a network $N$, NetScore is defined as:
\begin{equation}
	\Omega(N)
	=
	20\log_{10}
	\left(
	\frac{
		a(N)^{2}
	}{
		\sqrt{p(N) \cdot m(N)}
	}
	\right)
\end{equation}
where $a(N)$ is predictive accuracy, $p(N)$ is the parameter count, and $m(N)$ is the number of multiply-accumulate operations (MAC) required for inference. NetScore represented an important step towards resource-aware model comparison, but its computational term is an analytical operation count only. MAC count does not capture hardware efficiency, memory access, data movement, software implementation, or other system-level effects that can cause models with similar operation counts to consume different amounts of energy.

The authors in the paper \citep{gowda2024watt} subsequently proposed the Sustainable-Accuracy Metric (SAM) to combine predictive accuracy with electricity consumption. SAM is defined as:
\begin{equation}
	\mathrm{SAM}
	=
	\beta
	\frac{
		a^{\alpha}
	}{
		\log_{10}\left(E\right)
	}
\end{equation}
where $a$ is accuracy and $E$ is electricity consumption in kilowatt-hours. This formulation replaces an analytical computational proxy with electricity consumption. Here, $\beta$ is only a global scaling constant; it changes the numerical range of SAM but cannot change the weight of accuracy or energy. Also, SAM contains no independent representation of model scale, meaning that two models with similar accuracy and measured energy receive similar scores even when their storage and memory requirements differ substantially.

To address these limitations, we propose a novel metric, Sustainability-Aware Performance Index (SAPI). It combines predictive performance, energy consumption during inference and fine-tuning, and model parameter count through a weighted multiplicative formulation:
\begin{equation}
	\mathrm{SAPI}
	=
	\frac{
		P^{\alpha}
	}{
		\left[\log_{10}\left(E\right)\right]^{\beta} \cdot
		\left[\log_{10}\left(N_{\theta}\right)\right]^{\gamma}
	}
	\label{eq:sapi_general}
\end{equation}
where $P$ is the segmentation performance as defined in Equation~\ref{eq:performance_score}; $E$ is the energy consumed by the model, and $N_{\theta}$ is the total number of parameters in the model.

The base-10 logarithm is applied to $E$ and $N_{\theta}$ because both quantities vary over several orders of magnitude across model architectures, whereas the performance score ($P$) is bounded between zero and one. Using the unscaled energy and parameter values would allow their substantially larger numerical ranges to dominate the overall SAPI score and would impose a penalty that increases directly with absolute scale. The logarithmic transformation compresses these ranges while preserving their ordering, so that a model with greater energy consumption or more parameters always receives a larger penalty. It also introduces diminishing marginal penalties: an equal multiplicative increase in energy or parameter count produces an equal increment on the logarithmic scale, irrespective of the initial magnitude. Base 10 is used because it provides a direct interpretation in terms of orders of magnitude; for example, a tenfold increase in either quantity increases its logarithm by one. Under these fixed conventions, the logarithmic terms provide interpretable and comparable penalties for operational energy demand and model scale without allowing either component to overwhelm the performance term.

The exponents $\alpha$, $\beta$, and $\gamma$ are positive weights satisfying the constraint:
\begin{equation}
	\alpha+\beta+\gamma=1
	\label{eq:sapi_weight_constraint}
\end{equation}

The exponents $\alpha$, $\beta$ and $\gamma$ control the relative influence of performance, energy consumption, and model complexity, respectively. Requiring $\alpha,\beta,\gamma>0$ preserves the intended monotonic behaviour of the index: higher performance must increase SAPI, whereas higher energy consumption or a larger parameter count must decrease it. A zero exponent would remove the corresponding dimension from the index, while a negative exponent would reverse its meaning by rewarding lower performance or greater resource use. The constraint ($\alpha+\beta+\gamma=1$) is imposed to normalize the exponents into a common weighting scale. Without this constraint, multiplying all three exponents by the same positive constant would change the numerical magnitude and dispersion of SAPI without changing the relative weighting among its components or the resulting model ordering, because the transformed score would simply be a positive power of the original score. Enforcing a unit sum therefore removes this arbitrary degree of freedom and makes each weighting configuration uniquely defined and directly comparable with alternative configurations. It also permits relative priorities, such as (3{:}2{:}1), to be represented unambiguously as ($\alpha$=1/2), ($\beta$=1/3), and ($\gamma$=1/6).

For a model's zero-shot inference, the SAPI score is defined as:
\begin{equation}
	\mathrm{SAPI_{\mathrm{ZS}}}
	=
	\frac{
		P_{ZS}^{\alpha}
	}{
		\left[\log_{10}\left(E_i\right)\right]^{\beta} \cdot
		\left[\log_{10}\left(N_{\theta}\right)\right]^{\gamma}
	}
	\label{eq:sapi_zero_shoot}
\end{equation}
where $P_{ZS}$ is the zero-shot performance of the model and $E_{i}$ is the inference energy per test image as defined in Equation~\ref{eq:inference_energy_per_image}.

For a fine-tuned model, the energy term combines the inference energy of the adapted model per test image with the energy of the complete fine-tuning procedure per support image:
\begin{equation}
	E_{\mathrm{FT+I}}
	=
	\lambda E_{i}
	+
	\left(1-\lambda\right)E_{\mathrm{ft}}
	\label{eq:sapi_adaptation_energy}
\end{equation}
where $ \lambda \in[0,1]$. $E_{\mathrm{ft}}$ is the fine-tuning energy per support image as defined in Equation~\ref{eq:training_energy_per_image} and $E_i$ is the post fine-tuning inference energy per test image as defined in Equation~\ref{eq:inference_energy_per_image}.

The parameter $\lambda$ controls the relative emphasis assigned to inference energy, whereas ($1-\lambda$) controls the emphasis assigned to fine-tuning energy.
This weighting is motivated by the different temporal roles of the two energy costs. Fine-tuning is typically performed once, or only occasionally, for a given model and target dataset, whereas inference is repeated for every input processed after deployment. Consequently, the contribution of the initial adaptation cost becomes progressively amortized as the number of inferences increases, while cumulative inference energy continues to grow with deployment volume \citep{chien2023reducing, lbath2024energy, chou2025energy}.

For fine-tuning followed by inference, SAPI is defined as:
\begin{equation} 
	\mathrm{SAPI}_{\mathrm{FT+I}} = 
	\frac{ P_{\mathrm{FT}}^{\alpha} }
	{ \left[ \log_{10} \left( E_{\mathrm{FT+I}} \right) \right]^{\beta} \cdot \left[ \log_{10}\left(N_{\theta}\right) \right]^{\gamma} } 
	\label{eq:sapi_ft_inference} 
\end{equation}
where $P_{\mathrm{FT}}$ is the performance of the fine-tuned model.

\subsection{Experimental Study Design}
\label{sec:experment_design}

We evaluated the models under three experimental conditions: zero-shot inference, few-shot full-model fine-tuning, and few-shot fine-tuning with a frozen encoder. Because CellBinDB does not provide predefined training, validation, and test partitions, we defined the support and test sets for the few-shot experiments as described below. The same evaluation protocol was used for all applicable model-dataset combinations.

\subsubsection{Zero-shot Evaluation}

All 19 model configurations were evaluated in the zero-shot setting. Each model was applied to all the images in all the datasets supported by its respective pretrained checkpoint. No image was used to update model parameters, select inference thresholds, or modify model-specific post-processing settings. Predictions were generated using the native inference pipeline of each implementation. No manually supplied point, box, or mask prompts were used. Models that support prompt-based inference were instead evaluated using their automatic instance-segmentation procedures. Model-specific automatic settings, such as automatic diameter estimation in Cellpose, were retained where applicable. This experimental setting allowed us to assess how the models perform on the CellBinDB datasets out of the box, without any dataset-specific adaptation.

\subsubsection{Few-shot Adaptation}

Pretrained segmentation models can be adapted to a target microscopy domain using a small number of labeled images \citep{pachitariu2022cellpose,archit2025segment,dawoud2023knowing}. However, the effectiveness of few-shot adaptation can depend on the images selected for fine-tuning \citep{sheikh2026cgmd}. This is particularly relevant for microscopy instance segmentation, where producing accurate object-level annotations is laborious and time-consuming \citep{englbrecht2021automatic}.

For few-shot fine-tuning, we used the following procedure for selecting the support set and test set. For each dataset, we randomly selected one image, using a predefined seed, from a tissue group within the dataset as a support image. The total number of support images therefore equals to the number of tissues present in the dataset, as reported in Table~\ref{tbl:datasets}. All the remaining images in the dataset formed the test set. For a given dataset and seed, the same support and test sets was used for every compatible model.

We did not create a separate validation set because splitting the already small support set would further reduce the number of labeled images available for adaptation. An exploratory analysis of training-loss behavior was used to select a common training duration of 100 epochs. All fine-tuning experiments were therefore run for 100 epochs, and the model obtained after the final epoch was used for evaluation. Model-specific default training hyperparameters were retained with a batch size of 8 used for all models.

Models can be fine-tuned either by updating all parameters (full-model regime) or by keeping the encoder weights frozen while updating only the remaining trainable components (frozen-encoder regime). In the full-model regime, all model parameters were trainable from the first epoch. In the frozen-encoder regime, the encoder or backbone parameters were held fixed, while the model-specific decoder, prediction heads, and other designated adaptation components remained trainable. The relative effectiveness of these strategies depends on the architecture, the size of the support set, and the degree of domain shift between the pretraining data and the target dataset. Freezing the encoder can preserve transferable pretrained representations and reduce the number of trainable parameters, but it may also restrict the model’s ability to adapt its feature representation to the target domain. Previous studies have therefore reported differing outcomes for frozen-encoder and full-model adaptation \citep{xie2024sam,gu2024build}. We evaluate both possibilities under comparable conditions. We train and evaluate each fine-tunable model in both regimes using exactly the same support set and test test.

Fine-tuning was performed for 16 of the 19 model configurations. DeepCell NuclearSegmentation, Mesmer, and CellSAM were evaluated only in the zero-shot setting because their fine-tuning interfaces were not available to us. Information on the exact model checkpoint names and software versions are provided in the Supplementary material.

\subsubsection{Repeated Experiments}

Each experiment was repeated ten times using seeds 0 to 9. For each seed, a tissue-stratified support set was sampled independently. Repeating the experiments across these support sets reduced dependence on a single support-set selection. For a given dataset and seed, the support set and test was kept the same across all the models and both fine-tuning regimes. 

The seeds were propagated to Python, NumPy, PyTorch, and TensorFlow where applicable. Deterministic cuDNN execution was enabled and cuDNN benchmarking was disabled. These settings reduced uncontrolled variation from sampling, data ordering, augmentation, model initialization where applicable, and stochastic framework operations, although complete bitwise determinism cannot be guaranteed for every third-party implementation.

\subsubsection{Computational Setup}
All experiments were conducted on the same compute node equipped with two NVIDIA Tesla V100 PCIe GPUs with 32~GB memory each, two Intel Xeon Gold 6226R processors at 2.90~GHz with 16 physical cores per processor, and 501~GiB of system memory. Only GPU~0 was used for the experiments, and no other GPU processes were running on the node.

To reduce the effect of the GPU's thermal state on the energy measurements, the GPU was allowed to cool to 45$^\circ$C before each measurement. The same cooldown procedure was applied before each experiment. Energy consumption was then measured during the execution of the corresponding inference or fine-tuning workload using the procedure described in Section~\ref{subsec:energy_measurements}.

\section{Results}
\label{sec:results}

\begin{figure*}[tp]
	\centering
	\includegraphics[width=1.0\textwidth]{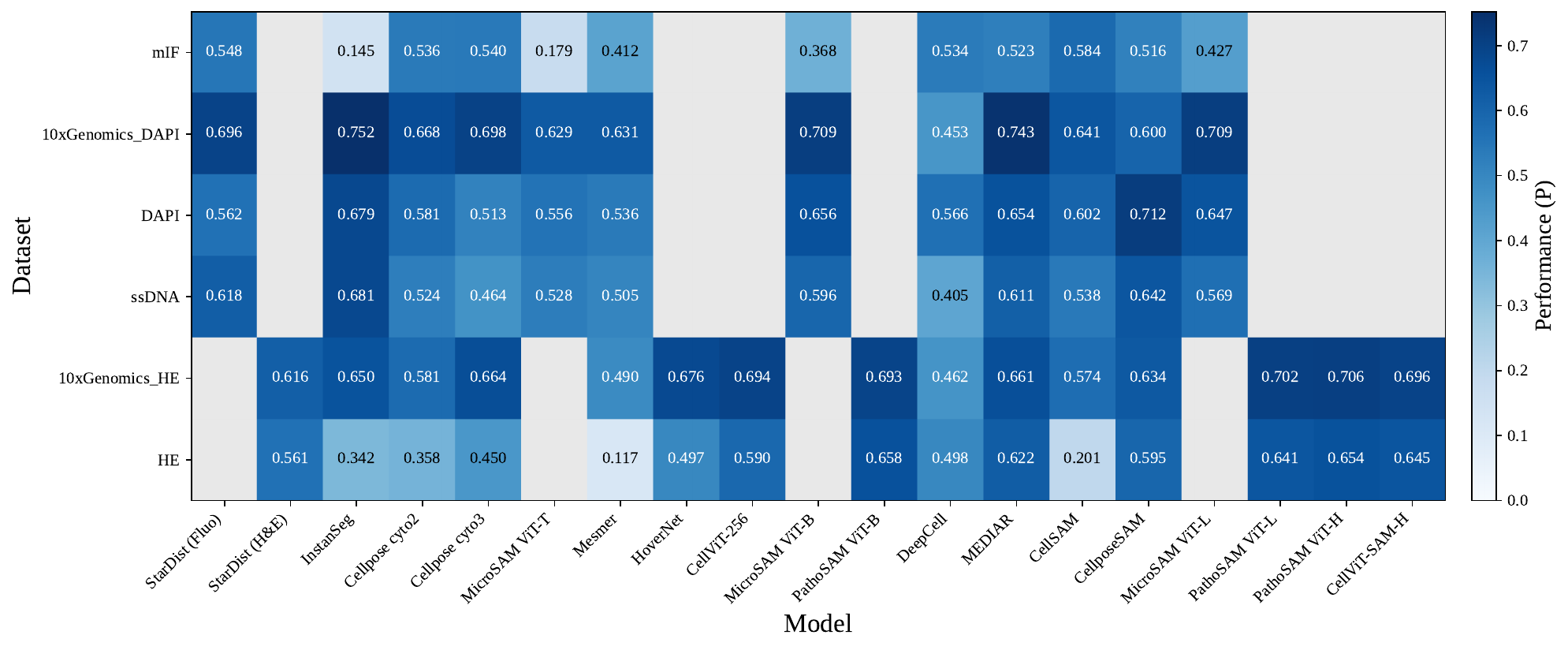}
	\caption{Zero-shot segmentation performance across compatible model--dataset combinations. Each cell reports the mean performance score $P$, calculated as the harmonic mean of AJI$^{+}$, PQ, and NSD. Grey cells denote model--dataset combinations that were not evaluated because the model is not intended to be used with that dataset modality.}
	\label{fig:zero_shot_performance}
\end{figure*}

All primary sustainability-aware comparisons were conducted using the predefined SAPI weights $\alpha$=1/2, $\beta$=1/3, and $\gamma$=1/6, corresponding to a performance–-energy–-size weighting ratio of (3{:}2{:}1). This weighting gives segmentation performance the greatest influence because predictive accuracy remains the primary requirement for model utility, while assigning substantial importance to energy consumption and a smaller, but non-negligible, contribution to model size. The combined energy term was calculated using $\lambda$=0.8, so that fine-tuning energy consumption contributed 20\% and post fine-tuning inference energy consumption contributed 80\%. This choice represents a deployment-oriented setting in which inference is expected to be performed repeatedly after adaptation, while still accounting for the computational cost incurred during fine-tuning. Unless otherwise stated, all SAPI rankings reported below use these fixed values.

\subsection{Zero-shot inference}
\label{subsec:zero_shot_results}

We performed zero-shot inference to compare the models without any adaptation to the CellBinDB datasets. The results showed clear differences in segmentation performance, inference energy consumption, and sustainability-aware utility across models and datasets. Performance also depended strongly on the dataset, and the models with the highest segmentation scores did not always achieve the highest $\mathrm{SAPI}_{\mathrm{ZS}}$ values. We therefore present the results in four parts: predictive performance, inference energy consumption, the joint relationship between performance, energy use, and model size, and the resulting $\mathrm{SAPI}_{\mathrm{ZS}}$ rankings. In the supplementary material we provide qualitative comparisons of the predicted masks produced by every compatible model for one sample image from each dataset, together with the corresponding input image and ground-truth mask.

\subsubsection{Performance}
\label{subsec:zero_shot_performance}

Figure~\ref{fig:zero_shot_performance} shows the zero-shot performance score $P$ for every compatible model-dataset combination. Across these combinations, $P$ ranged from 0.117 for Mesmer on H\&E to 0.752 for InstanSeg on 10xGenomics\_DAPI, demonstrating substantial variation in zero-shot
transferability across models and imaging modalities. 

The strongest model differed among the four fluorescence datasets. InstanSeg achieved the highest performance on 10xGenomics\_DAPI ($P$=0.752) and ssDNA ($P$=0.681), whereas CellposeSAM performed best on DAPI ($P$=0.712). CellSAM attained the highest score on mIF ($P$=0.584). The corresponding AJI$^{+}$, PQ, and NSD metrics are reported in the Supplementary material. The highest H\&E performance was obtained predominantly by histopathology-specific models. PathoSAM ViT-H achieved the highest score on both 10xGenomics\_HE ($P$=0.706) and HE ($P$=0.654) datasets.

\subsubsection{Inference energy consumption}
\label{subsec:zero_shot_energy}

\begin{figure*}[t]
	\centering
	\includegraphics[width=1.0\textwidth]{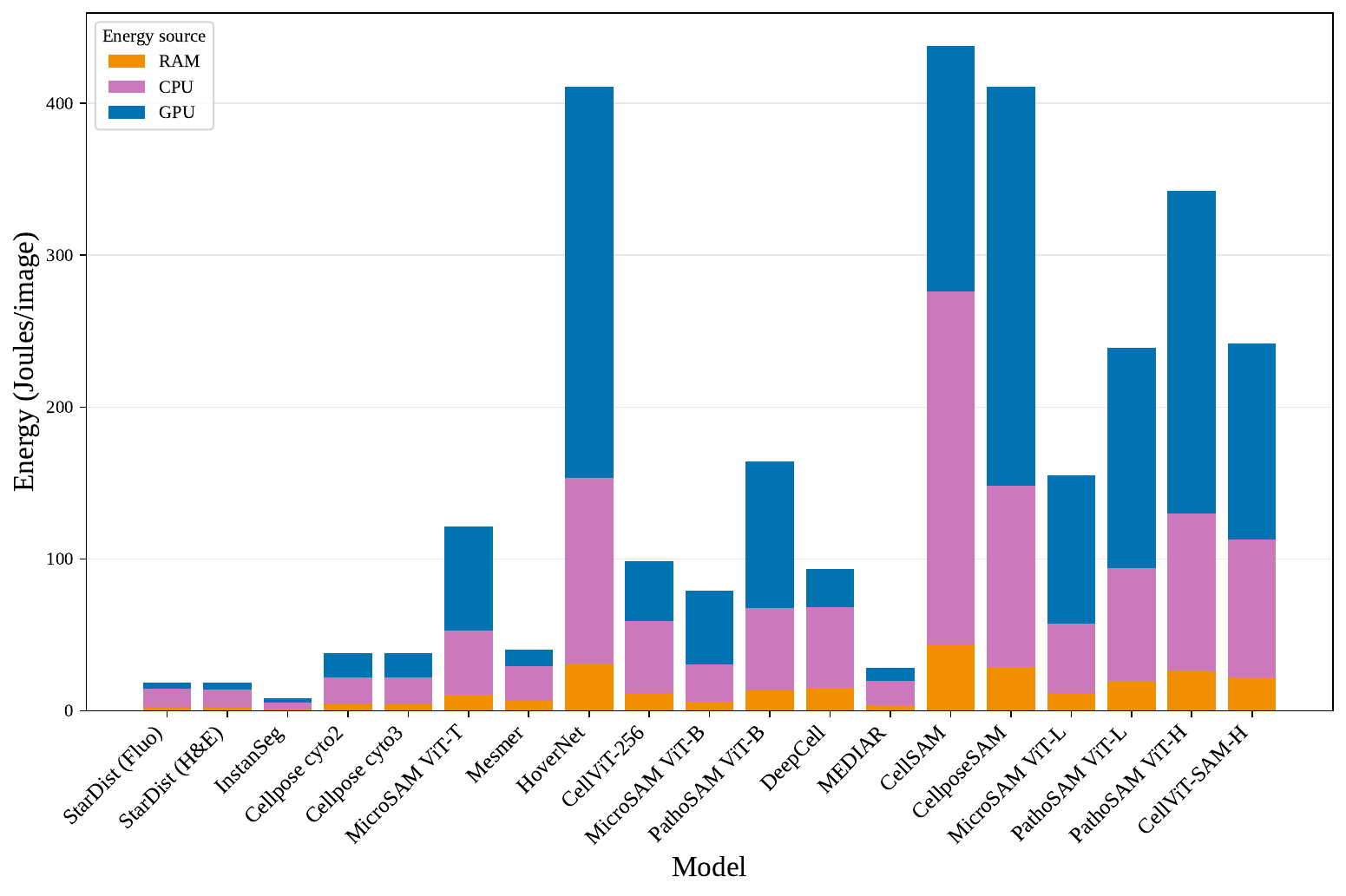}
	\caption{Mean zero-shot inference energy per test image, separated into GPU, CPU, and RAM contributions. Energy values were averaged across experimental seeds and the datasets compatible with each model. The total height of each stacked bar represents the total measured IT energy per image.}
	\label{fig:zero_shot_energy}
\end{figure*}

We measured the zero-shot inference energy used by the GPU, CPU, and RAM for each model. Figure~\ref{fig:zero_shot_energy} shows the mean energy consumed per test image, averaged across experimental seeds and the datasets compatible with each model. The total energy consumption varied widely across the model zoo.

InstanSeg consumed the least energy, with a mean of 8.36~J per image. The two StarDist configurations also had relatively low energy requirements, consuming around 8~J per image. MEDIAR followed at 28.27~J per image. In contrast, CellSAM consumed the most energy at 437.65~J per image. CellposeSAM and HoVer-Net had similarly high energy
requirements, consuming 410.91 and 410.72~J per image, respectively. CellSAM therefore consumed approximately 52 times more energy per image than InstanSeg.

The contribution of each hardware component also differed among models. CPU energy formed the largest component for several of the lower-energy models, including InstanSeg and both StarDist configurations. GPU energy dominated the total consumption of most of the higher-energy models, including CellposeSAM, HoVer-Net, and the PathoSAM variants. CellSAM was a notable exception: despite having the highest total energy consumption, most of its measured energy came from the CPU. RAM made the smallest contribution for every model. These results show that zero-shot inference energy differed not only in total magnitude but also in how the workload was distributed across the measured hardware
components.

\subsubsection{Performance vs Energy vs Model size}
\label{subsec:zero_shot_energy_vs_p}

\begin{figure*}[tp]
	\centering
	\includegraphics[width=1.0\textwidth]{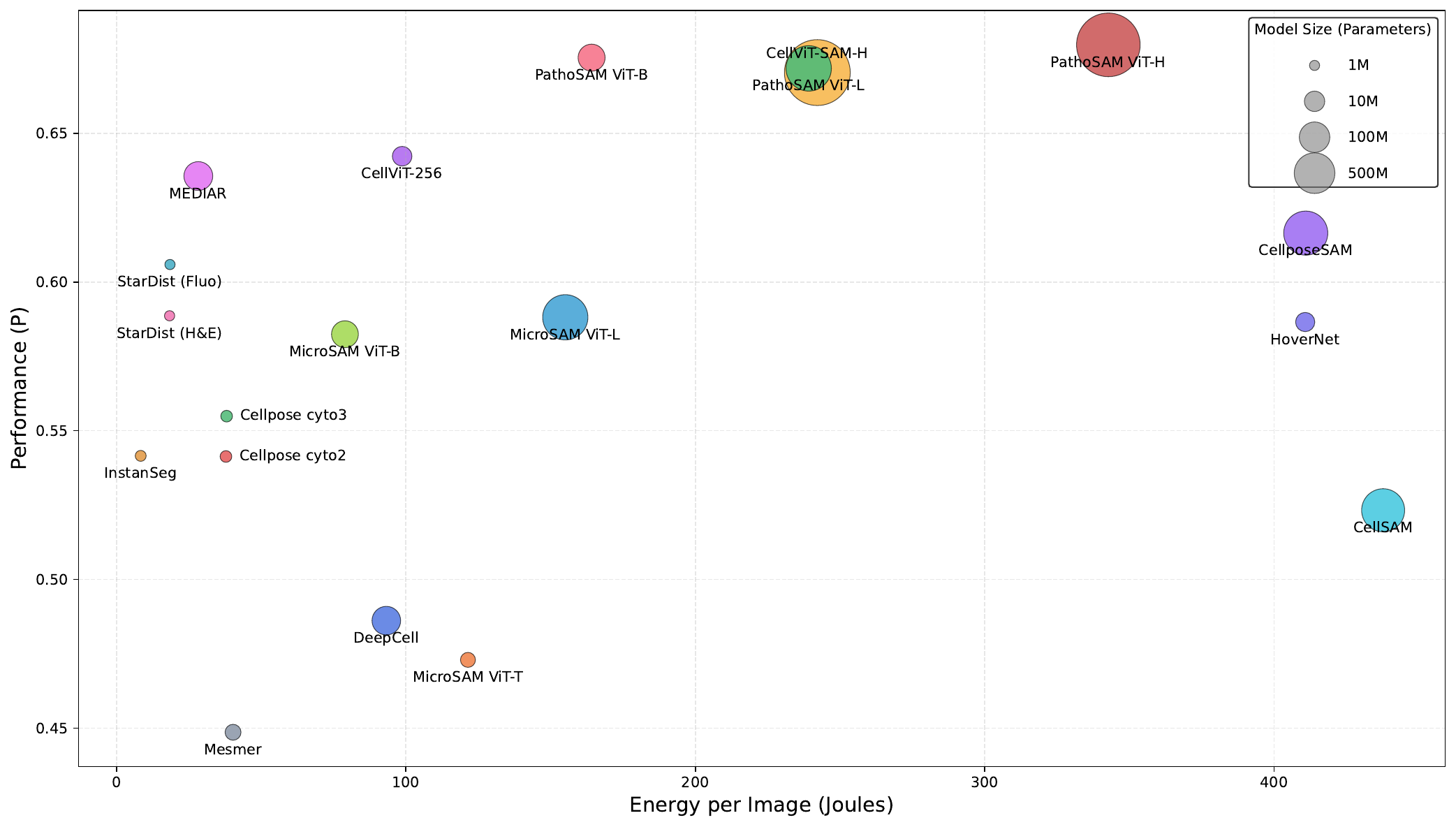}
	\caption{Relationship between zero-shot segmentation performance, inference energy consumption, and model size. Each point represents one model. Performance and energy were averaged across ten runs and then across the datasets compatible with that model. The horizontal axis	shows total inference energy per image on a logarithmic scale, and marker area represents the total number of model parameters.}
	\label{fig:zero_shot_energy_vs_p}
\end{figure*}

We compared mean zero-shot performance with inference energy and model size in Figure~\ref{fig:zero_shot_energy_vs_p}. Across the 19 model configurations, Spearman correlation analysis showed almost no monotonic association between performance and inference energy ($\rho=-0.020$) and only a weak positive association between performance and parameter count ($\rho=0.197$). In contrast, parameter count showed a strong positive association with inference energy ($\rho=0.747$). Thus, larger models generally required more energy for inference, but greater model size or energy consumption was not consistently associated with higher zero-shot segmentation performance.

Several relatively small and energy-efficient models achieved competitive performance. StarDist (Fluo), for example, reached a mean performance of 0.606 while using only 18.47~J per image with approximately 1.4 million parameters. MEDIAR also combined strong performance with low energy use. It achieved a mean performance of 0.636 across all six datasets while consuming 28.27~J per image. In comparison, CellposeSAM consumed 410.91~J per image but achieved a lower mean performance of 0.616 over the same datasets. CellSAM and HoVer-Net also had high energy requirements without reaching the highest performance levels. These results show that high inference energy consumption did not necessarily translate into better zero-shot segmentation.

Among the models evaluated on the two H\&E datasets, the PathoSAM variants and the two CellViT configurations achieved the highest mean performance. However, this pattern did not extend to all pathology-specific models, as HoVer-Net and StarDist (H\&E) achieved lower mean performance values. The PathoSAM family provides a clear example of the trade-off between performance, energy consumption, and model size. PathoSAM ViT-B achieved a mean performance of 0.675 with 104.8 million parameters and consumed 164.15~J per image. PathoSAM ViT-H increased the mean performance only slightly to 0.680, while increasing the model size to 652.1 million parameters and the energy consumption to 342.71~J per image. Thus, the ViT-H variant used approximately twice as much energy and more than six times as many parameters as the ViT-B variant for only a tiny increase in segmentation performance.

\subsubsection{$\mathrm{\textbf{SAPI}_{\mathrm{\textbf{ZS}}}}$}

\begin{figure*}[tp]
	\centering
	
	\begin{subfigure}[t]{0.97\textwidth}
		\centering
		\includegraphics[width=\linewidth]
		{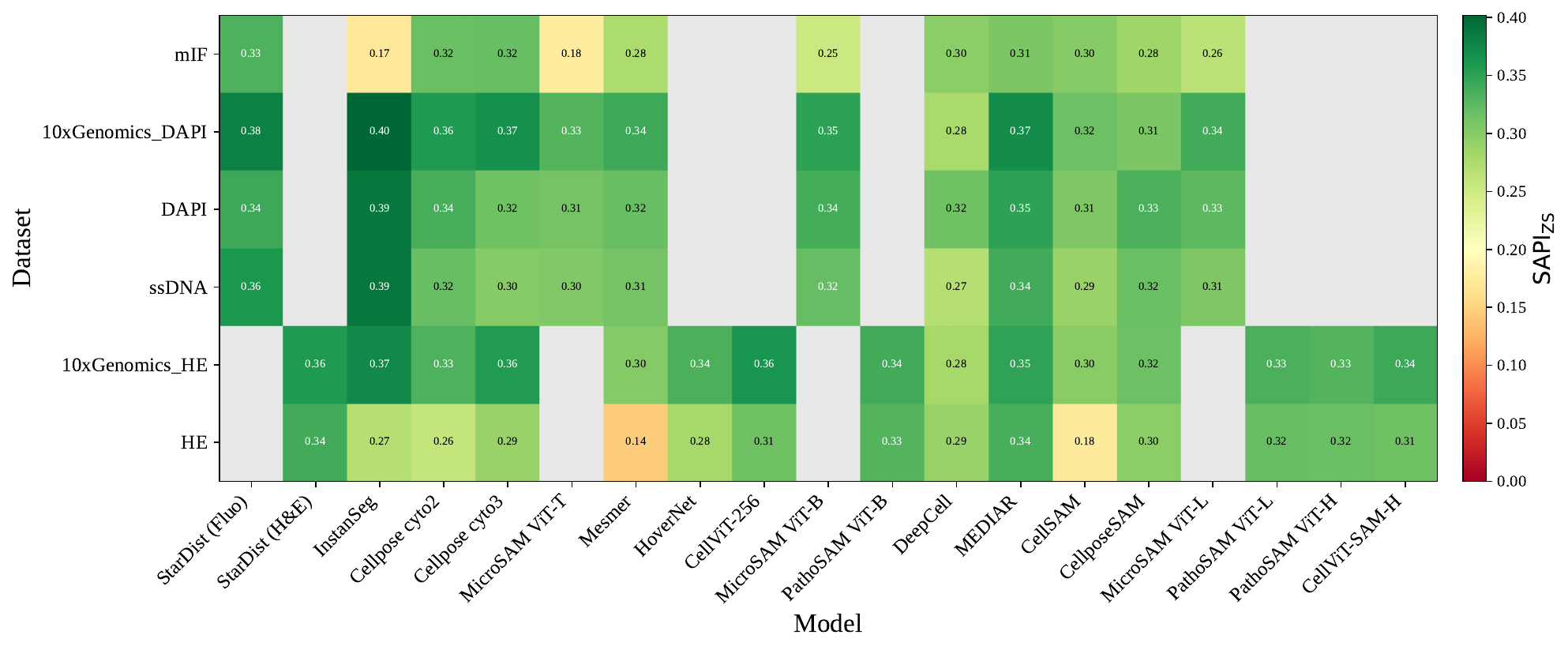}
		\caption{Zero-shot Sustainability-Aware Performance Index ($\mathrm{SAPI}_{\mathrm{ZS}}$) across compatible model-dataset combinations. Each cell reports the mean score across ten runs. Grey cells indicate combinations that were not evaluated because the model is not intended to be used with the corresponding image modality.}
		\label{fig:zero_shot_sapi_heatmap}
	\end{subfigure}
	
	\vspace{0.8em}
	
	\begin{subfigure}[t]{0.97\textwidth}
		\centering
		\includegraphics[width=\linewidth]
		{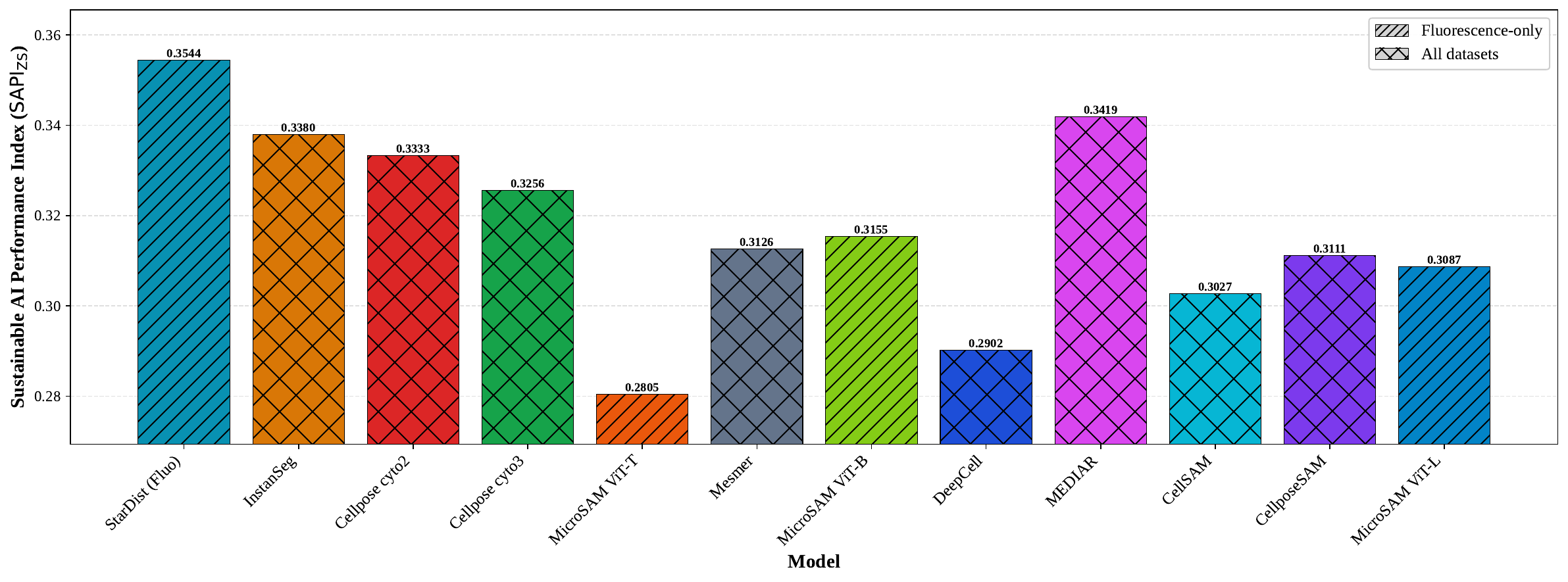}
		\caption{$\mathrm{SAPI}_{\mathrm{ZS}}$ scores average across ten runs and the four fluorescence datasets: mIF, DAPI, ssDNA and 10xGenomics\_DAPI. Hatching pattern distinguishes fluorescence-only models from models evaluated on all the datasets.}
		\label{fig:zero_shot_sapi_bar_fluo}
	\end{subfigure}
	
	\vspace{0.8em}
	
	\begin{subfigure}[t]{0.97\textwidth}
		\centering
		\includegraphics[width=\linewidth]
		{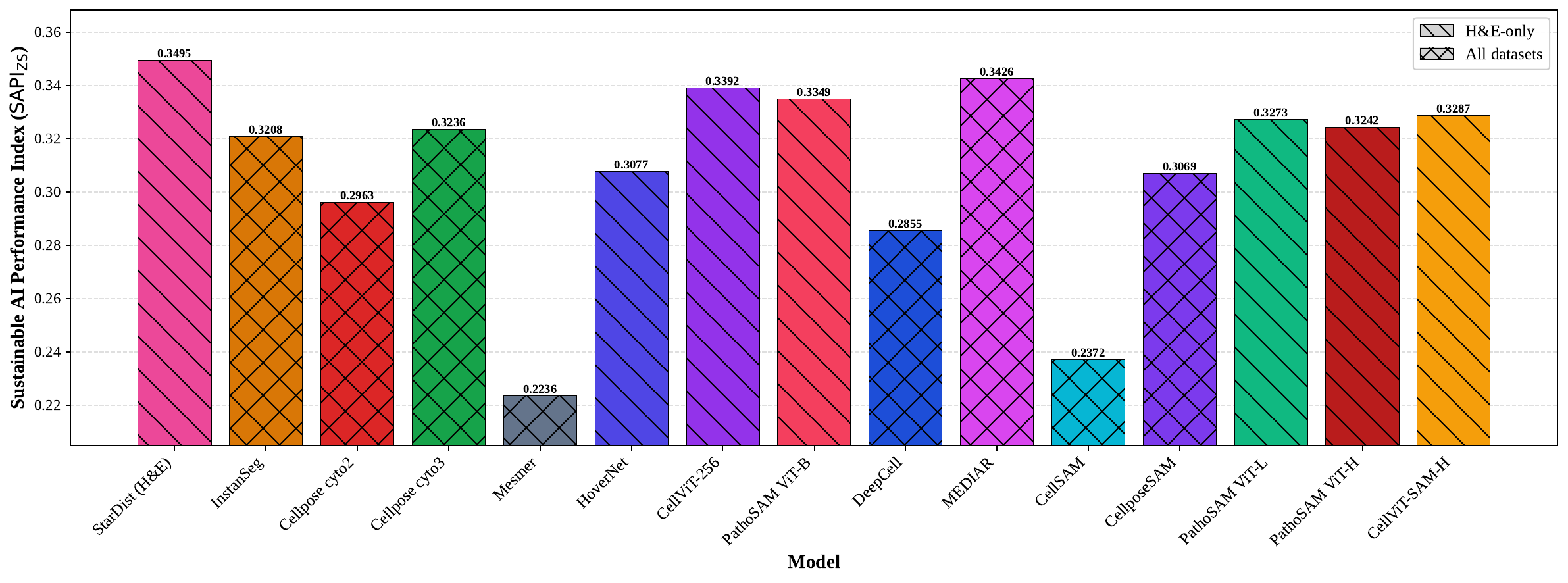}
		\caption{$\mathrm{SAPI}_{\mathrm{ZS}}$ scores average across ten runs and the two H\&E datasets: HE and 10xGenomics\_HE. Hatching pattern distinguishes H\&E-only models from models evaluated on all the datasets.}
		\label{fig:zero_shot_sapi_bar_he}
	\end{subfigure}
	
	\caption{Zero-shot sustainability-aware performance across models and
		datasets.}
	\label{fig:zero_shot_sapi}
\end{figure*}

We calculated $\mathrm{SAPI}_{\mathrm{ZS}}$ for each compatible
model-dataset combination to assess segmentation performance together with inference energy consumption and model size. Figure~\ref{fig:zero_shot_sapi_heatmap}
shows that the $\mathrm{SAPI}_{\mathrm{ZS}}$ score for each compatible
model-dataset combination.

InstanSeg achieved the highest $\mathrm{SAPI}_{\mathrm{ZS}}$ on four datasets: 10xGenomics\_DAPI (0.4019), DAPI (0.3869), ssDNA (0.3890), and 10xGenomics\_H\&E (0.3729). StarDist (Fluo) ranked first on mIF with a score of 0.3336, while StarDist (H\&E) ranked first on the HE dataset with a score of 0.3404. These results show that $\mathrm{SAPI}_{\mathrm{ZS}}$ rankings differ from the rankings based on performance alone.

Figure~\ref{fig:zero_shot_sapi_bar_fluo} compares the mean
$\mathrm{SAPI}_{\mathrm{ZS}}$ across the four fluorescence datasets.
StarDist (Fluo) achieved the highest mean score at 0.3544, followed by MEDIAR at 0.3419 and InstanSeg at 0.3380. Although InstanSeg ranked first on three of the four fluorescence datasets, its substantially lower score on mIF reduced its overall mean. MEDIAR achieved the highest mean segmentation performance across the fluorescence datasets, but StarDist (Fluo) ranked higher under $\mathrm{SAPI}_{\mathrm{ZS}}$ because it combined competitive performance with much lower energy consumption and a smaller model size. By contrast, CellposeSAM achieved strong segmentation performance but received a lower mean $\mathrm{SAPI}_{\mathrm{ZS}}$ of 0.3111 because of its greater energy and parameter requirements.

A similar pattern appeared across the two H\&E datasets
(Figure~\ref{fig:zero_shot_sapi_bar_he}). StarDist (H\&E) achieved the highest mean $\mathrm{SAPI}_{\mathrm{ZS}}$ at 0.3495, followed by MEDIAR at 0.3426, CellViT-256 at 0.3392, and PathoSAM ViT-B at 0.3349. The larger PathoSAM variants achieved some of the highest mean segmentation scores on the H\&E datasets, but their higher energy consumption and parameter counts reduced their $\mathrm{SAPI}_{\mathrm{ZS}}$ values. For example, PathoSAM ViT-H achieved the highest mean H\&E performance but obtained a lower mean $\mathrm{SAPI}_{\mathrm{ZS}}$ of 0.3242.

Overall, $\mathrm{SAPI}_{\mathrm{ZS}}$ did not simply reproduce the
performance ranking. It favoured models that maintained strong segmentation performance while using less energy and fewer parameters. The variation across datasets also shows that the most favourable model depended on the imaging modality and dataset characteristics.

\subsection{Few-shot Adaptation}
\label{subsec:few_shot_results}

For each dataset, we randomly selected one image from every tissue group to form the support set and used all the remaining images as the test set. We fine-tuned the model on the support set using two strategies i.e. full-model fine-tuning and frozen-encoder fine-tuning. In full-model fine-tuning, we updated all model parameters. In frozen-encoder fine-tuning, we kept the encoder weights fixed and updated the remaining trainable parameters. We evaluated the zero-shot checkpoint and both fine-tuned variants on the same test set. We repeated the complete support-set selection, fine-tuning, and evaluation procedure with ten different random seeds to capture variation caused by support set and test set composition.

\subsubsection{Performance}

\begin{figure*}[tp]
	\centering
	\includegraphics[width=1.0\textwidth]{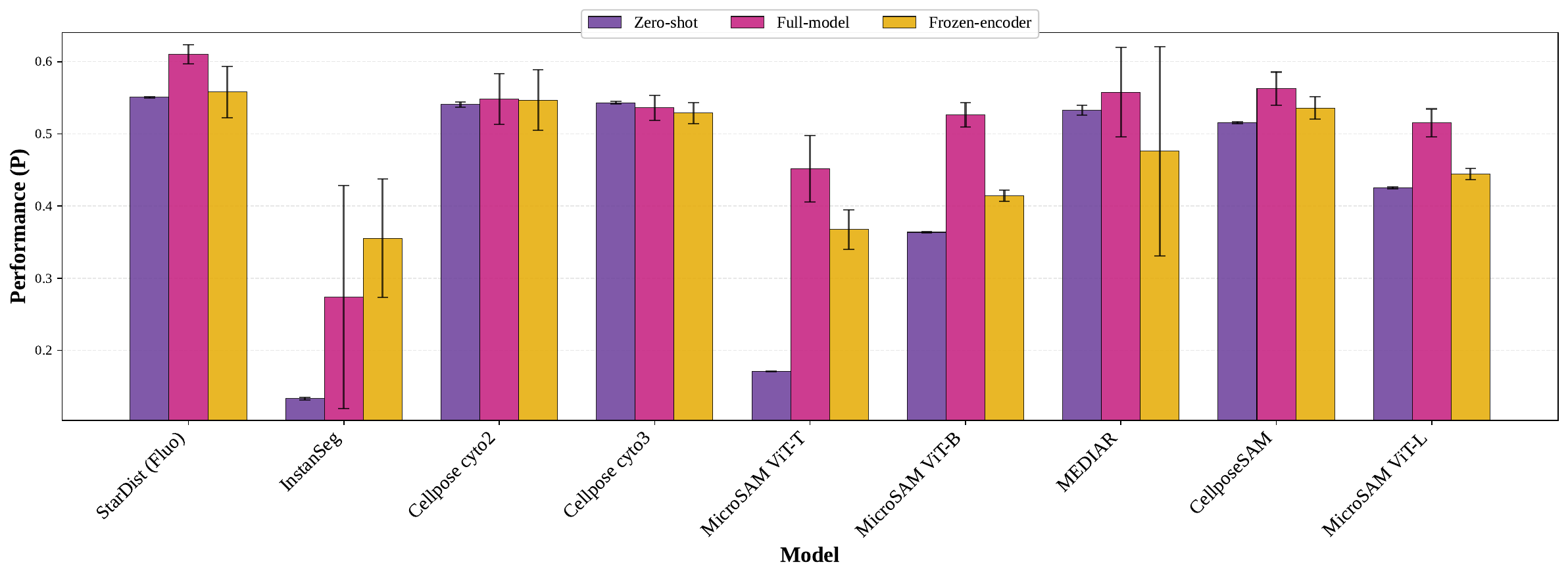}
	\caption{Segmentation performance $P$ on the mIF dataset under zero-shot inference, full-model fine-tuning and frozen-encoder fine-tuning. The plot reports the mean value of $P$ across ten runs for each compatible model with error bars indicating standard deviation.}
	\label{fig:performance_mIF}
\end{figure*}

\begin{figure*}[tp]
	\centering
	\includegraphics[width=1.0\textwidth]{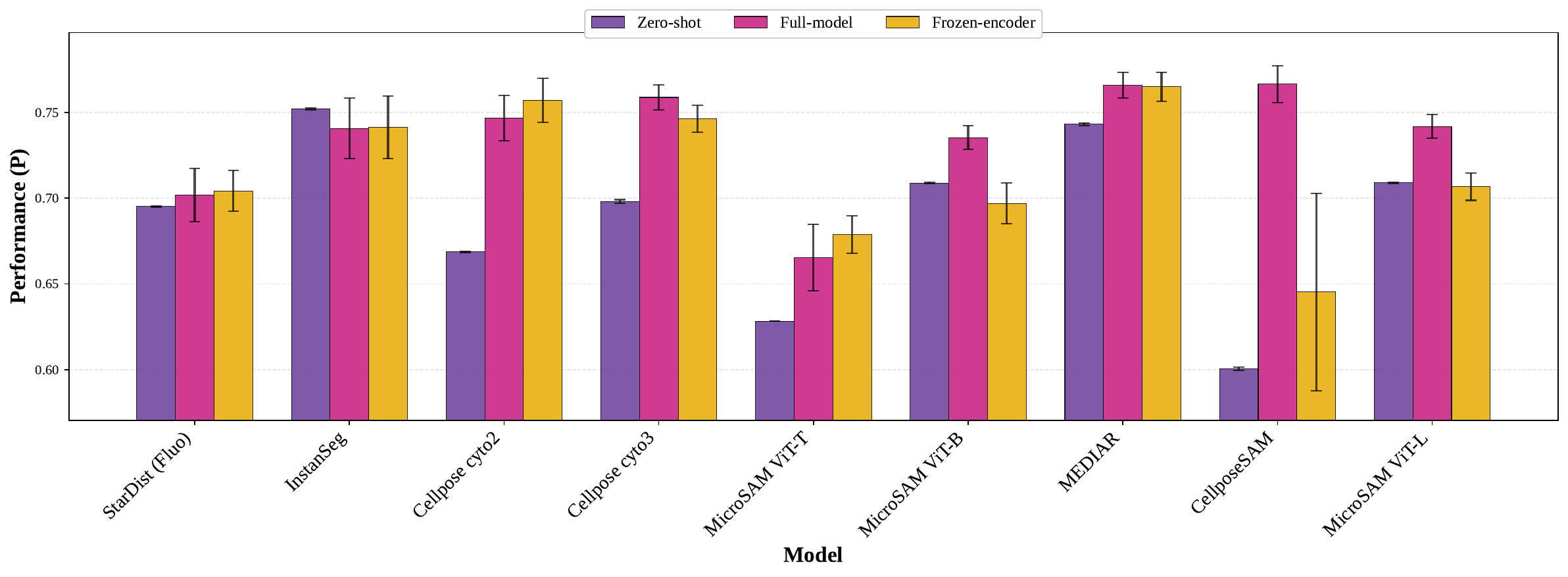}
	\caption{Segmentation performance $P$ on the 10xGenomics\_DAPI dataset under zero-shot inference, full-model fine-tuning and frozen-encoder fine-tuning. The plot reports the mean value of $P$ across ten runs for each compatible model with error bars indicating standard deviation.}
	\label{fig:performance_10xGenomics_DAPI}
\end{figure*}

\begin{figure*}[tp]
	\centering
	\includegraphics[width=1.0\textwidth]{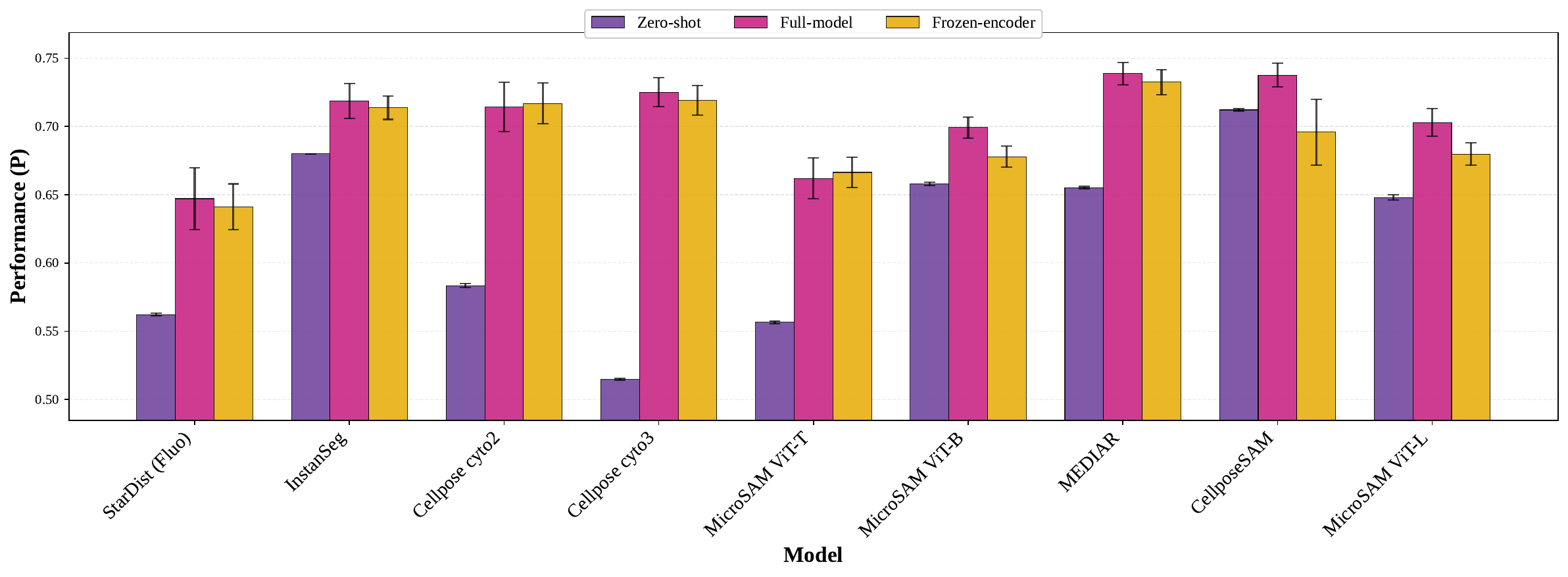}
	\caption{Segmentation performance $P$ on the DAPI dataset under zero-shot inference, full-model fine-tuning and frozen-encoder fine-tuning. The plot reports the mean value of $P$ across ten runs for each compatible model with error bars indicating standard deviation.}
	\label{fig:performance_DAPI}
\end{figure*}

\begin{figure*}[tp]
	\centering
	\includegraphics[width=1.0\textwidth]{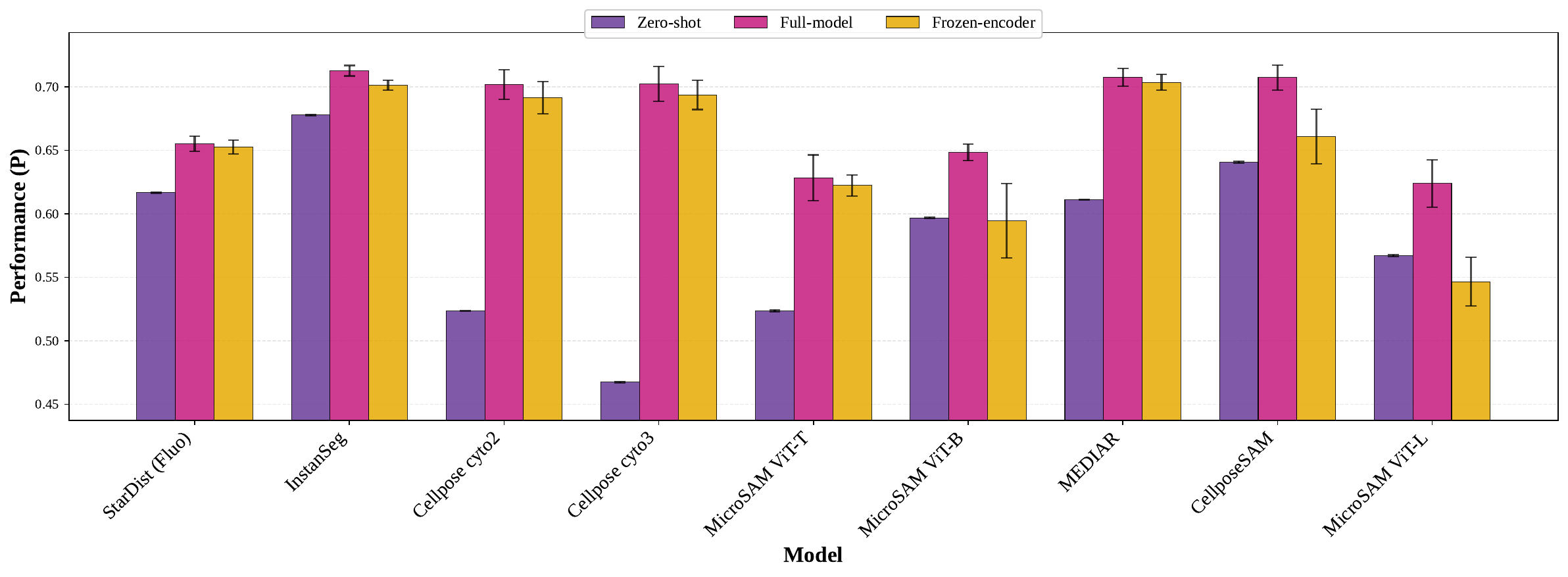}
	\caption{Segmentation performance $P$ on the ssDNA dataset under zero-shot inference, full-model fine-tuning and frozen-encoder fine-tuning. The plot reports the mean value of $P$ across ten runs for each compatible model with error bars indicating standard deviation.}
	\label{fig:performance_ssDNA}
\end{figure*}

\begin{figure*}[tp]
	\centering
	\includegraphics[width=1.0\textwidth]{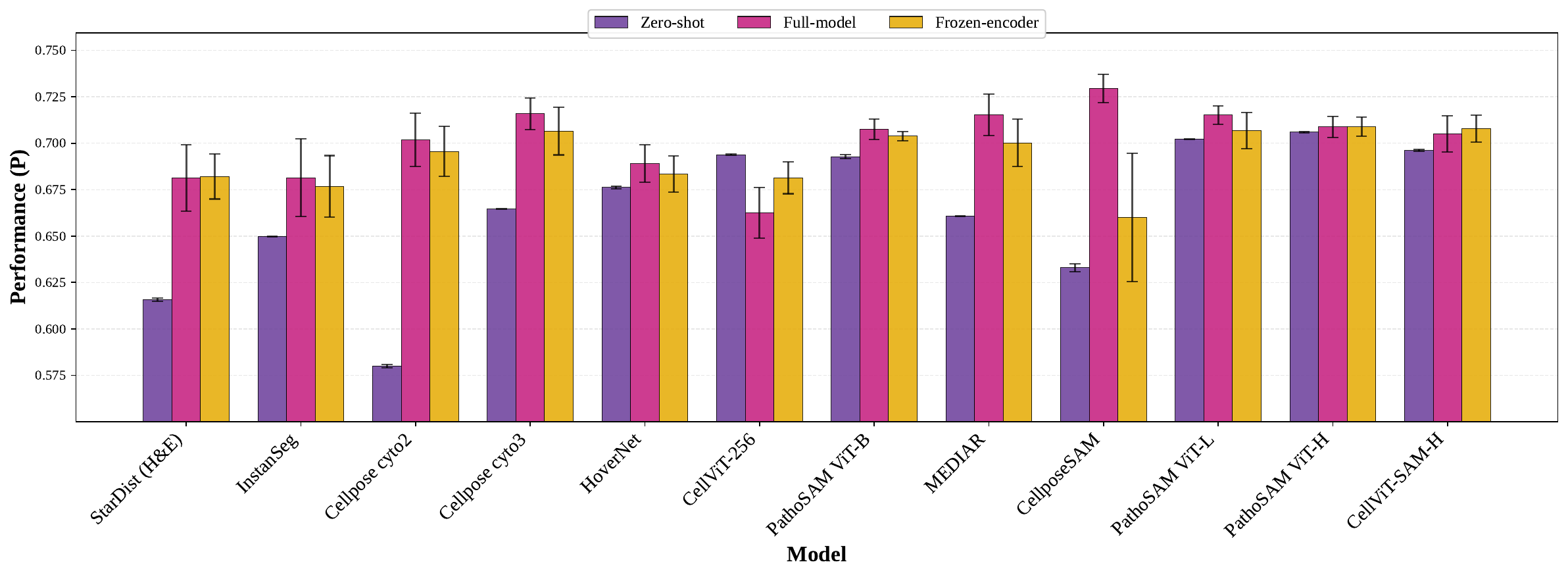}
	\caption{Segmentation performance $P$ on the 10xGenomics\_HE dataset under zero-shot inference, full-model fine-tuning and frozen-encoder fine-tuning. The plot reports the mean value of $P$ across ten runs for each compatible model with error bars indicating standard deviation.}
	\label{fig:performance_10xGenomics_HE}
\end{figure*}

\begin{figure*}[tp]
	\centering
	\includegraphics[width=1.0\textwidth]{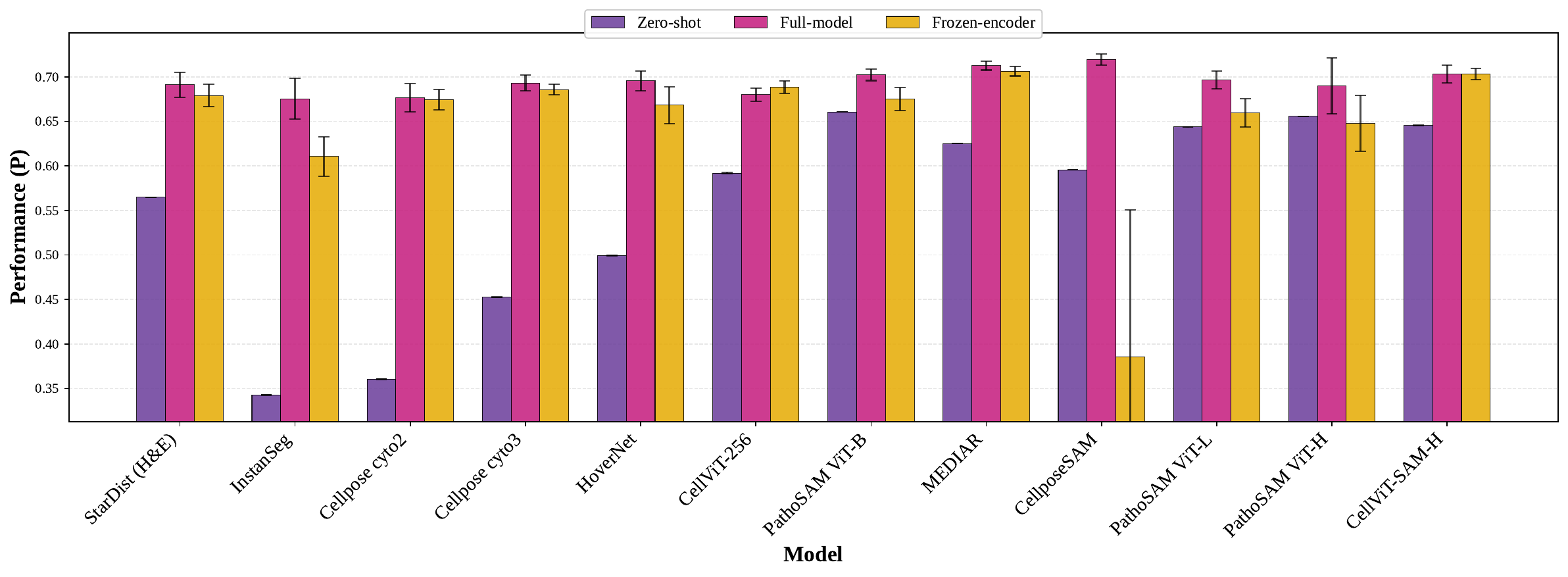}
	\caption{Segmentation performance $P$ on the HE dataset under zero-shot inference, full-model fine-tuning and frozen-encoder fine-tuning. The plot reports the mean value of $P$ across ten runs for each compatible model with error bars indicating standard deviation.}
	\label{fig:performance_HE}
\end{figure*}

Few-shot adaptation improved segmentation performance for most compatible
model-dataset combinations (Figures~\ref{fig:performance_mIF}--\ref{fig:performance_HE}). The largest gains generally occurred when the corresponding zero-shot model performed poorly. This pattern was particularly clear on HE, DAPI, and ssDNA, where adaptation substantially improved Cellpose, InstanSeg, Hover-Net, and MicroSAM models. By contrast, the gains were usually smaller on the two 10xGenomics datasets, where several models already achieved relatively strong zero-shot performance.

Full-model fine-tuning generally produced stronger performance than frozen-encoder fine-tuning. Updating the encoder allowed the model to adjust its learned representation to the target images, which was especially useful for models that transferred poorly in the zero-shot setting. However,
frozen-encoder fine-tuning sometimes remained close to full-model fine-tuning and occasionally performed slightly better. For example, freezing the encoder improved the results of CellViT-256 on both the H\&E datasets. InstanSeg on mIF shows a more pronounced exception, as the frozen-encoder strategy outperformed full-model fine-tuning considerably.

The response to adaptation also differed across datasets. Performance on mIF remained lower and more variable than on the other datasets, although StarDist (Fluo) retained strong performance and several general-purpose models improved after fine-tuning. DAPI and ssDNA showed more consistent gains, particularly for Cellpose cyto2 and Cellpose cyto3. On the H\&E datasets, adaptation substantially improved models with weak zero-shot transfer, including InstanSeg and the Cellpose cyto models. CellposeSAM and MEDIAR achieved strong adapted performance on several datasets, while the preferred model still depended on the image modality and dataset.

Adaptation did not improve every model. Cellpose cyto3 showed little benefit on mIF, while InstanSeg on 10xGenomics\_DAPI and CellViT-256 on 10xGenomics\_HE performed slightly worse than their zero-shot baselines. While the performance of most models is stable across the ten runs, CellposeSAM with frozen-encoder finetuning shows greater variance in performance on most datasets. Also, InstanSeg and MEDIAR have larger performance variations for the mIF dataset. These exceptions show that few-shot adaptation depends on both the target dataset and the pretrained model. They also show that updating all the parameters does not always provide the most stable or better result.

\subsubsection{Energy Consumption}

\begin{figure*}[tp]
	\centering
	\includegraphics[width=1.0\textwidth]{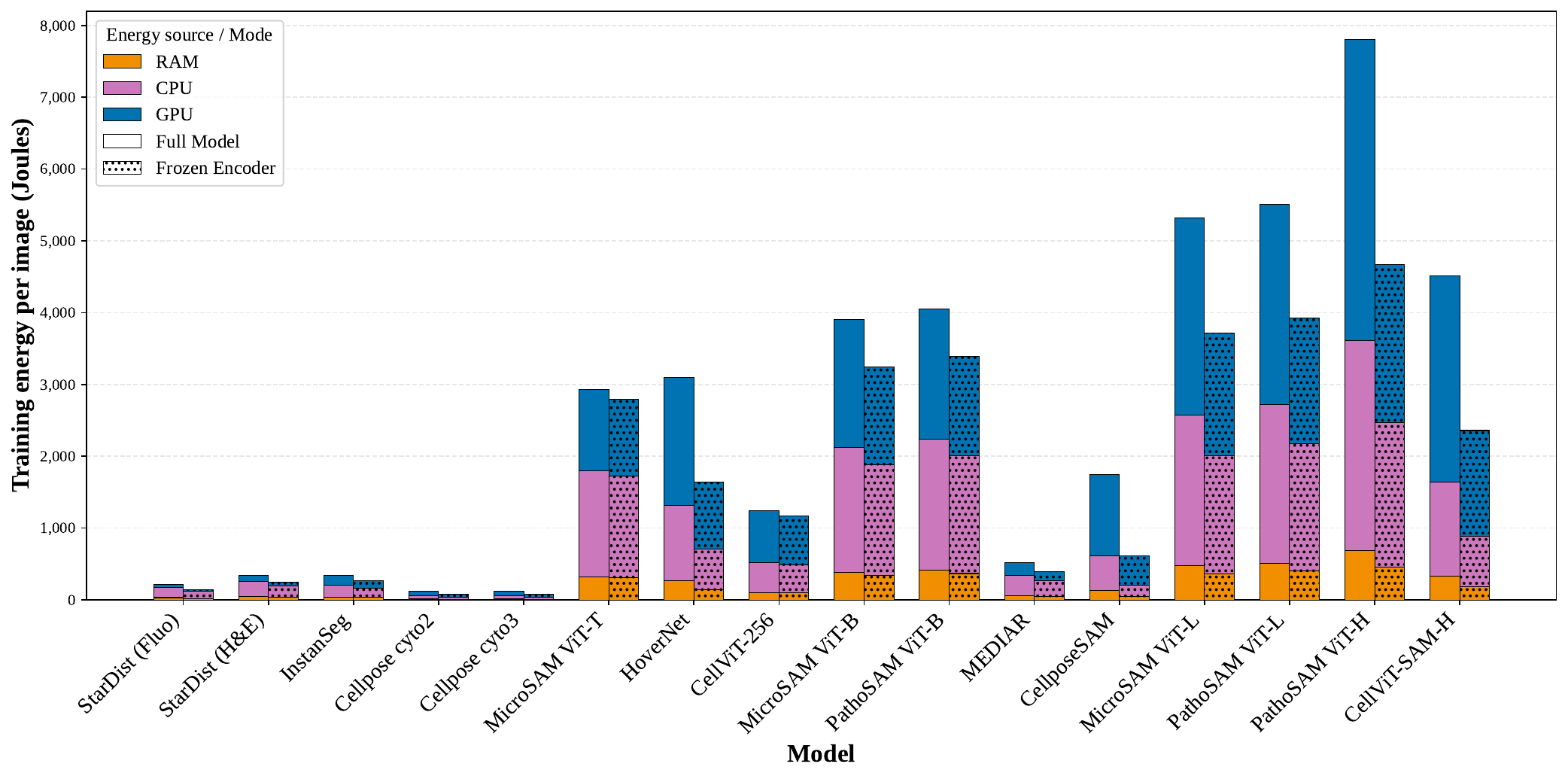}
	\caption{Mean fine-tuning energy per support image under full-model and frozen-encoder adaptation. The total bar height represents the combined GPU, CPU, and RAM energy consumption averaged across compatible datasets.}
	\label{fig:plot_stacked_finetuning_training_energy}
\end{figure*}

Fine-tuning energy \(E_{\mathrm{ft}}\) varied considerably across models (Figure~\ref{fig:plot_stacked_finetuning_training_energy}). The figure reports the energy consumed per support image, averaged across the datasets compatible with each model and across the repeated runs. Cellpose cyto2 and cyto3 required the least energy under both adaptation strategies. InstanSeg, MEDIAR, and StarDist also remained among the lower-energy models. In contrast, the larger MicroSAM, PathoSAM, and CellViT variants occupied the high-energy end of the comparison. PathoSAM ViT-H consumed the most energy during full-model fine-tuning.

Freezing the encoder during fine-tuning reduced the training energy \(E_{\mathrm{ft}}\) for all models. However, the amount of the reduction depended strongly on the model. The largest reductions appeared for CellposeSAM, CellViT-SAM-H, Hover-Net, and PathoSAM ViT-H. Freezing the encoder
also produced clear savings for the larger MicroSAM and PathoSAM variants. By contrast, the difference between the two fine-tuning strategies was small for StarDist, Cellpose, InstanSeg, MEDIAR, CellViT-256, and MicroSAM ViT-T. 

Across all the models, GPU and CPU activity accounted for the largest shares of fine-tuning energy ($E_\mathrm{ft}$), while RAM contributed relatively little. Larger models generally consumed more fine-tuning energy, but model size alone did not explain the observed differences. Models with similar parameter counts sometimes showed distinct energy profiles, and encoder freezing produced different levels of savings across architectures. Fine-tuning energy therefore reflected both model scale and the computational structure of the adaptation process.

\subsubsection{$\mathrm{\textbf{SAPI}_{\mathrm{\textbf{FT+I}}}}$}

\begin{figure*}[tp]
	\centering
	\begin{subfigure}[t]{\textwidth}
		\centering
		\includegraphics[width=0.9\textwidth]{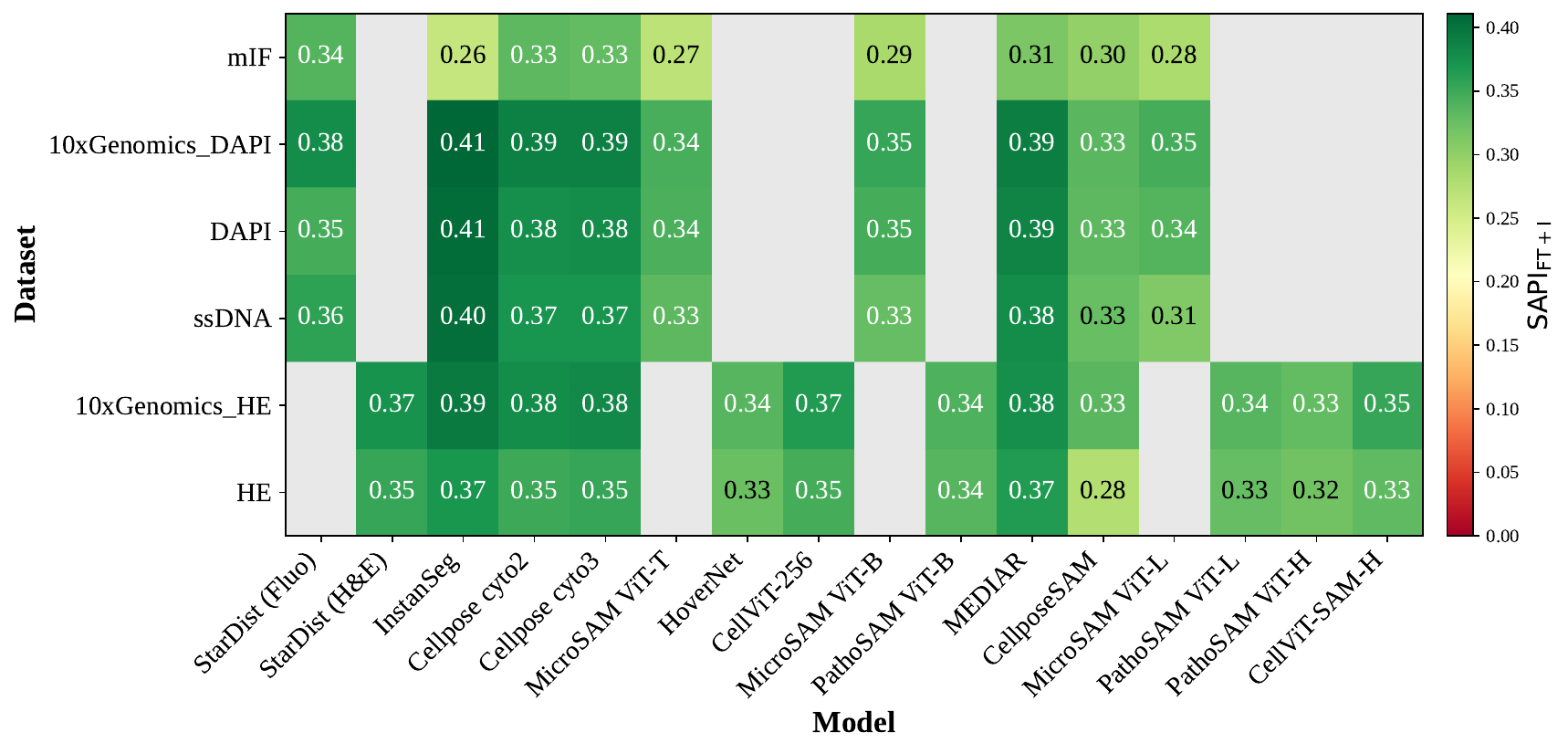}
		\caption{\(\mathrm{SAPI}_{\mathrm{FT+I}}\) for each compatible
			model-dataset combination. Grey cells indicate combinations that
			were not evaluated because the model is not intended to be used with the
			corresponding image modality.}
		\label{fig:plot_sapi_heatmap_finetuning}
	\end{subfigure}
	
	\vspace{0.8em}
	
	\begin{subfigure}[t]{\textwidth}
		\centering
		\includegraphics[width=0.9\linewidth]
		{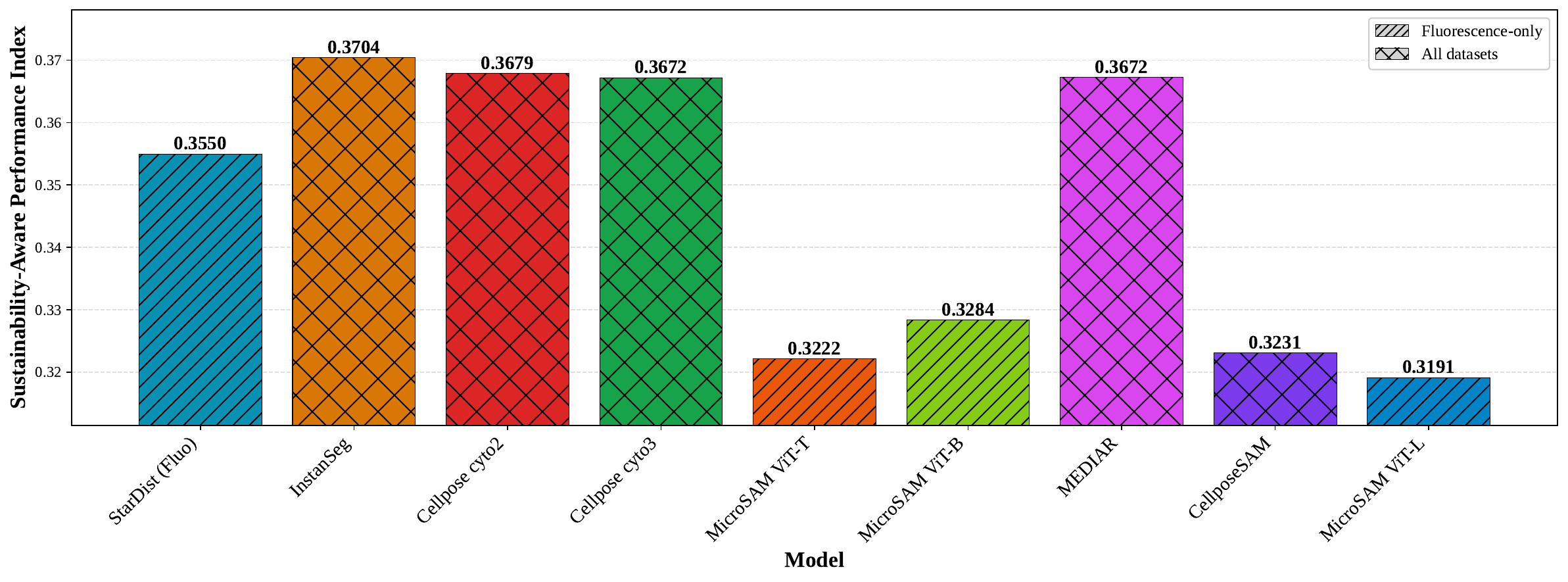}
		\caption{Mean \(\mathrm{SAPI}_{\mathrm{FT+I}}\) across the four
			fluorescence datasets: mIF, DAPI, ssDNA, and 10xGenomics\_DAPI.
			Hatching distinguishes fluorescence-specific models from models
			evaluated on all datasets.}
		\label{fig:plot_sapi_bar_finetuning_fluorescence}
	\end{subfigure}
	
	\vspace{0.8em}
	
	\begin{subfigure}[t]{\textwidth}
		\centering
		\includegraphics[width=0.9\textwidth]{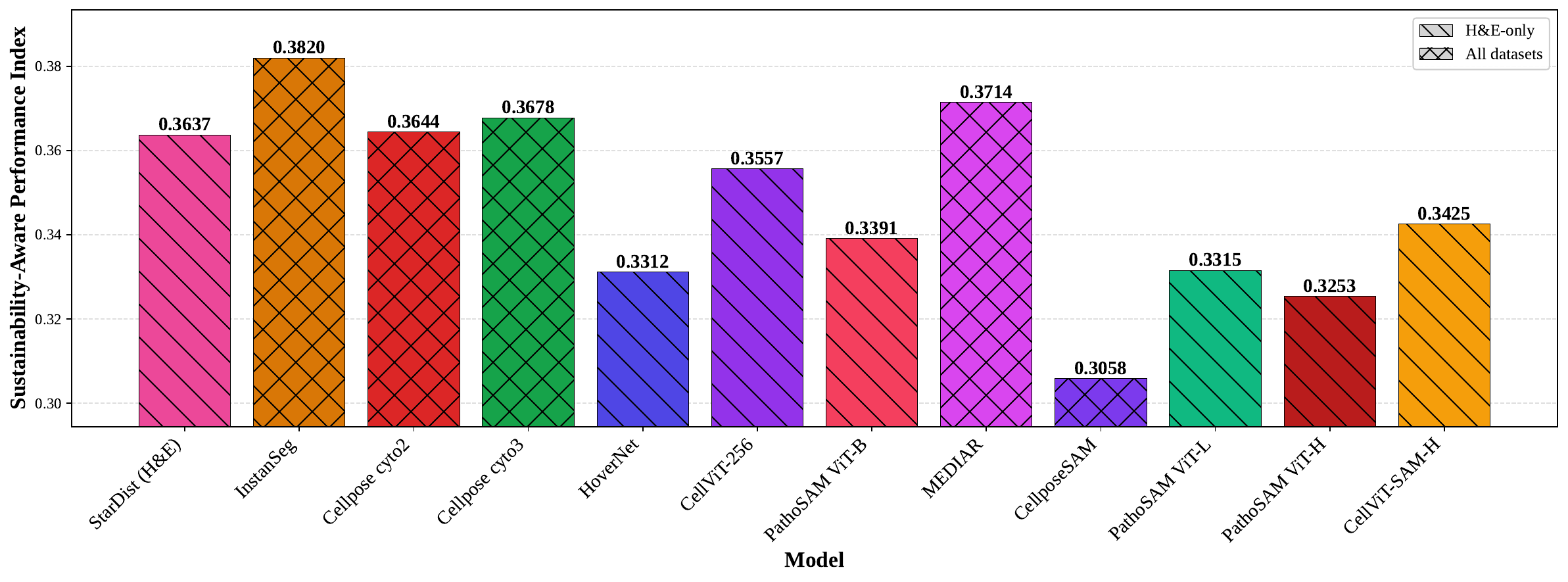}
		\caption{Mean \(\mathrm{SAPI}_{\mathrm{FT+I}}\) across the HE and
			10xGenomics\_HE datasets. Hatching distinguishes H\&E-specific
			models from models evaluated on all datasets.}
		\label{fig:plot_sapi_bar_finetuning_he}
	\end{subfigure}
	
	\caption{Sustainability-aware performance after few-shot adaptation. The results combine segmentation performance with fine-tuning and post-fine-tuning inference energy and model size. Scores are averaged across ten independent runs.}
	\label{fig:few_shot_sapi}
\end{figure*}

The \(\mathrm{SAPI}_{\mathrm{FT+I}}\) results show clear differences across models and datasets (Figure~\ref{fig:few_shot_sapi}). Smaller models that combined strong adapted performance with low fine-tuning and inference energy generally received the highest scores. As a result, the sustainability-aware ranking did not always follow the performance ranking. Several large models achieved competitive segmentation performance after adaptation, but their higher energy consumption and parameter counts reduced their \(\mathrm{SAPI}_{\mathrm{FT+I}}\) scores.

Across the fluorescence datasets, InstanSeg achieved the highest average \(\mathrm{SAPI}_{\mathrm{FT+I}}\). Cellpose cyto2, Cellpose cyto3, and MEDIAR followed closely. StarDist (Fluo) also remained competitive.  The heatmap shows that these rankings were not consistent across all fluorescence datasets. InstanSeg performed strongly on DAPI, ssDNA, and 10xGenomics\_DAPI, but obtained a much lower score on mIF. The mIF dataset also produced lower scores for several MicroSAM variants, showing that the sustainability-aware benefit of adaptation remained dataset-dependent.

A similar pattern appeared on the two H\&E datasets. InstanSeg obtained the highest average score, followed by MEDIAR. Cellpose cyto2, Cellpose cyto3, and StarDist (H\&E) also achieved strong results. In contrast, the larger PathoSAM and CellViT variants generally received lower scores, despite their competitive adapted performance because their additional energy and model-complexity costs were not fully offset by their performance gains.

Fine-tuning also changed the relative \(\mathrm{SAPI}\) ranking of several models. In the zero-shot setting, StarDist (Fluo) and MEDIAR ranked among the strongest models because they combined competitive segmentation performance with low inference energy and small model size. After few-shot adaptation, InstanSeg moved to the highest average \(\mathrm{SAPI}_{\mathrm{FT+I}}\) score across both fluorescence and H\&E datasets, while Cellpose cyto2 and cyto3 also improved their relative positions. This shift reflects the effect of adaptation on both segmentation performance and the total energy cost considered by \(\mathrm{SAPI}_{\mathrm{FT+I}}\). Models that gained substantially in segmentation performance could therefore improve their sustainability-aware ranking, whereas models with high adaptation or inference costs remained disadvantaged despite competitive adapted performance. The results show that the most favourable model under zero-shot inference is not necessarily the most favourable model after target-specific adaptation.

Overall, the results favor models that achieve a suitable balance between segmentation performance, adaptation cost, inference cost, and model complexity. No model achieved the highest score on every dataset, and some models showed markedly different results across image modalities. Nevertheless, InstanSeg remains the most sustainable model on an average across all the datasets, followed by MEDIAR and then the two Cellpose cyto variants. The choice of model should therefore reflect both the target dataset and the sustainability requirements of the intended deployment scenario.

\section{Discussion}
\label{sec:discussion}

This study examined whether the segmentation gains from large pre-trained and foundation models were proportionate to their computational requirements. We evaluated 19 model configurations across six CellBinDB datasets and considered segmentation performance, inference energy, fine-tuning energy, and model complexity. Three findings define the main narrative of the study. First, increasing model size did not lead to a consistent or proportional improvement in segmentation performance. Second, large foundation models often achieved a marginally higher performance, but they required order of magnitude more energy and trainable parameters. Third, several compact models remained competitive and provided a better balance between segmentation quality and computational cost. These results show that the model with the highest segmentation score is not necessarily the most suitable model for resource constrained deployment scenarios.

The main issue is therefore proportionality. A larger computational burden is reasonable when it produces a clear and practically important improvement. However, a marginal performance gain may not justify a large increase in inference energy, adaptation cost, or model complexity. The decision depends on the intended application. A marginal gain may be important when it prevents a consequential segmentation error, but it may have limited value when competing models already meet the required performance level. These findings support the Green AI principle that predictive quality and computational efficiency should be evaluated together \citep{schwartz2020green,henderson2020towards}. Our study extends this principle to cell and nuclear instance segmentation through a controlled comparison of predictive performance, energy use, and model complexity.

\subsection{Model scale did not determine segmentation performance}

The results do not support a simple relationship between parameter count and segmentation quality. Several large pre-trained and foundation models performed strongly, but their advantage varied across datasets and adaptation settings. Smaller architectures also remained competitive across multiple datasets. Model size alone therefore provided limited information about the performance that a model would achieve on a specific cell or nuclear segmentation task.

This result reflects the demands of instance segmentation in biomedical images. A model must separate densely packed and touching objects, preserve fine boundaries, and remain robust to changes in staining, tissue structure, image contrast, object density, and imaging modality. Generic visual representations do not automatically satisfy these requirements. Performance also depends on the correspondence between the pre-training domain and the target images, the model architecture, the instance-separation method, the inference procedure, and the model-specific pre-processing and post-processing steps.

Large-scale pre-training can still provide useful representations. Foundation models may transfer across datasets and can support tasks that are not measured in this benchmark, including prompted segmentation and reuse across several applications \citep{kirillov2023segment,ma2024segment}. Their broader capabilities may justify their size when a single shared model replaces several specialized systems or supports a wide range of workloads. However, these benefits should be demonstrated for the intended deployment setting. For a fixed and repetitive segmentation task, a specialized model may provide a better balance when the additional capabilities of a general-purpose model remain unused.

The results also show why parameter count and measured energy should not be treated as equivalent measures. Parameter count describes model scale and influences hardware requirements, but it does not fully determine computational cost. Energy use also depends on activation sizes, operator types, memory transfers, numerical precision, image tiling, prompt generation, post-processing, framework overhead, and hardware utilization. A model with fewer parameters may therefore consume more energy than its size suggests. Conversely, a larger architecture may use the hardware more efficiently. Direct energy measurements provide stronger evidence for deployment decisions than parameter count alone. Energy and complexity should therefore be reported as related but distinct properties.

Large-scale pre-training often improve cell and nuclear segmentation. However, the additional performance was usually accompanied by greater inference energy and a much larger parameter footprint. The relevant question is therefore not only whether a foundation model performs better, but whether the size of the improvement justifies the additional recurring costs.

This distinction is especially important for inference. Fine-tuning normally occurs a limited number of times, whereas inference may be repeated over thousands or millions of images. A difference that appears small at the level of one image can become substantial in a high-volume pipeline. Repeated inference affects energy consumption, processing time, accelerator availability, infrastructure requirements, cooling demand, and the number of analyses that can run concurrently. These effects are relevant to digital pathology, spatial biology, fluorescence microscopy, and high-content screening, where segmentation is often applied to large image collections.

The practical importance of the performance gain must therefore guide model selection. A more computationally demanding model may be justified when it corrects errors that affect downstream measurements, biological conclusions, or clinical interpretation. In contrast, the lower-cost model is usually more preferable when two models produce results that are sufficiently similar for the intended task. This does not mean that energy should take priority over segmentation quality. Candidate models should first meet a task-specific performance requirement. Models that satisfy this requirement can then be compared using inference energy, runtime, memory demand, parameter count, maintainability, and hardware accessibility.

This approach avoids two problematic choices. It prevents the selection of an inadequate model only because it consumes less energy. It also prevents a marginal performance gain from automatically outweighing a much larger computational burden. The appropriate weight will differ between applications and should reflect the consequences of segmentation errors in the downstream workflow.

\subsection{Few-shot adaptation should be justified by its incremental benefit}

The comparison between zero-shot inference, frozen-encoder adaptation, and full-model fine-tuning shows that adaptation should be treated as a deployment investment. Fine-tuning can improve the alignment between a pre-trained model and a target dataset. However, the gain should be evaluated against the required annotation effort, training energy, memory demand, and implementation complexity. Full-model fine-tuning is difficult to justify when it provides only a tiny improvement over zero-shot inference or frozen-encoder adaptation.

Frozen-encoder adaptation can provide a useful compromise. It updates only part of the model and can reduce the computational and memory requirements of training. It may work well when the pre-trained encoder already provides suitable representations for the target images. However, freezing the encoder can restrict adaptation when the target modality differs substantially from the pre-training domain. The relative value of frozen and full adaptation therefore depends on both the model and the dataset. Our results do not establish one adaptation strategy as universally superior.

The controlled few-shot protocol is important for this interpretation. Each model received the same tissue-stratified support set and the same training budget. This design allowed us to compare how efficiently different models used a limited support set. It did not attempt to find the maximum possible performance of every architecture. Model-specific tuning of the learning rate, augmentation, training duration, or support-set size could improve some results. However, such optimization would require additional computation and would reduce comparability between models. The benchmark instead addresses a practical question: under the same limited annotation and adaptation budget, which models convert these resources into useful performance gains most effectively?

The value of adaptation also depends on how often the resulting model will be used. An expensive fine-tuning procedure may be reasonable when it produces a meaningful improvement and the adapted model is then applied to a large number of images. The same procedure may be less attractive when models must be retrained frequently for different sites, staining protocols, tissues, or imaging devices. Adaptation and inference costs should therefore be interpreted in relation to the expected deployment volume.

The $\mathrm{SAPI}_{\mathrm{FT+I}}$ formulation makes this distinction explicit by combining fine-tuning and post-fine-tuning inference energy. The parameter $\lambda$ controls their relative contribution. A larger value gives more importance to recurring inference, while a smaller value gives greater importance to adaptation. This weighting represents an assumed deployment pattern rather than a complete lifecycle model. It should therefore be selected and reported according to the expected use of the adapted model.

\subsection{SAPI makes the performance-resource trade-off explicit}
No single measurement was sufficient to identify the most appropriate model. A ranking based only on segmentation performance favors the most accurate models without considering the resources they require. A ranking based only on energy or parameter count favors compact models even when their segmentation quality is insufficient. The joint analysis of performance, energy, and model size showed that models occupied different positions across these dimensions.

The Sustainability-Aware Performance Index (SAPI) addresses this problem by combining the three dimensions in a transparent manner. Under the default weights, ($\alpha$=1/2, $\beta$=1/3, and $\gamma$=1/6), performance, energy, and complexity have a relative importance of (3:2:1). Performance receives the greatest weight because segmentation quality remains the primary requirement. Energy receives the second-largest weight because inference can create a recurring operational cost. Parameter count receives the smallest weight but still captures relevant constraints related to storage, memory, model transfer, and access to suitable hardware.

This weighting does not reward low energy use at the expense of inadequate segmentation. Instead, it assesses whether the achieved performance is proportionate to the required energy and model complexity. A compact model can rank favorably when it achieves competitive performance at substantially lower cost. A larger model can also rank highly when its performance advantage is sufficiently large to compensate for its greater resource consumption.

SAPI does not define a universal and permanent ordering of models. Its ranking depends on the priorities expressed through the weights. A high-throughput laboratory may assign more importance to inference energy. An application with severe consequences for segmentation errors may place greater emphasis on performance. A resource-constrained setting may give additional importance to model complexity. A change in ranking under different weights is therefore not a failure of the SAPI index. Instead, it shows that the preferred model depends on the intended use case.

\subsection{Energy consumption and implications for operational carbon emissions}

Energy consumption has direct implications for the operational carbon emissions associated with computation. Electricity-related operational emissions depend on the energy consumed and the carbon intensity of the electricity supplying the computation. Under the same electricity and data-center conditions, a proportional increase in energy consumption therefore produces a proportional increase in operational emissions. This makes energy efficiency an important consideration when selecting models for deployment.

The implications become more substantial as computational workloads increase. A small difference in energy per image can accumulate when a model processes thousands or millions of images or is deployed continuously across multiple sites. This is particularly relevant for high-throughput applications such as digital pathology, biopharmaceutical screening, and large-scale microscopy analysis. When models provide only marginal performance gains but require substantially more computation, the additional energy demand can translate into a corresponding increase in electricity related operational emissions. Fine-tuning and repeated model adaptation add further computational demand and should therefore also be considered when evaluating the overall resource requirements of a model.

Carbon-aware practices can reduce the emissions associated with a given computational workload by using electricity with lower carbon intensity or shifting computation to periods when cleaner electricity is available. These measures can complement improvements in model and workflow efficiency, which directly reduce the amount of computation required. The energy results reported in this study should, however, not be interpreted as a complete environmental assessment. They quantify computational demand within the defined experimental setting and do not capture embodied emissions from hardware and infrastructure or the energy and emissions associated with pre-training externally developed models. Our results, therefore, provide an important indicator of the potential operational carbon implications of model choice.

\subsection{Implications for sustainable biomedical image segmentation}
\label{sec:deployment_implications}

Our findings have direct implications for the design and deployment of biomedical image-segmentation systems. Cell and nuclear segmentation often forms an early stage of a larger analytical pipeline. Its outputs may be used for feature extraction, cell counting, spatial analysis, population statistics, phenotypic profiling, treatment-response modeling, or quality control. The value of a performance improvement should therefore be judged by its effect on these downstream analyses rather than by the segmentation score alone.

This consideration is particularly important in high-content screening, drug discovery, toxicology, spatial biology, and large microscopy studies. These applications may process large numbers of images across many experimental conditions. A small increase in per-image energy or runtime can accumulate over the full workload. A more expensive model may also reduce throughput or require additional accelerators. The resulting infrastructure burden may limit the number of samples, compounds, tissues, or experimental conditions that can be analyzed within a fixed budget.

Our results indicate that researchers should not select the largest available model by default. They should first define the minimum segmentation quality required by the downstream task. Among models that meet this requirement, they should prefer the model that provides the most favorable balance of inference energy, adaptation energy, runtime, memory use, parameter count, and maintenance effort. This approach makes computational cost part of the model-selection process.

Model developers should also report efficiency measurements alongside segmentation results. Parameter count alone is not sufficient. Studies should report measured inference energy, runtime, memory use, hardware, numerical precision, image size, tiling strategy, and relevant pre-processing and post-processing boundaries. Fine-tuning studies should additionally report the training budget and the energy required for adaptation. These details are necessary for making informed decisions about using the model.

Adaptation strategies should follow the same principle. Full-model fine-tuning should not be treated as the automatic choice. Frozen-encoder adaptation may be preferable when it achieves a similar improvement at lower energy cost. Zero-shot inference may remain the most suitable option when adaptation provides marginal benefit. The selected strategy should depend on the incremental performance gain, the frequency of retraining, and the expected number of future inferences.

Our findings are also relevant to shared research infrastructures and institutions with limited access to high-memory accelerators. A compact model that runs on widely available hardware may support broader reproduction and adoption, even when its performance is marginally below that of the largest model. Computational accessibility affects which laboratories can use a model, the size of the studies they can conduct, and their ability to reproduce published results. Efficiency is therefore connected not only to environmental sustainability but also to scientific accessibility.

A foundation model may still provide the best system-level solution when it supports several modalities or tasks and removes the need to maintain separate specialized models. In that case, the relevant comparison should include the full anticipated workload. The cost of one shared model should be compared with the combined cost and maintenance burden of the systems it replaces. Our benchmark does not directly measure this multi-task value, but it shows that broad capability should not be assumed to compensate automatically for high recurring cost.

For clinical and translational applications, computational efficiency can influence processing time, local hardware requirements, service capacity, and maintainability. However, this study did not evaluate clinical decision-making, patient outcomes, regulatory requirements, or prospective workflow performance. Our results concern computational and operational suitability. Any clinical application would still require external validation, detailed failure analysis, robustness testing, and prospective evaluation.

The broader implication is that sustainable model selection does not mean selecting the model with the lowest energy consumption. It means using computational resources in proportion to the value they provide. The most suitable model is one that meets the required performance standard without imposing an unnecessary computational burden and adaptation cost.

\subsection{Limitations}
Several limitations affect the interpretation of our results. First, we performed the measurements on one fixed GPU and CPU configuration. This choice controlled hardware variation and supported consistent comparisons. However, absolute energy consumption may differ across accelerators, numerical precisions, software environments, and deployment platforms. Relative efficiency may also change when an architecture is better optimized for a different hardware system. The reported energy values are therefore empirical measurements for the evaluated configuration rather than fixed properties of the model architectures.

Second, the benchmark covers operational energy within defined inference and fine-tuning boundaries. It does not include the energy consumed during the original pre-training of the published checkpoints. Reliable checkpoint-specific pre-training information was not available for all models. 

Third, the fixed adaptation protocol was designed to enable controlled comparison across models rather than maximize the performance of each model individually. The use of limited support sets, a fixed training duration, and no validation-based early stopping ensures that all models are evaluated under the same adaptation conditions, but may favor models that adapt efficiently within these constraints. Model-specific optimization, such as using additional support images, longer training, different learning-rate schedules, or model-specific augmentation, could improve the performance of individual models. However, such optimization would introduce additional sources of variation, increase computational cost, and make direct comparisons across models less controlled.

Fourth, the six CellBinDB subsets cover several imaging modalities and tissue strata, but they do not represent the full diversity of microscopy and pathology data. Differences in specimen preparation, staining, imaging devices, acquisition settings, artifacts, and disease composition may affect model performance. External evaluation on independently collected datasets is needed before extending the reported rankings to other settings. Undocumented overlap between model pre-training data and the source data underlying a benchmark may also be difficult to exclude for externally developed checkpoints.

Fifth, for the zero-shot inference experiments, we evaluated model-native automatic inference without dataset-specific threshold optimization or manual prompting. This protocol supports scalable and consistent comparison, but it may not show the maximum performance of interactive or prompt-dependent models. The conclusions therefore apply to automatic segmentation under the evaluated settings. They should not be directly extended to interactive workflows in which expert prompts or manual corrections form part of the inference process.

Sixth, parameter count represents only one aspect of model complexity. It does not directly describe activation memory, operator efficiency, dependency management, loading time, software stability, or maintenance effort. The measured energy and memory values provide complementary information, but a complete deployment analysis would also require throughput testing, reliability assessment, hardware cost, and integration with the intended analytical system.

Finally, SAPI contains explicit choices. The weights assigned to performance, energy, and parameter count, and the value of $\lambda$ affect the resulting ranking. The index is configurable so that these choices can reflect different deployment priorities. However, every application should state and justify its selected weights. SAPI rankings should always be reported together with its component weights.

\section{Conclusion}
\label{sec:conclusion}

This study presents a sustainability-aware benchmark of 19 pretrained cell and nuclear instance segmentation models across six CellBinDB datasets. We evaluated zero-shot inference and few-shot adaptation using segmentation performance, operational energy consumption, and model size. Our results show that these factors vary substantially across models and datasets, and that higher performance does not always justify greater computational cost. Model selection should therefore consider image modality, adaptation needs, deployment scale, and resource use rather than segmentation accuracy alone.

We introduced SAPI, a novel measure, to provide a transparent way to examine the trade-offs between performance, energy use, and model complexity. The explicit weightage assigned to each component of SAPI allows researchers and practitioners to adjust the weighting according to their priorities, providing a practical framework for selecting cell segmentation models that are both effective and efficient.

\section*{CRediT authorship contribution statement}
\textbf{Eiram Mahera Sheikh:} Conceptualization, Methodology, Software, Formal analysis, Investigation, Visualization,
Writing - Original Draft. \textbf{Alaa Tharwat:} Writing - Review \& Editing. \textbf{Wolfram Schenck:} Supervision, Funding acquisition, Writing - Review \& Editing

\section*{Declaration of competing interest}

The authors declare that they have no known competing financial interests or personal relationships that could have appeared to influence the work reported in this paper.

\section*{Acknowledgments}
This research was conducted within the framework of the project “SAIL: SustAInable Lifecycle of Intelligent SocioTechnical Systems” (grant no. NW21-059B). SAIL is receiving funding from the programme “Netzwerke 2021”, an initiative of the Ministry of Culture and Science of the State of North Rhine-Westphalia, Germany. The sole responsibility for the content of this publication lies with the authors. We thank Christoph Ostrau for reviewing the manuscript and providing valuable comments and suggestions that contributed to its improvement.

\section*{Data availability}

The code developed for this study is publicly available at \url{https://github.com/eiram-mahera/sapi}.
The CellBinDB datasets analyzed in this study are publicly available through Zenodo at \url{https://doi.org/10.5281/zenodo.15370205}. 

\section*{Supplementary data}
Supplementary material associated with this article is provided in ~\ref{app:supp}.

\bibliographystyle{unsrtnat}
\bibliography{main}

\clearpage
\onecolumn
\appendix
\renewcommand{\thesection}{Appendix \Alph{section}}
\renewcommand{\thesubsection}{\Alph{section}.\arabic{subsection}}

\setcounter{figure}{0}
\setcounter{table}{0}
\renewcommand{\thefigure}{S\arabic{figure}}
\renewcommand{\thetable}{S\arabic{table}}
\def\theHfigure{S\arabic{figure}}
\def\theHtable{S\arabic{table}}

\section{Supplementary Material}
\label{app:supp}

	This supplementary material provides additional information on model and software, together with detailed quantitative and qualitative results. The experimental protocol, metric definitions, energy measurement procedure, carbon-emission calculation, and formulation of the Sustainable AI Performance Index are described in the main manuscript. Unless stated otherwise, quantitative results report the mean across ten runs. Qualitative zero-shot examples use the experiment associated with random seed 0.
	
\subsection{Model and Software Details}
\label{subsec:model-details}
	
	Table~\ref{tab:model-checkpoints} lists every evaluated model in the benchmark, the exact pre-trained checkpoint used to initialize it, and the software version used to run the experiments. For software installed directly from a source repository, the table reports the corresponding Git commit instead of a package version. The exact inference and training configurations and the code developed for this study are publicly available at \url{https://github.com/eiram-mahera/sapi}.
	
\subsection{Zero-Shot Evaluation}
\label{subsec:zero-shot}
	
\subsubsection{Performance Metrics}
\label{subsubsec:performance_metrics}
	
\begin{figure*}[tp]
		\centering
		\includegraphics[width=1.0\textwidth]{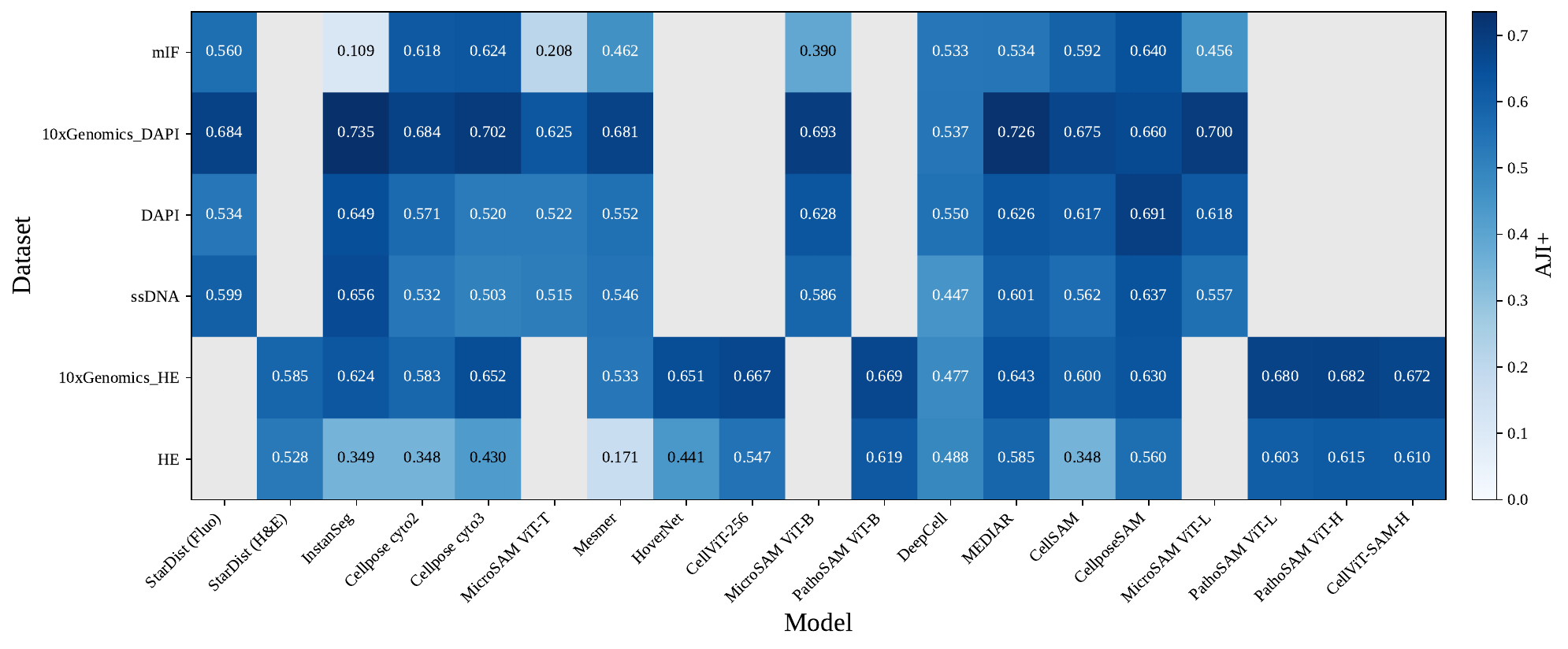}
		\caption{Zero-shot AJI+ scores across compatible model--dataset combinations. Grey cells denote model--dataset combinations that were not evaluated because the model is not intended to be used with that image modality.}
		\label{fig:zero_shot_aji}
	\end{figure*}

\begin{figure*}[tp]
		\centering
		\includegraphics[width=1.0\textwidth]{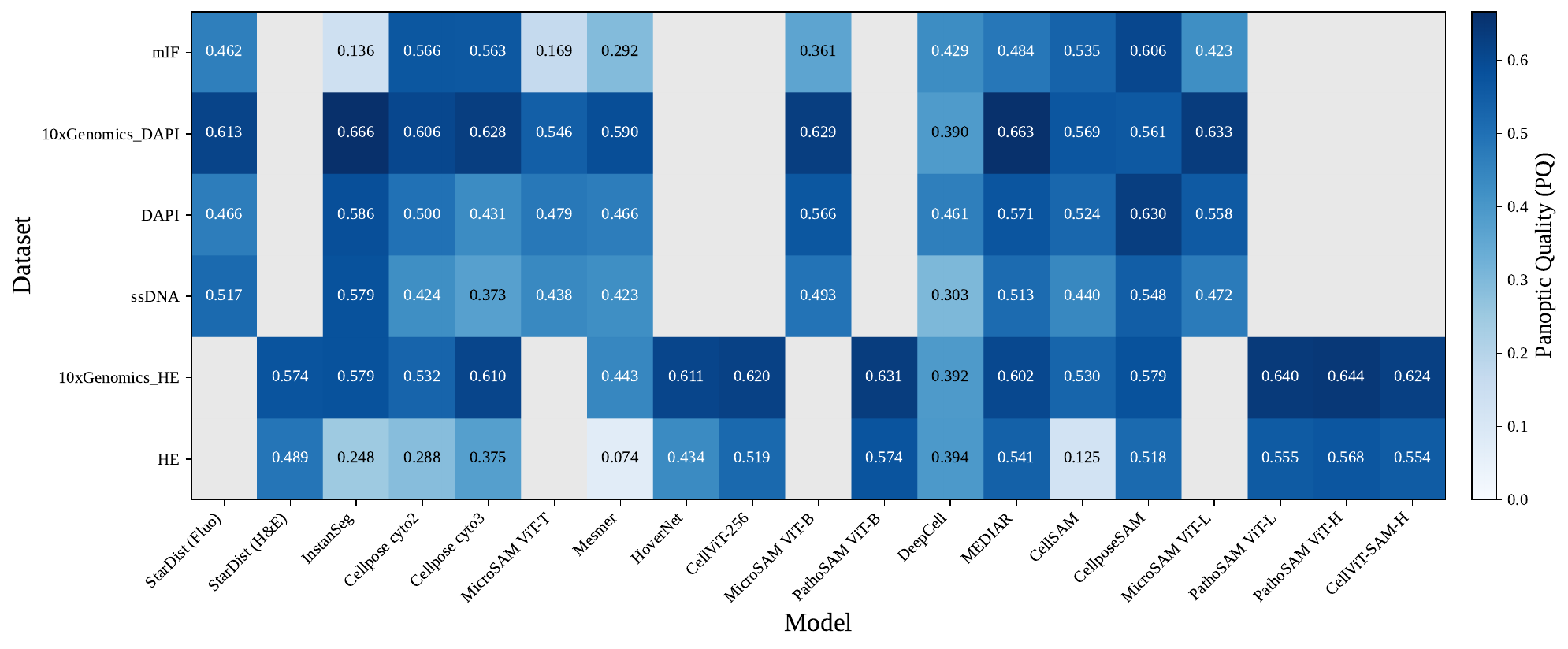}
		\caption{Zero-shot Panoptic Quality (PQ) scores across compatible model--dataset combinations. Grey cells denote model--dataset combinations that were not evaluated because the model is not intended to be used with that image modality.}
		\label{fig:zero_shot_pq}
	\end{figure*}

\begin{figure*}[tp]
		\centering
		\includegraphics[width=1.0\textwidth]{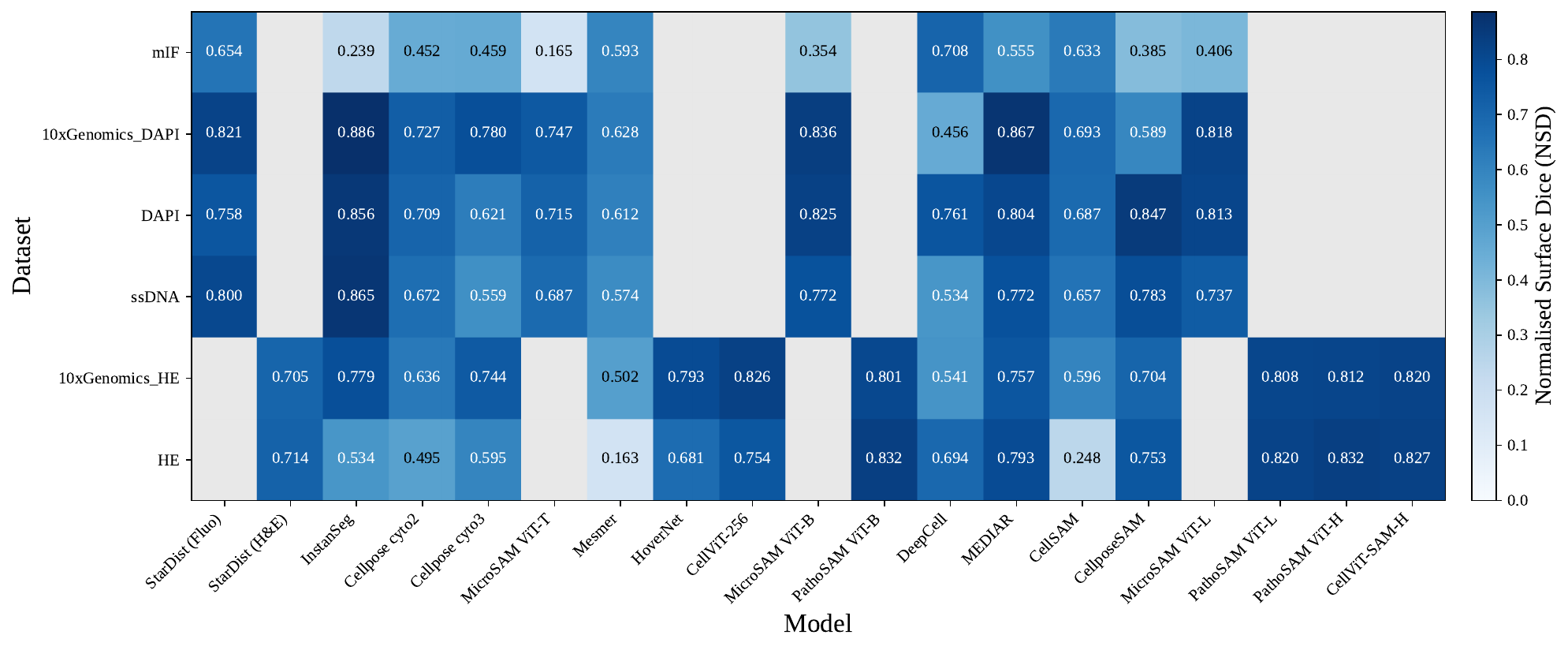}
		\caption{Zero-shot Normalized Surface Dice (NSD) scores across compatible model--dataset combinations. Grey cells denote model--dataset combinations that were not evaluated because the model is not intended to be used with that image modality.}
		\label{fig:zero_shot_nsd}
	\end{figure*}

	We evaluated each pre-trained model without dataset-specific adaptation. Each model was assessed only on datasets compatible with its intended image modality. The main manuscript reports the composite performance score ($P$), whereas Figures~\ref{fig:zero_shot_aji}, \ref{fig:zero_shot_pq} and \ref{fig:zero_shot_nsd} provide the corresponding AJI+, Panoptic Quality (PQ), and Normalized Surface Dice (NSD) results.
	
\subsubsection{Qualitative Predictions}
\label{subsubsec:qualitative_predictions}
	
\begin{figure*}[tp]
		\centering
		\includegraphics[width=1\textwidth]{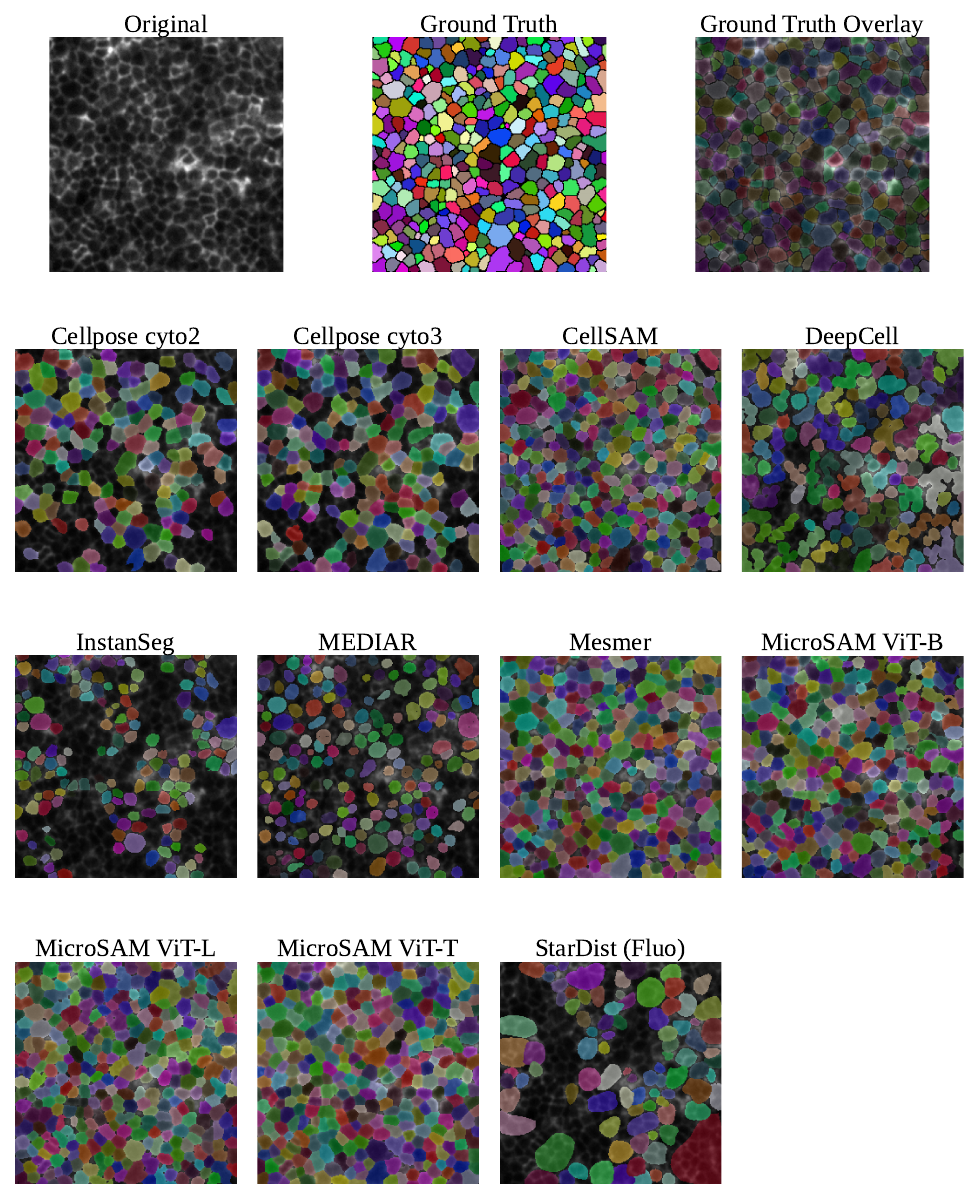}
		\caption{Qualitative zero-shot comparison for one image sampled from the mIF dataset using random seed 0. The figure shows the input image, the ground-truth instance mask, the ground-truth overlay, and predictions from all models compatible with the image modality. Each prediction is overlaid on the input image.}
		\label{fig:dataset_grid_mIF}
	\end{figure*}

\begin{figure*}[tp]
		\centering
		\includegraphics[width=1\textwidth]{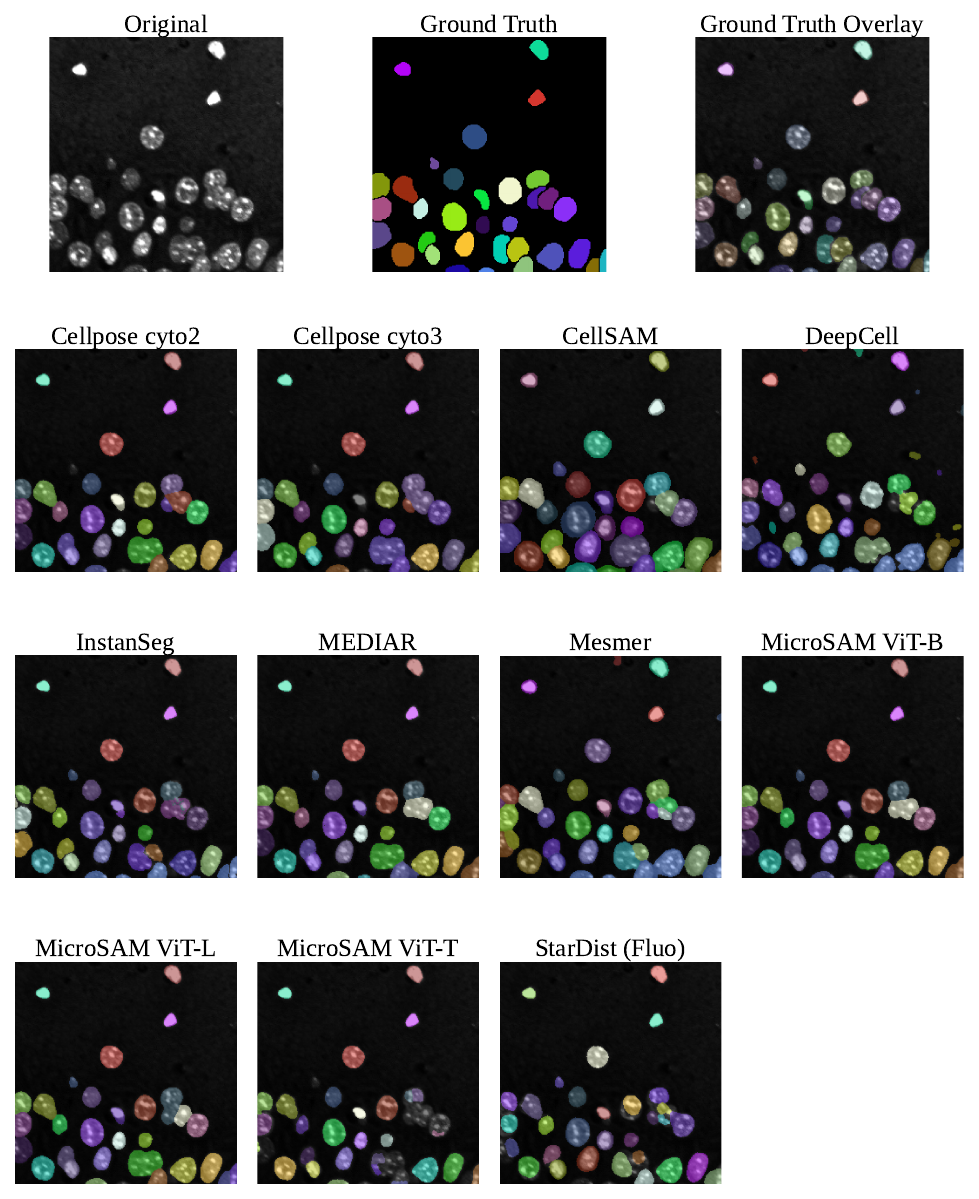}
		\caption{Qualitative zero-shot comparison for one image sampled from the DAPI dataset using random seed 0. The figure shows the input image, the ground-truth instance mask, the ground-truth overlay, and predictions from all models compatible with the image modality. Each prediction is overlaid on the input image.}
		\label{fig:dataset_grid_DAPI}
	\end{figure*}

\begin{figure*}[tp]
		\centering
		\includegraphics[width=1\textwidth]{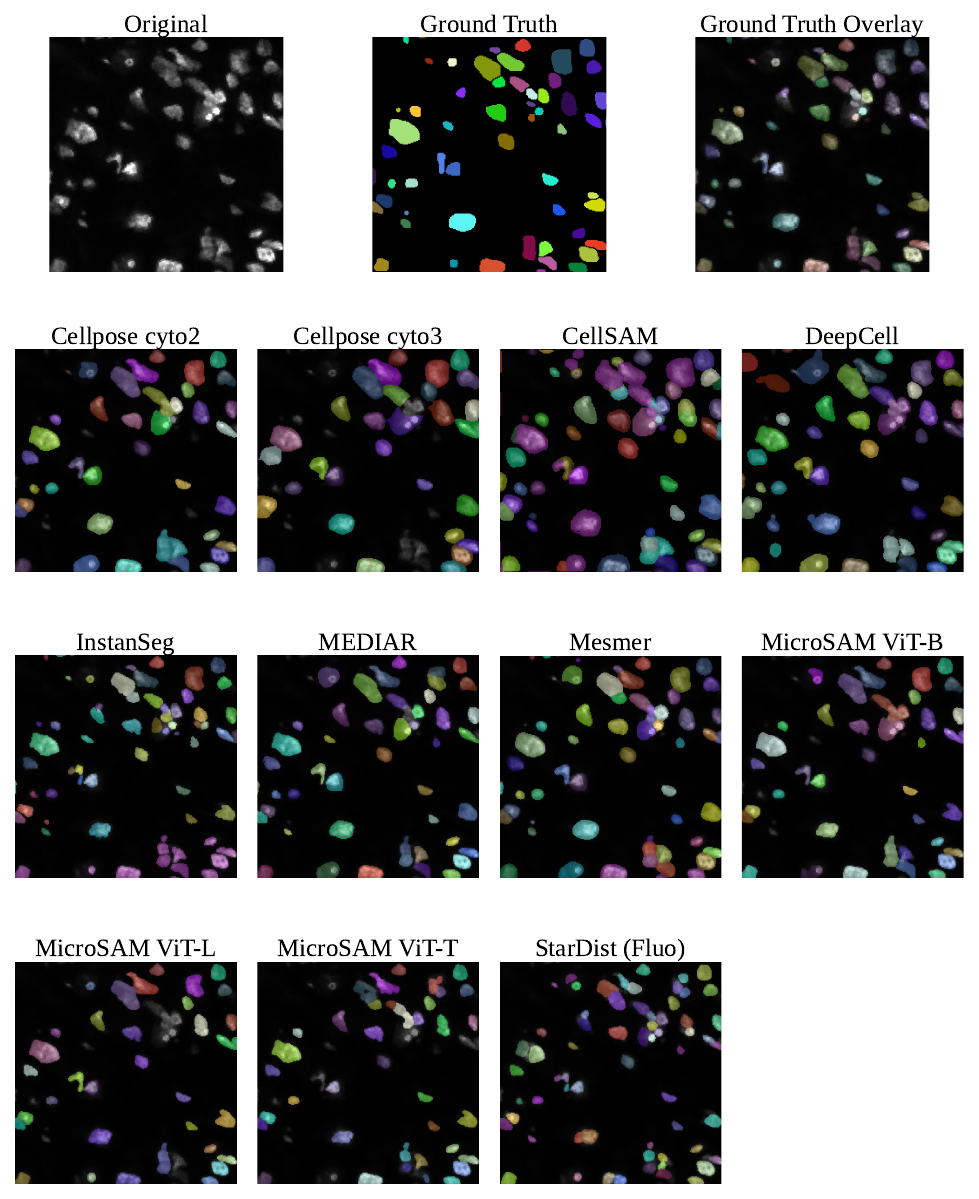}
		\caption{Qualitative zero-shot comparison for one image sampled from the ssDNA dataset using random seed 0. The figure shows the input image, the ground-truth instance mask, the ground-truth overlay, and predictions from all models compatible with the image modality. Each prediction is overlaid on the input image.}
		\label{fig:dataset_grid_ssDNA}
	\end{figure*}

\begin{figure*}[tp]
		\centering
		\includegraphics[width=1\textwidth]{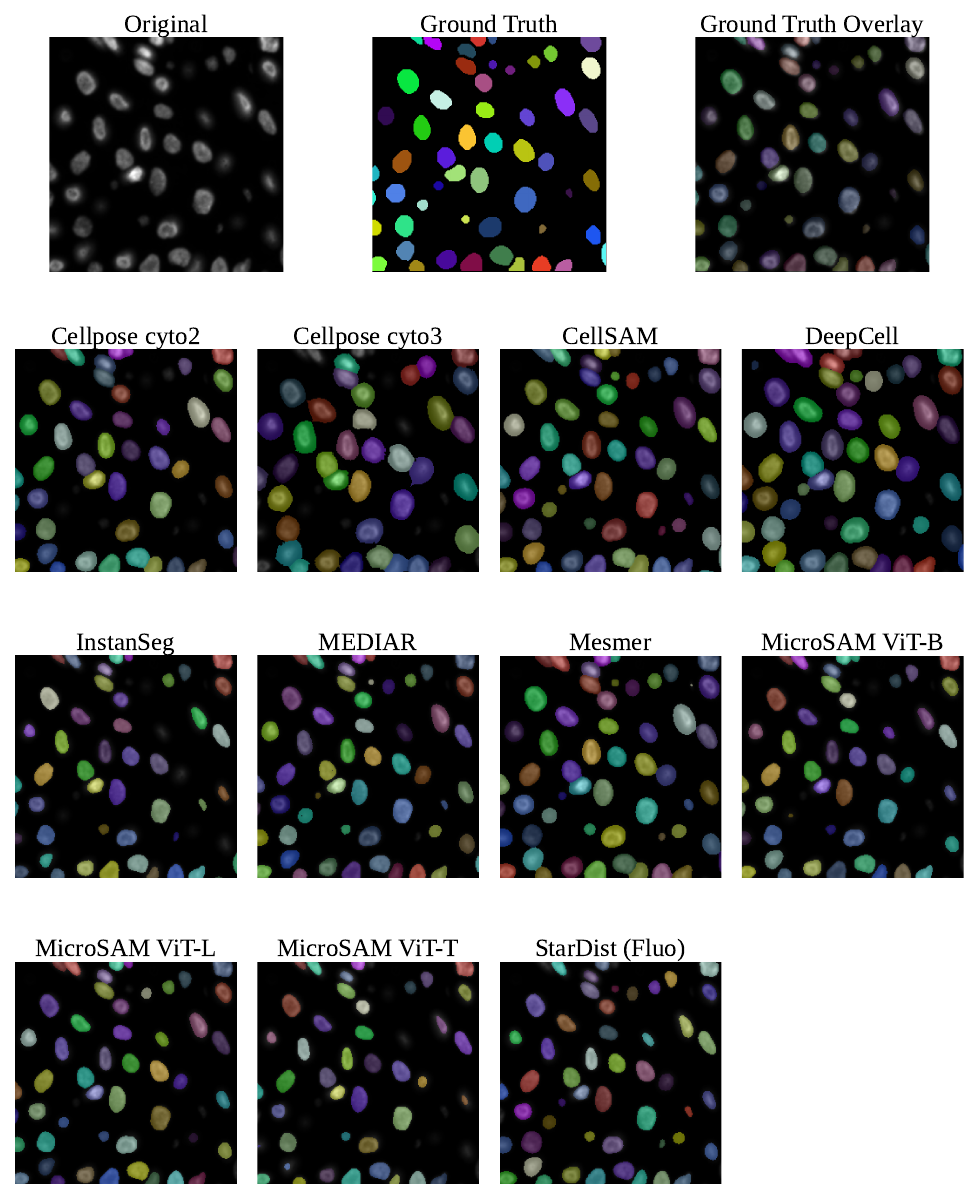}
		\caption{Qualitative zero-shot comparison for one image sampled from the 10xGenomics\_DAPI dataset using random seed 0. The figure shows the input image, the ground-truth instance mask, the ground-truth overlay, and predictions from all models compatible with the image modality. Each prediction is overlaid on the input image.}
		\label{fig:dataset_grid_10xGenomics_DAPI}
	\end{figure*}

\begin{figure*}[tp]
		\centering
		\includegraphics[width=0.8\textwidth]{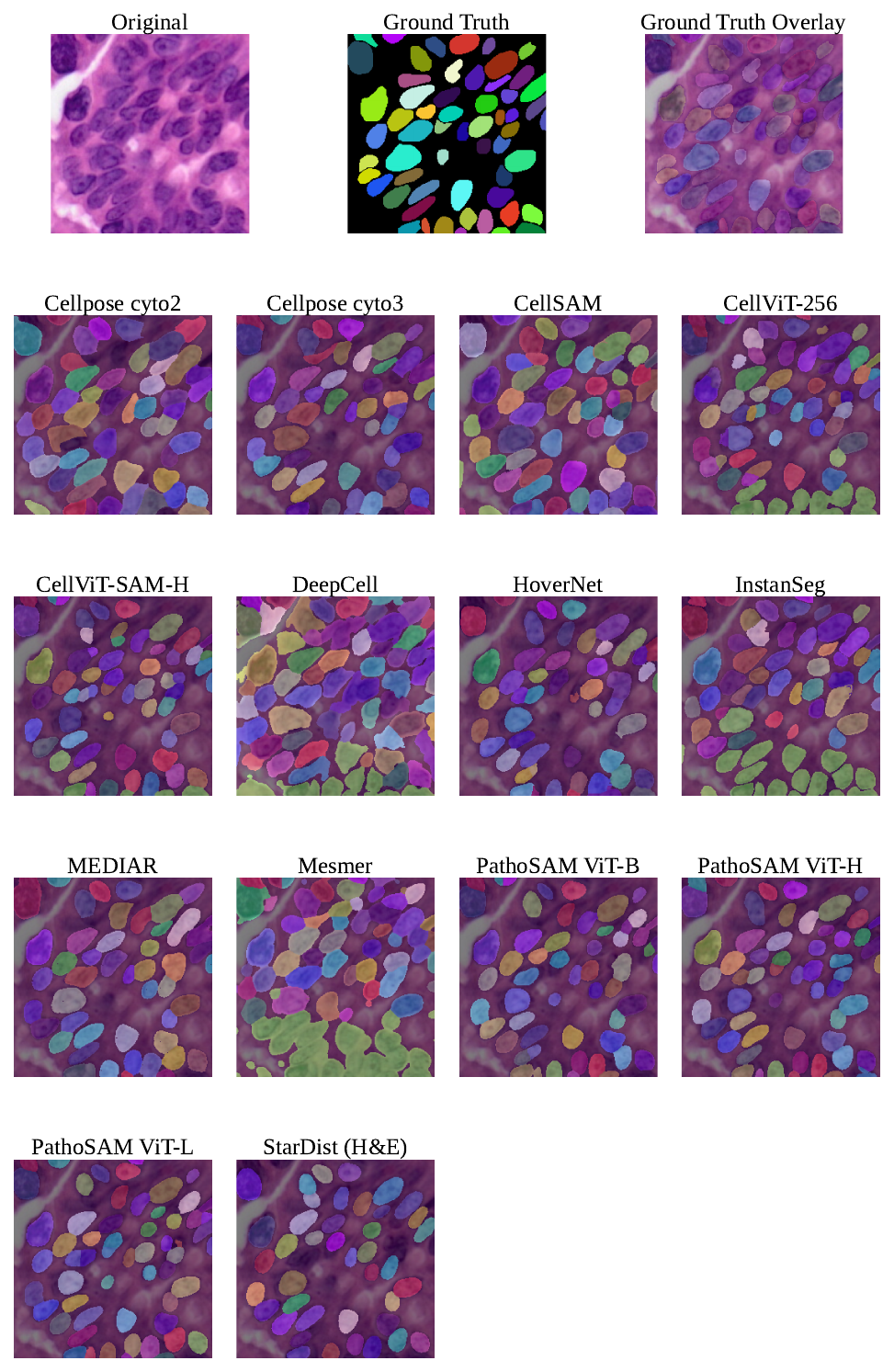}
		\caption{Qualitative zero-shot comparison for one image sampled from the 10xGenomics\_HE dataset using random seed 0. The figure shows the input image, the ground-truth instance mask, the ground-truth overlay, and predictions from all models compatible with the image modality. Each prediction is overlaid on the input image.}
		\label{fig:dataset_grid_10xGenomics_HE}
	\end{figure*}

\begin{figure*}[tp]
		\centering
		\includegraphics[width=0.8\textwidth]{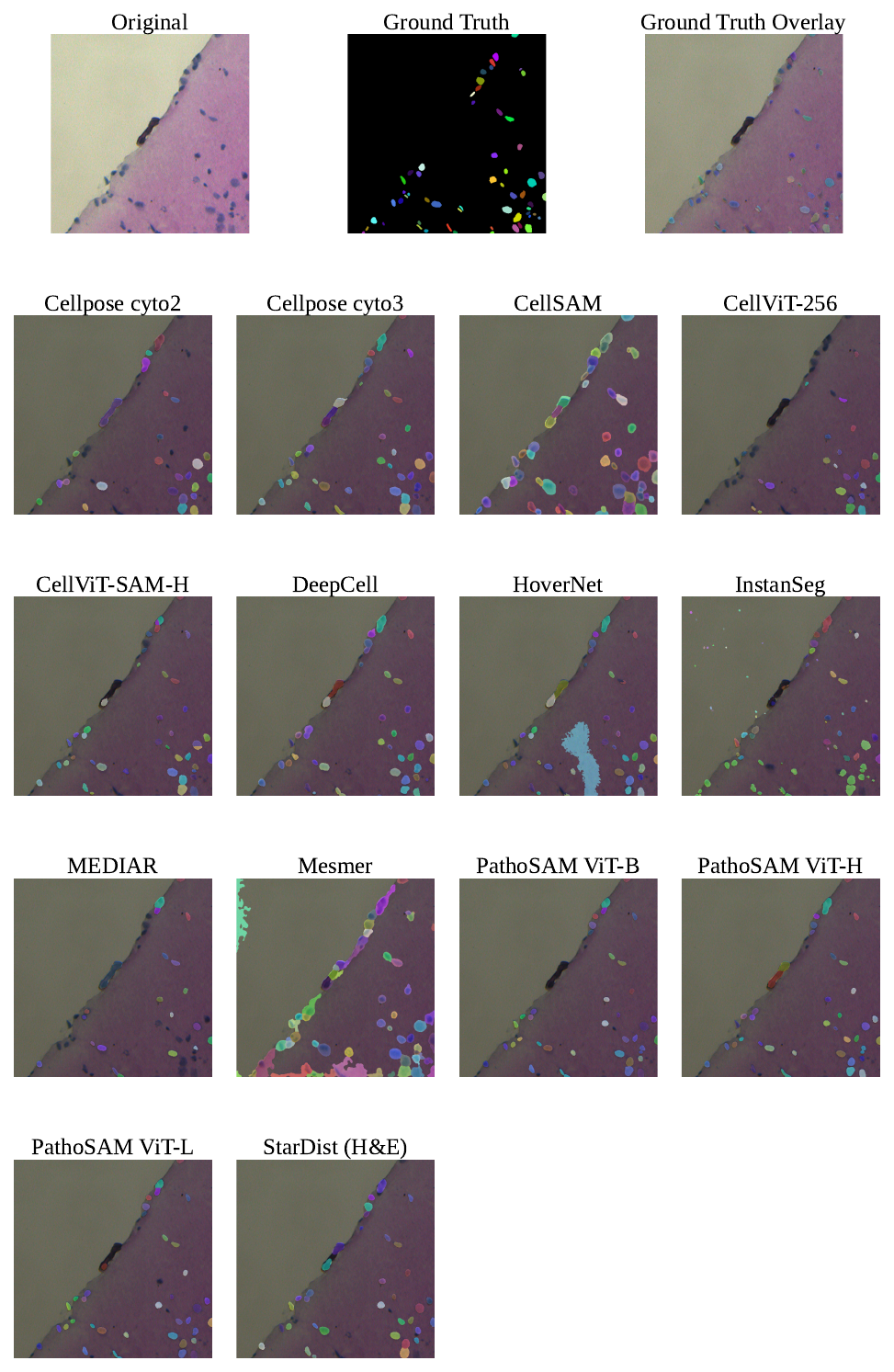}
		\caption{Qualitative zero-shot comparison for one image sampled from the HE dataset using random seed 0. The figure shows the input image, the ground-truth instance mask, the ground-truth overlay, and predictions from all models compatible with the image modality. Each prediction is overlaid on the input image.}
		\label{fig:dataset_grid_HE}
	\end{figure*}

	To provide a qualitative comparison of the segmentation outputs, we selected one test image from each dataset using a fixed random seed of 0. For each selected image, the corresponding supplementary figure (Figures~\ref{fig:dataset_grid_mIF}, \ref{fig:dataset_grid_DAPI}, \ref{fig:dataset_grid_ssDNA}, \ref{fig:dataset_grid_10xGenomics_DAPI}, \ref{fig:dataset_grid_10xGenomics_HE} and \ref{fig:dataset_grid_HE}) presents the original image, the ground-truth instance mask, an overlay of the ground-truth instance mask on the original image, and the predictions generated by all applicable models, overlaid on the original image.
	
	\begin{table*}[t]
		\centering
		\caption{Models, pretrained checkpoints, and software versions used in the benchmark.}
		\label{tab:model-checkpoints}
		
		\small
		\renewcommand{\arraystretch}{1.0}
		
		\begin{tabular}{@{}l p{0.27\textwidth} l p{0.30\textwidth}@{}}
			\toprule
			\textbf{Model name}
			& \textbf{Software name}
			& \textbf{Software version}
			& \textbf{Checkpoint name} \\
			\midrule
			
			StarDist (Fluo)    & stardist        & 0.9.2    & 2D\_versatile\_fluo \\
			StarDist (HE)      & stardist        & 0.9.2    & 2D\_versatile\_he \\
			InstanSeg          & instanseg-torch & 0.1.2    & single\_channel\_nuclei \\
			Cellpose cyto2     & cellpose        & 3.1.1    & cyto2torch\_0 \\
			Cellpose cyto3     & cellpose        & 3.1.1    & cyto3 \\
			Cellpose-SAM       & cellpose        & 4.0.8    & cpsam \\
			DeepCell           & DeepCell        & 0.12.10  & NuclearSegmentation-8.tar.gz \\
			Mesmer             & DeepCell        & 0.12.10  & MultiplexSegmentation-9.tar.gz \\
			CellSAM            & DeepCell        & 0.12.10  & cellsam-models\_v1.2.tar.gz \\
			MicroSAM ViT-T     & micro-sam       & 1.7.7    & vit\_t\_lm \\
			MicroSAM ViT-B     & micro-sam       & 1.7.7    & vit\_b\_lm \\
			MicroSAM ViT-L     & micro-sam       & 1.7.7    & vit\_l\_lm \\
			PathoSAM ViT-B     & micro-sam       & 1.7.7    & vit\_b\_histopathology \\
			PathoSAM ViT-L     & micro-sam       & 1.7.7    & vit\_l\_histopathology \\
			PathoSAM ViT-H     & micro-sam       & 1.7.7    & vit\_h\_histopathology \\
			HoverNet           & \url{github.com/vqdang/hover_net.git}
			& main branch
			& hovernet\_\allowbreak original\_\allowbreak kumar\_\allowbreak notype\_\allowbreak pytorch.tar \\
			MEDIAR             & \url{github.com/Lee-Gihun/MEDIAR}
			& main branch
			& from\_phase2.pth \\
			CellViT-256        & \url{github.com/TIO-IKIM/CellViT}
			& main branch
			& CellViT-256-x20.pth \\
			CellViT-SAM-H      & \url{github.com/TIO-IKIM/CellViT}
			& main branch
			& CellViT-SAM-H-x20.pth \\
			
			\bottomrule
		\end{tabular}
	\end{table*}
	
\subsection{Few-Shot Adaptation}
\label{subsec:fewshot_adaptation}
	
	We grouped the images in a dataset by tissue and sampled one image from every tissue group to form the support set for fine-tuning. The remaining images formed the test set. We first evaluated each compatible model on this test set before adaptation and then evaluated it again after fine-tuning on the support set. Fine-tuning was performed for 100 epochs under two regimes: full-model fine-tuning, in which all the model parameters were updated, and frozen-encoder fine-tuning, in which the encoder weights were not updated during training. We repeated the complete procedure ten times using random seeds 0--9.
	
\subsubsection{Aggregated Jaccard Index Plus (AJI$^{+}$)}
\label{subsubsec:aji_plus}
	
\begin{figure*}[tp]
		\centering
		\includegraphics[width=0.9\textwidth]{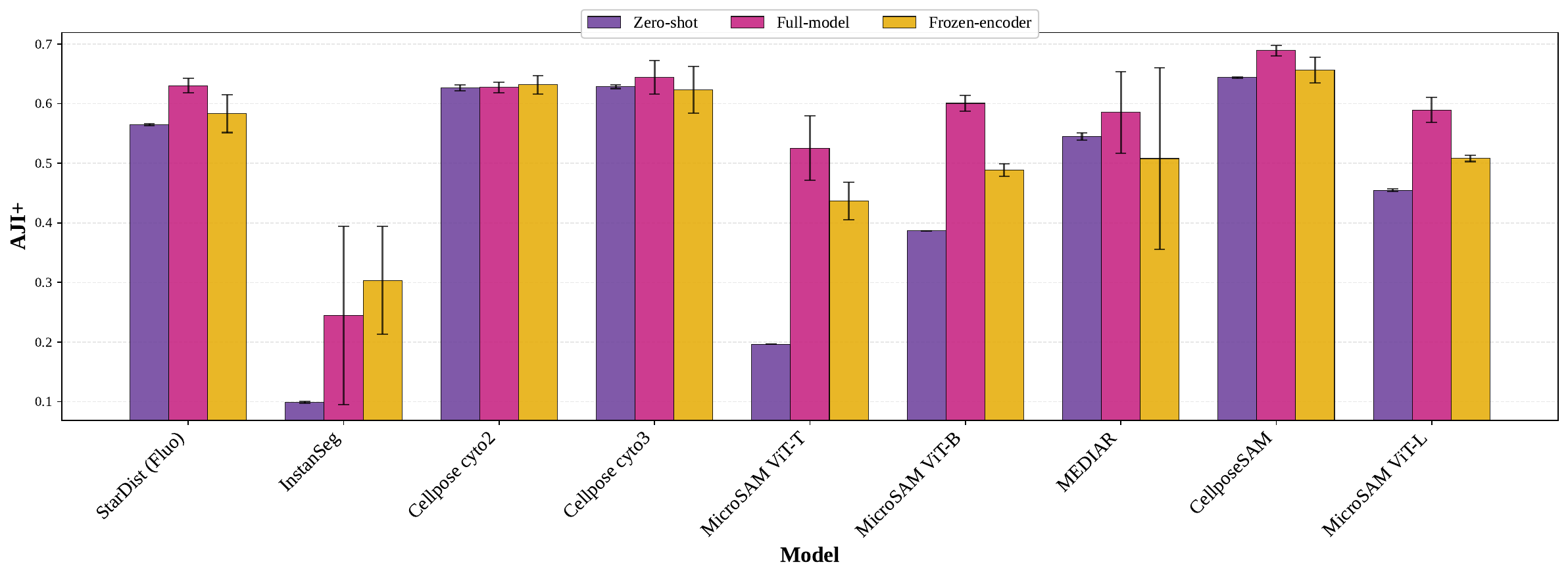}
		\caption{AJI$^{+}$ results for few-shot adaptation on the mIF dataset. The figure compares compatible, fine-tunable models under zero-shot inference, frozen-encoder and full-model fine-tuning. Each value reports the mean across ten runs.}
		\label{fig:aji_plus_mIF}
	\end{figure*}

\begin{figure*}[tp]
		\centering
		\includegraphics[width=0.9\textwidth]{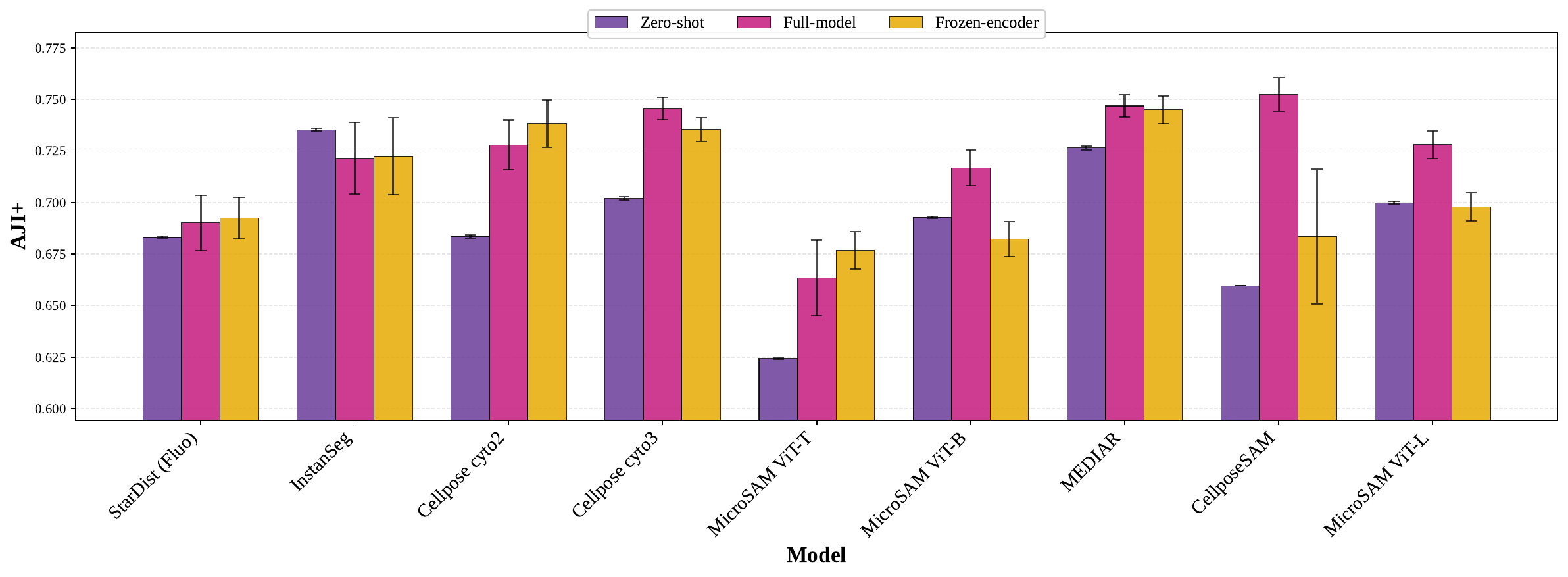}
		\caption{AJI$^{+}$ results for few-shot adaptation on the 10xGenomics\_DAPI dataset. The figure compares compatible, fine-tunable models under zero-shot inference, frozen-encoder and full-model fine-tuning. Each value reports the mean across ten runs.}
		\label{fig:aji_plus_10xGenomics_DAPI}
	\end{figure*}

\begin{figure*}[tp]
		\centering
		\includegraphics[width=0.9\textwidth]{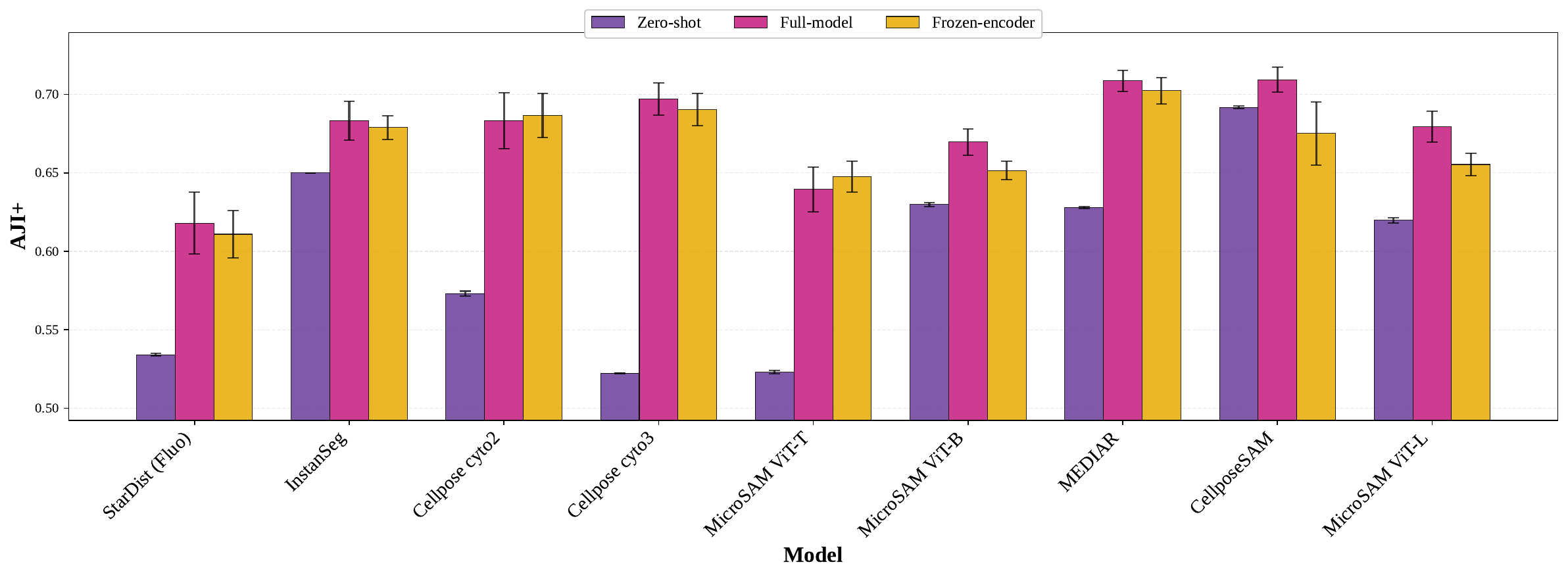}
		\caption{AJI$^{+}$ results for few-shot adaptation on the DAPI dataset. The figure compares compatible, fine-tunable models under zero-shot inference, frozen-encoder and full-model fine-tuning. Each value reports the mean across ten runs.}
		\label{fig:aji_plus_DAPI}
	\end{figure*}

\begin{figure*}[tp]
		\centering
		\includegraphics[width=0.9\textwidth]{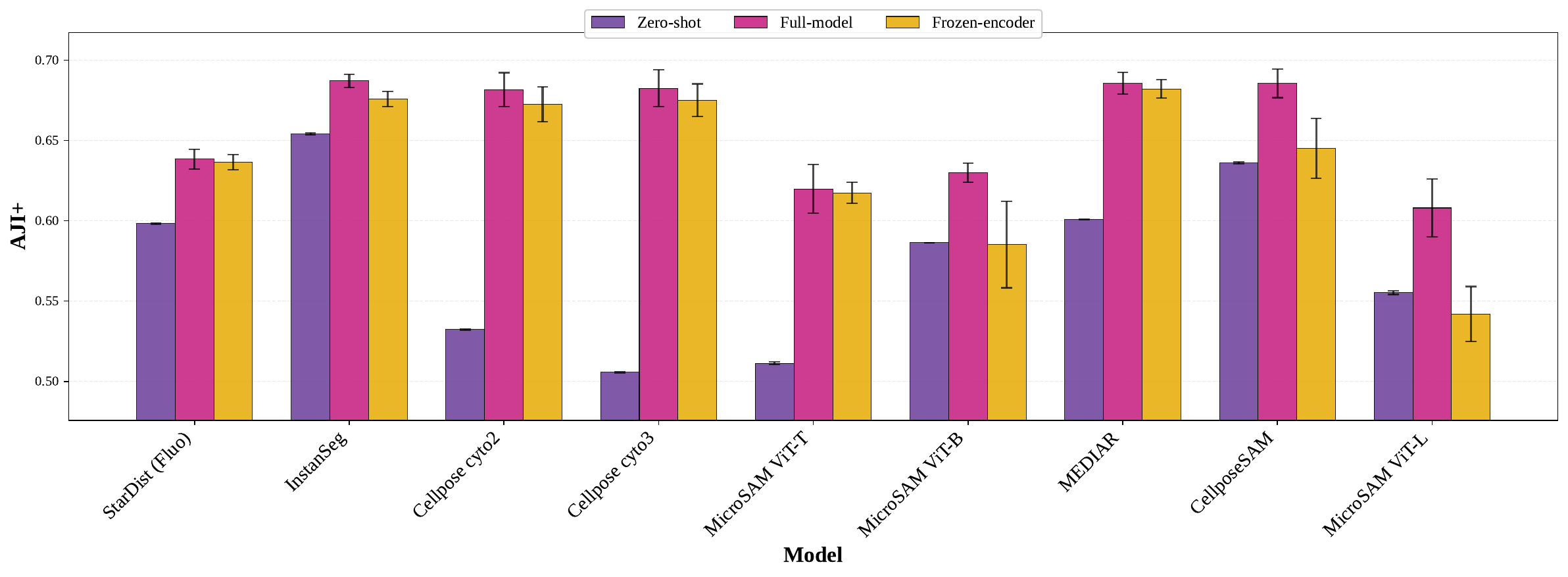}
		\caption{AJI$^{+}$ results for few-shot adaptation on the ssDNA dataset. The figure compares compatible, fine-tunable models under zero-shot inference, frozen-encoder and full-model fine-tuning. Each value reports the mean across ten runs.}
		\label{fig:aji_plus_ssDNA}
	\end{figure*}

\begin{figure*}[tp]
		\centering
		\includegraphics[width=0.9\textwidth]{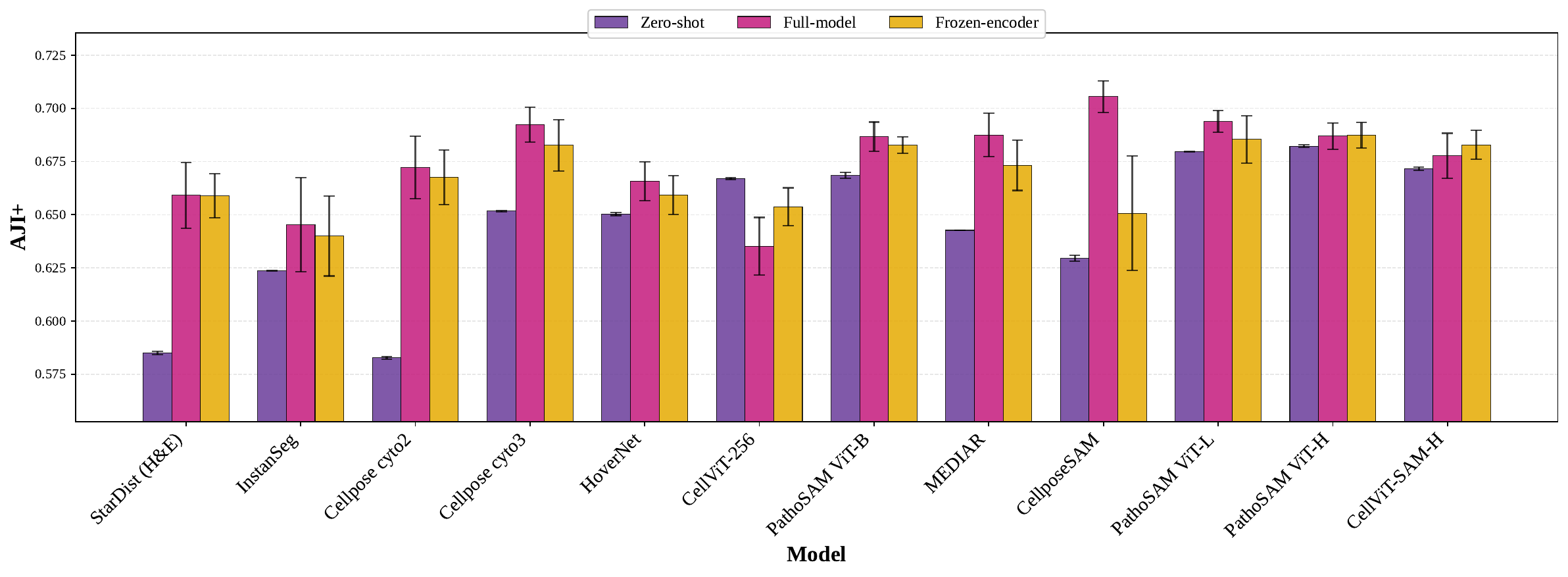}
		\caption{AJI$^{+}$ results for few-shot adaptation on the 10xGenomics\_HE dataset. The figure compares compatible, fine-tunable models under zero-shot inference, frozen-encoder and full-model fine-tuning. Each value reports the mean across ten runs.}
		\label{fig:aji_plus_10xGenomics_HE}
	\end{figure*}

\begin{figure*}[tp]
		\centering
		\includegraphics[width=0.9\textwidth]{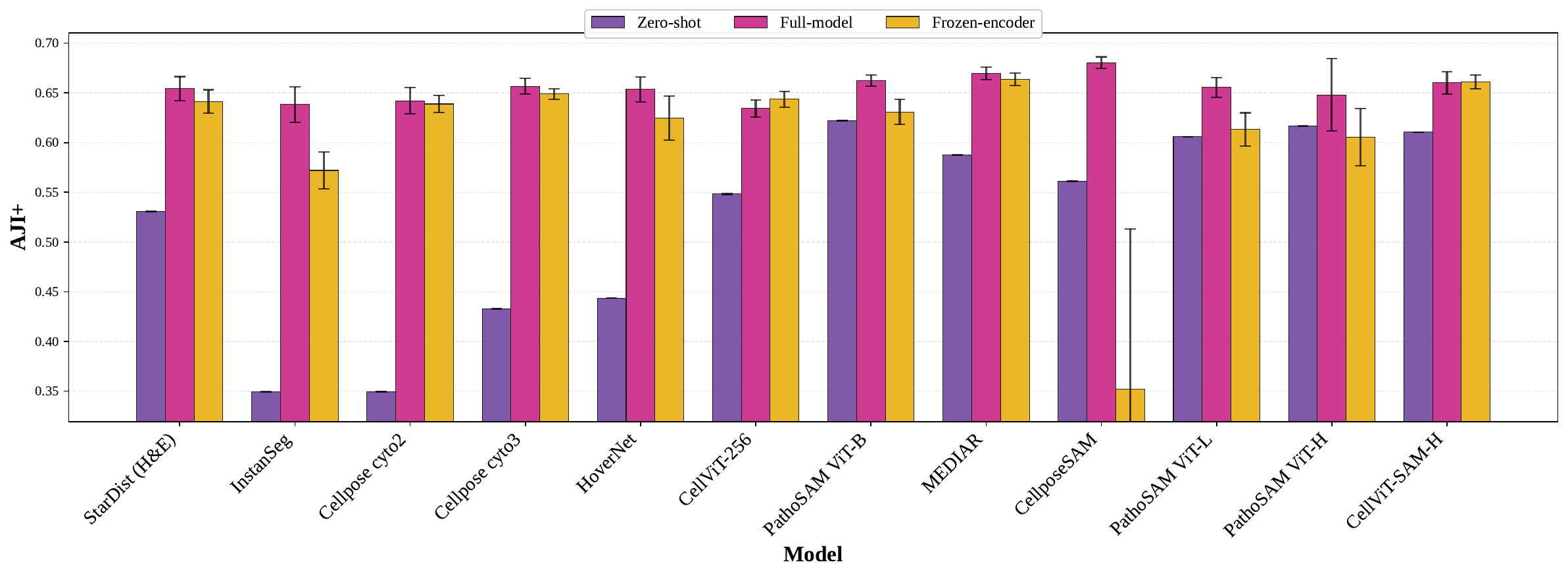}
		\caption{AJI$^{+}$ results for few-shot adaptation on the HE dataset. The figure compares compatible, fine-tunable models under zero-shot inference, frozen-encoder and full-model fine-tuning. Each value reports the mean across ten runs.}
		\label{fig:aji_plus_HE}
	\end{figure*}

	Figures~\ref{fig:aji_plus_mIF}, \ref{fig:aji_plus_10xGenomics_DAPI}, \ref{fig:aji_plus_DAPI}, \ref{fig:aji_plus_ssDNA}, \ref{fig:aji_plus_10xGenomics_HE}, and \ref{fig:aji_plus_HE} visualize the AJI$^{+}$ results for each dataset.
	
\subsubsection{Panoptic Quality (PQ)}
\label{subsubsec:pq}
	
\begin{figure*}[tp]
		\centering
		\includegraphics[width=0.9\textwidth]{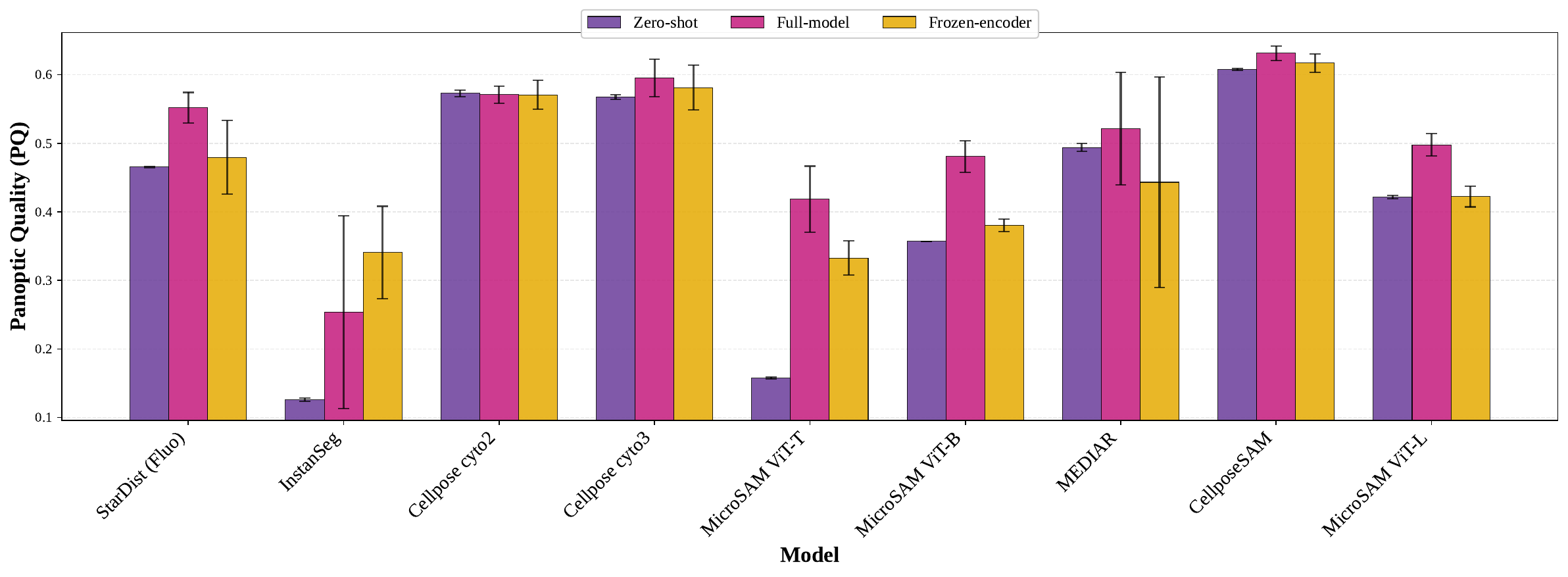}
		\caption{PQ results for few-shot adaptation on the mIF dataset. The figure compares compatible, fine-tunable models under zero-shot inference, frozen-encoder and full-model fine-tuning. Each value reports the mean across ten runs.}
		\label{fig:pq_mIF}
	\end{figure*}

\begin{figure*}[tp]
		\centering
		\includegraphics[width=0.9\textwidth]{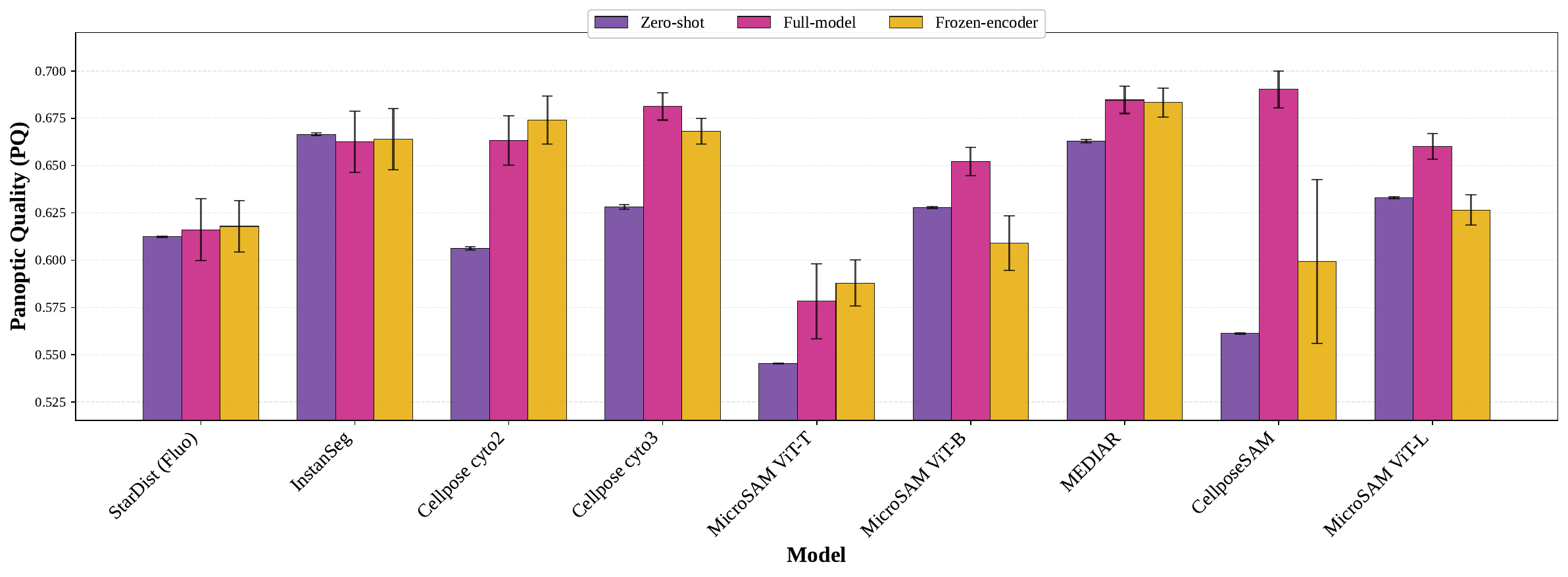}
		\caption{PQ results for few-shot adaptation on the 10xGenomics\_DAPI dataset. The figure compares compatible, fine-tunable models under zero-shot inference, frozen-encoder and full-model fine-tuning. Each value reports the mean across ten runs.}
		\label{fig:pq_10xGenomics_DAPI}
	\end{figure*}

\begin{figure*}[tp]
		\centering
		\includegraphics[width=0.9\textwidth]{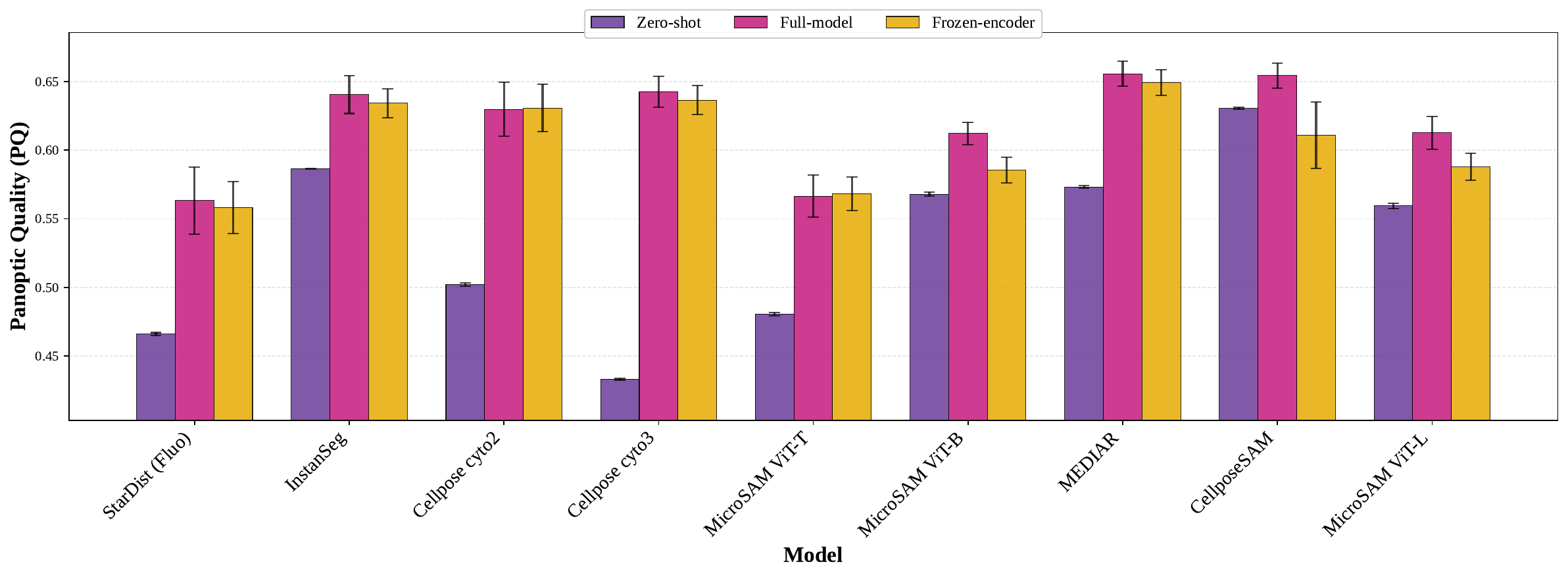}
		\caption{PQ results for few-shot adaptation on the DAPI dataset. The figure compares compatible, fine-tunable models under zero-shot inference, frozen-encoder and full-model fine-tuning. Each value reports the mean across ten runs.}
		\label{fig:pq_DAPI}
	\end{figure*}

\begin{figure*}[tp]
		\centering
		\includegraphics[width=0.9\textwidth]{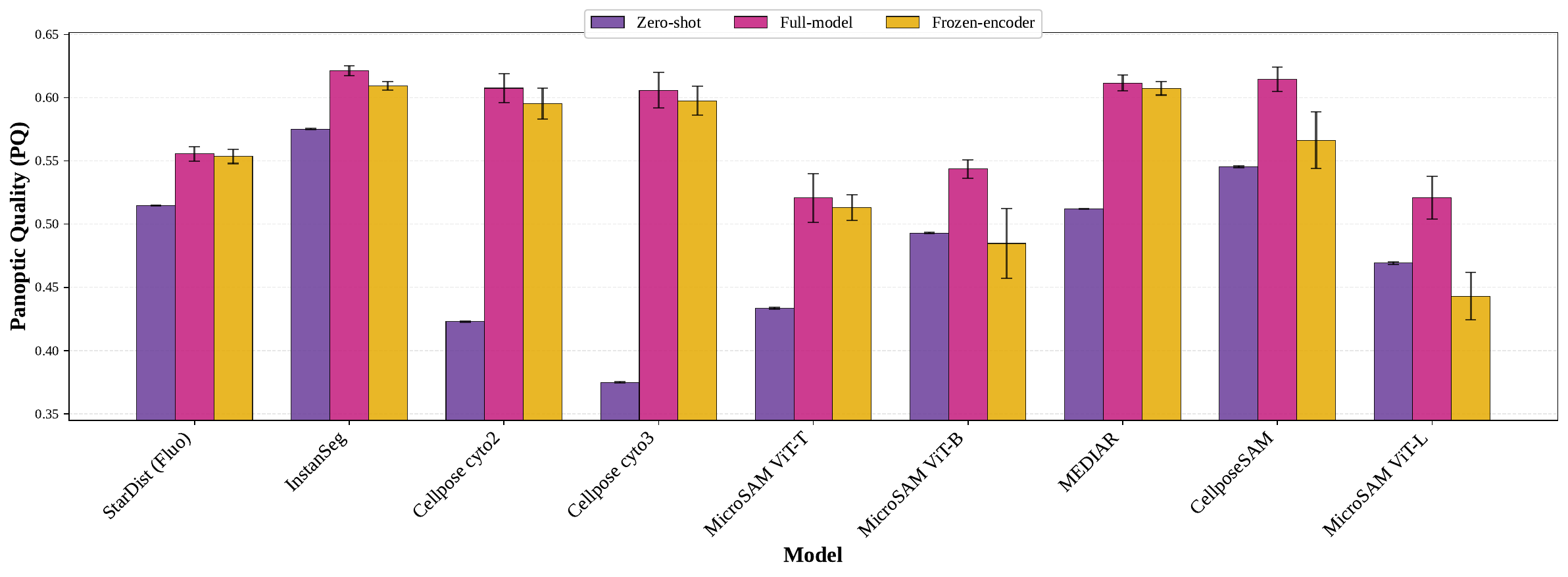}
		\caption{PQ results for few-shot adaptation on the ssDNA dataset. The figure compares compatible, fine-tunable models under zero-shot inference, frozen-encoder and full-model fine-tuning. Each value reports the mean across ten runs.}
		\label{fig:pq_ssDNA}
	\end{figure*}

\begin{figure*}[tp]
		\centering
		\includegraphics[width=0.9\textwidth]{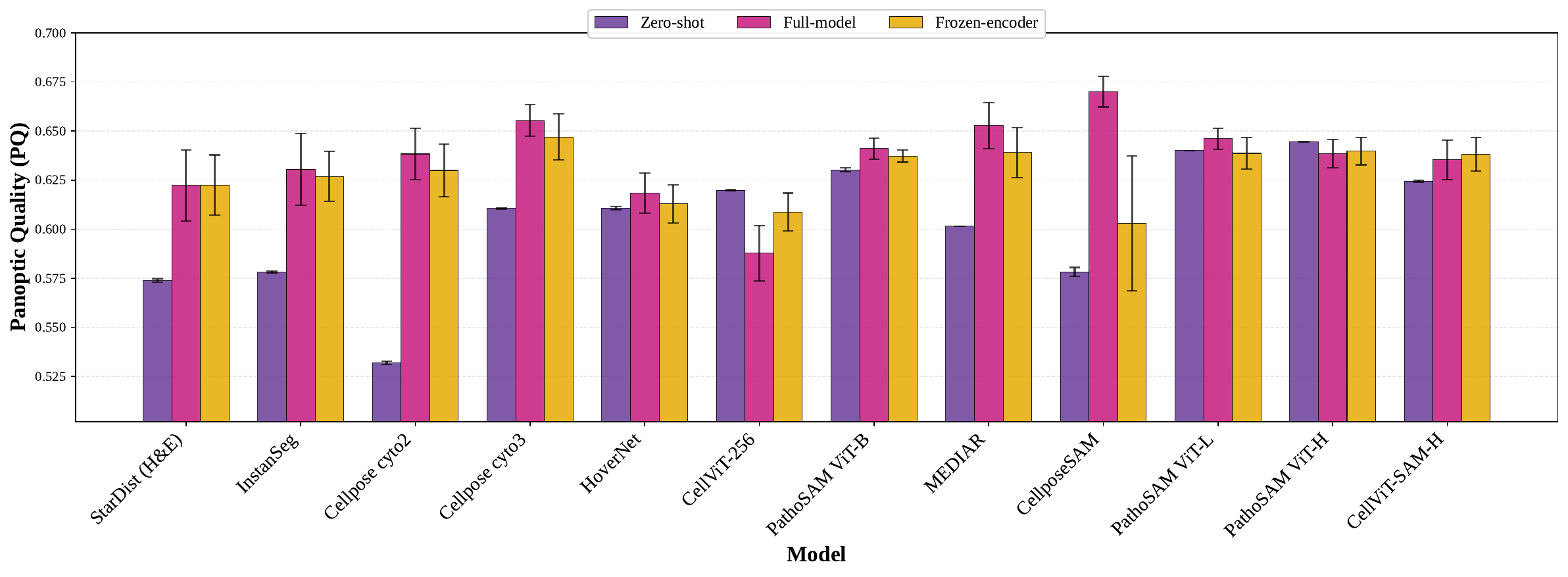}
		\caption{PQ results for few-shot adaptation on the 10xGenomics\_HE dataset. The figure compares compatible, fine-tunable models under zero-shot inference, frozen-encoder and full-model fine-tuning. Each value reports the mean across ten runs.}
		\label{fig:pq_10xGenomics_HE}
	\end{figure*}

\begin{figure*}[tp]
		\centering
		\includegraphics[width=0.9\textwidth]{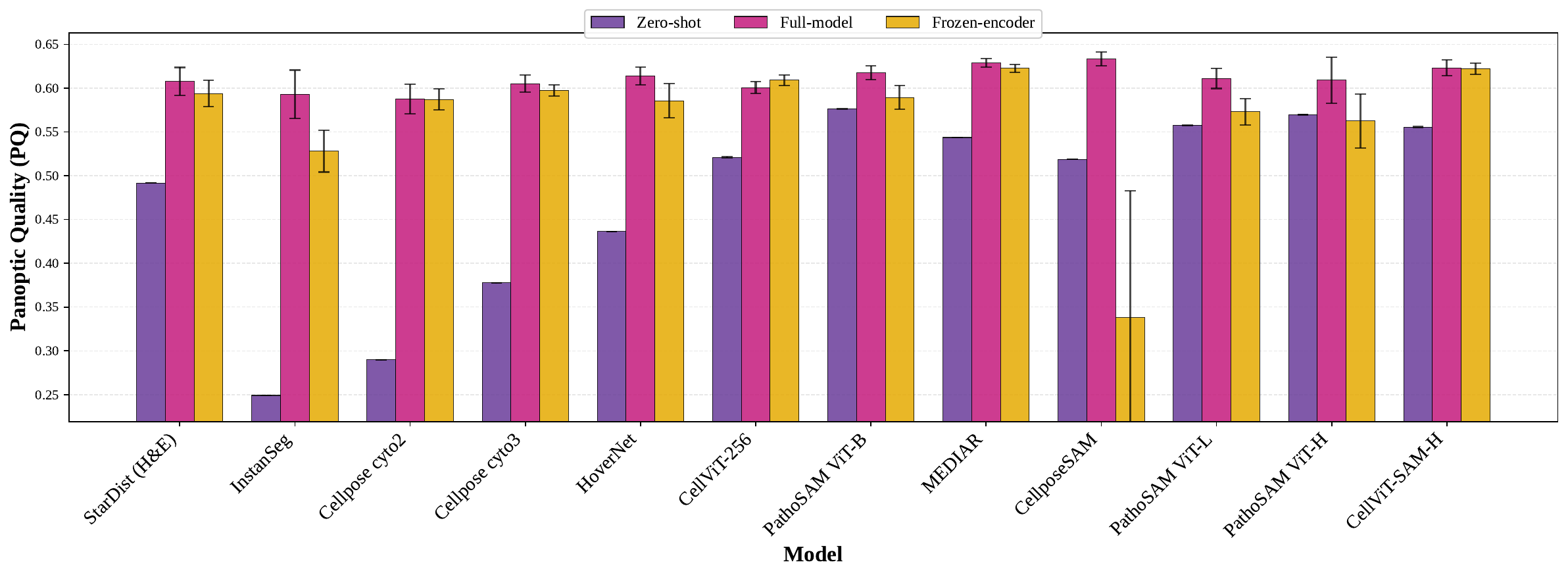}
		\caption{PQ results for few-shot adaptation on the HE dataset. The figure compares compatible, fine-tunable models under zero-shot inference, frozen-encoder and full-model fine-tuning. Each value reports the mean across ten runs.}
		\label{fig:pq_HE}
	\end{figure*}

	Figures~\ref{fig:pq_mIF}, \ref{fig:pq_10xGenomics_DAPI}, \ref{fig:pq_DAPI}, \ref{fig:pq_ssDNA}, \ref{fig:pq_10xGenomics_HE}, and \ref{fig:pq_HE} visualize the PQ metrics for each dataset.
	
\subsubsection{Normalized Surface Dice (NSD)}
\label{subsubsec:nsd}
	
\begin{figure*}[tp]
		\centering
		\includegraphics[width=0.9\textwidth]{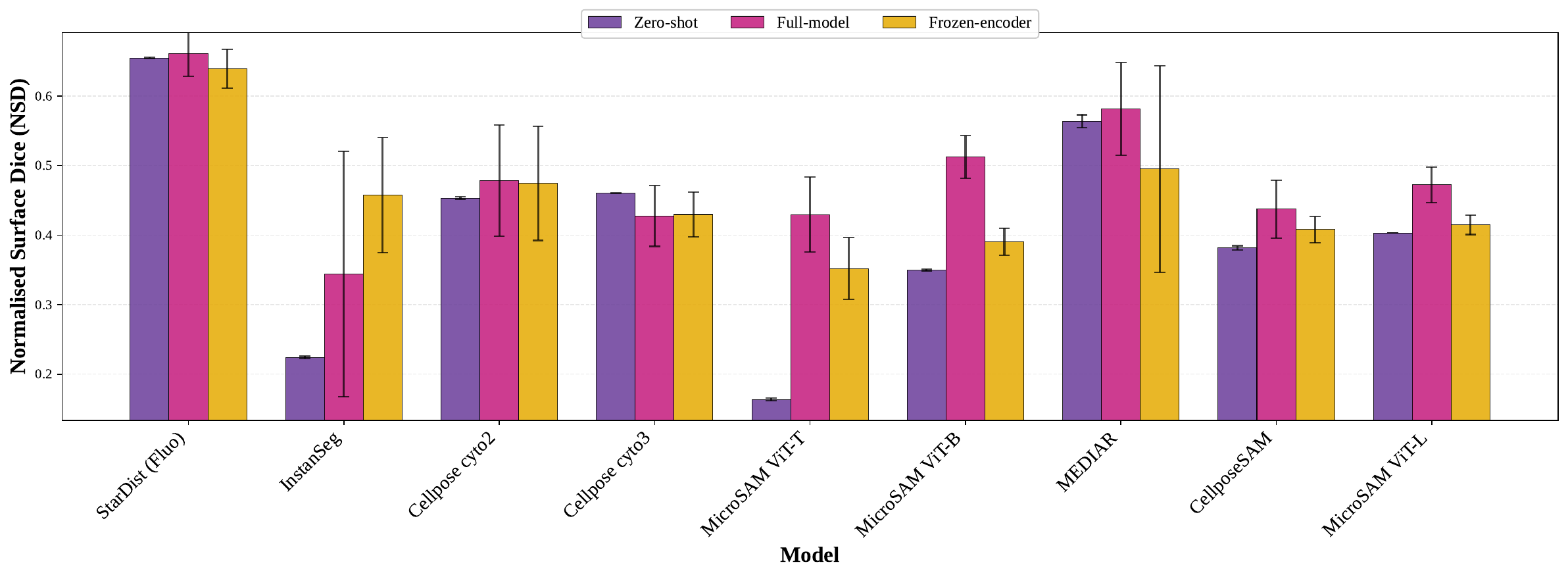}
		\caption{NSD results for few-shot adaptation on the mIF dataset. The figure compares compatible, fine-tunable models under zero-shot inference, frozen-encoder and full-model fine-tuning. Each value reports the mean across ten runs.}
		\label{fig:nsd_mIF}
	\end{figure*}

\begin{figure*}[tp]
		\centering
		\includegraphics[width=0.9\textwidth]{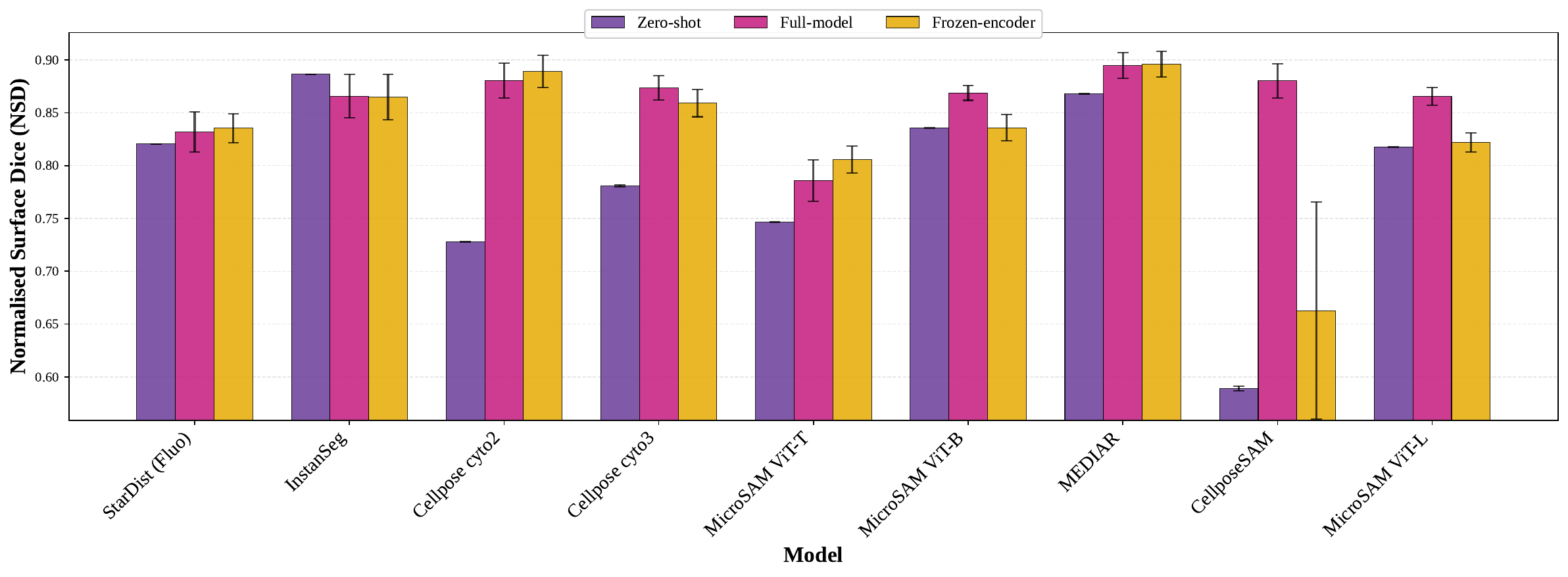}
		\caption{NSD results for few-shot adaptation on the 10xGenomics\_DAPI dataset. The figure compares compatible, fine-tunable models under zero-shot inference, frozen-encoder and full-model fine-tuning. Each value reports the mean across ten runs.}
		\label{fig:nsd_10xGenomics_DAPI}
	\end{figure*}

\begin{figure*}[tp]
		\centering
		\includegraphics[width=0.9\textwidth]{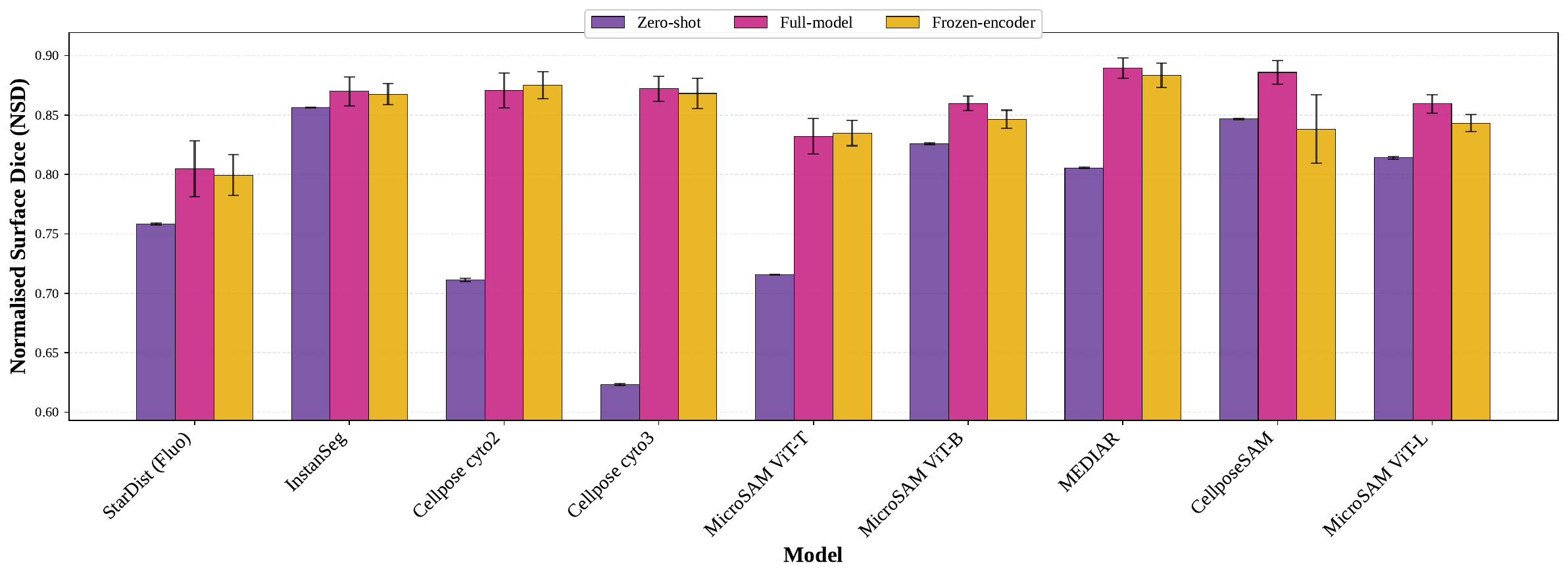}
		\caption{NSD results for few-shot adaptation on the DAPI dataset. The figure compares compatible, fine-tunable models under zero-shot inference, frozen-encoder and full-model fine-tuning. Each value reports the mean across ten runs.}
		\label{fig:nsd_DAPI}
	\end{figure*}

\begin{figure*}[tp]
		\centering
		\includegraphics[width=0.9\textwidth]{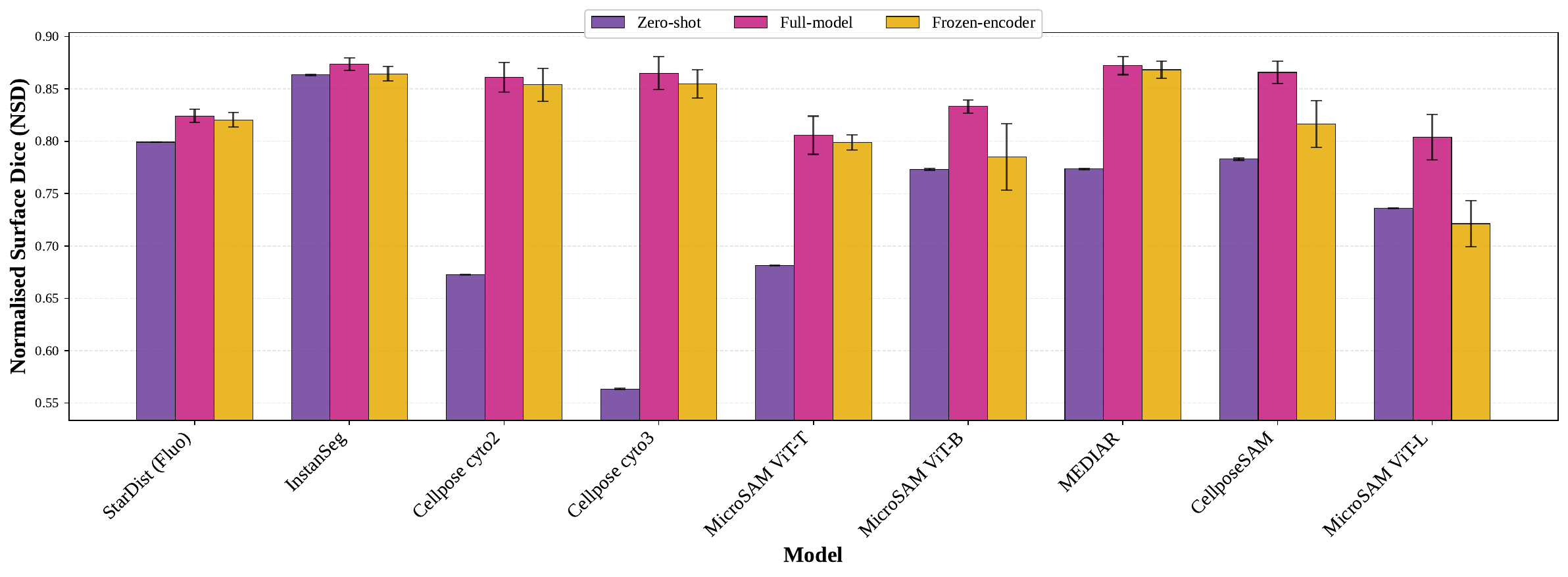}
		\caption{NSD results for few-shot adaptation on the ssDNA dataset. The figure compares compatible, fine-tunable models under zero-shot inference, frozen-encoder and full-model fine-tuning. Each value reports the mean across ten runs.}
		\label{fig:nsd_ssDNA}
	\end{figure*}

\begin{figure*}[tp]
		\centering
		\includegraphics[width=0.9\textwidth]{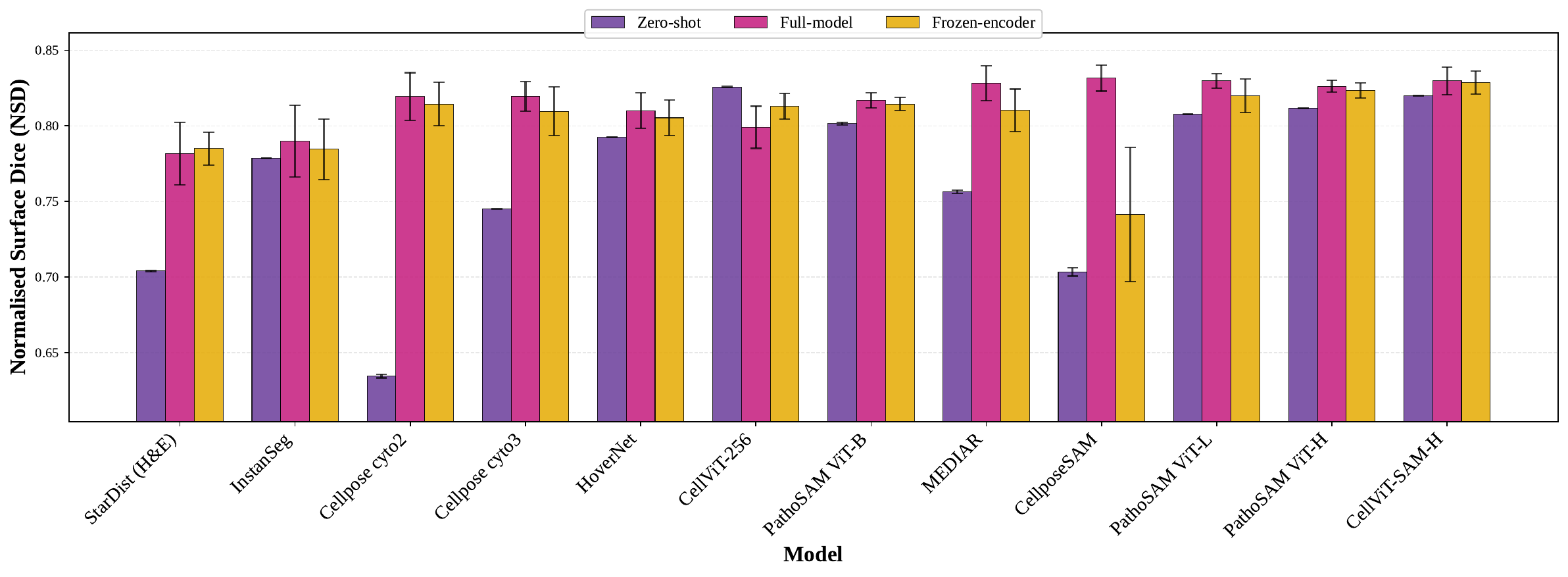}
		\caption{NSD results for few-shot adaptation on the 10xGenomics\_HE dataset. The figure compares compatible, fine-tunable models under zero-shot inference, frozen-encoder and full-model fine-tuning. Each value reports the mean across ten runs.}
		\label{fig:nsd_10xGenomics_HE}
	\end{figure*}

\begin{figure*}[tp]
		\centering
		\includegraphics[width=0.9\textwidth]{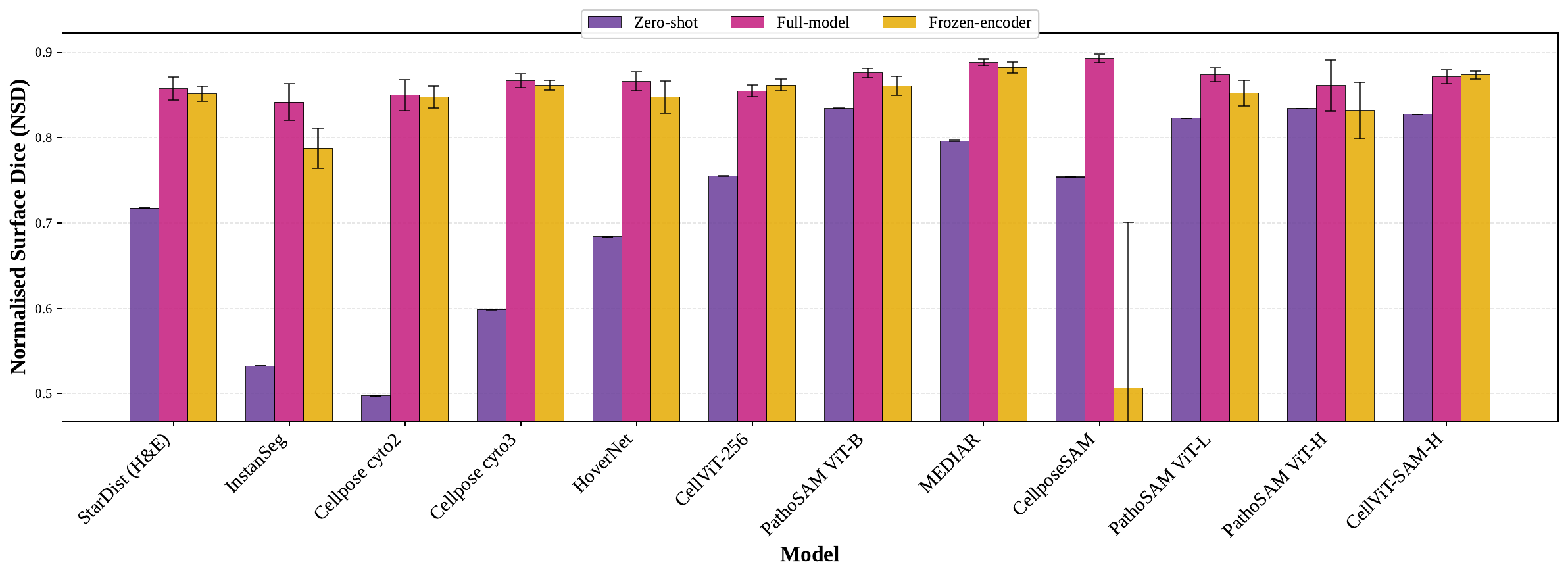}
		\caption{NSD results for few-shot adaptation on the HE dataset. The figure compares compatible, fine-tunable models under zero-shot inference, frozen-encoder and full-model fine-tuning. Each value reports the mean across ten runs.}
		\label{fig:nsd_HE}
	\end{figure*}

	Figures~\ref{fig:nsd_mIF}, \ref{fig:nsd_10xGenomics_DAPI}, \ref{fig:nsd_DAPI}, \ref{fig:nsd_ssDNA}, \ref{fig:nsd_10xGenomics_HE}, and \ref{fig:nsd_HE} visualize the NSD metrics for each dataset.
	
\end{document}